\pdfoutput=1
\documentclass[acmsmall,screen]{acmart}

\usepackage{longtable}
\usepackage[flushleft]{threeparttable}
\usepackage{threeparttablex}
\usepackage{lscape}
\usepackage{xcolor}
\usepackage{colortbl}

\AtBeginDocument{%
  }

\setcopyright{acmlicensed}
\copyrightyear{2024}
\acmYear{2024}
\acmDOI{XXXXXXX.XXXXXXX}

\acmJournal{TIST}
\acmVolume{37}
\acmNumber{4}
\acmArticle{111}
\acmMonth{8}

\newcommand{\ie}{\emph{i.e.}}
\newcommand{\eg}{\emph{e.g.}}
\newcommand{\etc}{\emph{etc}}

\newcommand{\review}[1]{\textcolor{black}{#1}}

\usepackage{todonotes}

\begin{document}

%\title{Concept Drift Adaptation in Text Stream Mining Settings: A Comprehensive Review}
\title{Concept Drift Adaptation in Text Stream Mining Settings: A {Systematic} Review}

\author{Cristiano Mesquita Garcia}
\email{cristiano.garcia@ifsc.edu.br}
\orcid{0000-0002-7475-146X}
\affiliation{%
  \institution{Instituto Federal de Santa Catarina}
  \city{Caçador}
  \country{Brazil}
}
\affiliation{%
  \institution{Programa de Pós-Graduação em Informática (PPGIa), Pontifícia Universidade Católica do Paraná (PUCPR)}
  \city{Curitiba}
  \country{Brazil}
}

\author{Ramon Abilio}
\orcid{0000-0002-7197-5951}
\affiliation{%
  \institution{Instituto Federal de São Paulo}
  \city{Capivari}
  \country{Brazil}}
\affiliation{%
  \institution{Universidade de Campinas}
  \city{Limeira}
  \country{Brazil}
}
\email{ramon.abilio@ifsp.edu.br}

\author{Alessandro Lameiras Koerich}
\orcid{0000-0001-5879-7014}
\affiliation{%
  \institution{École de Technologie Supérieure, Université du Québec}
  \city{Montréal}
  \country{Canada}
}
\email{alessandro.koerich@etsmtl.ca}

\author{Alceu de Souza Britto Jr}
\orcid{0000-0002-3064-3563}
\affiliation{%
 \institution{Programa de Pós-Graduação em Informática (PPGIa), Pontifícia Universidade Católica do Paraná (PUCPR)}
 \city{Curitiba}
 \country{Brazil}}
 \affiliation{%
 \institution{Universidade Estadual de Ponta Grossa}
 \city{Ponta Grossa}
 \country{Brazil}}
\email{alceu@ppgia.pucpr.br}

\author{Jean Paul Barddal}
\orcid{0000-0001-9928-854X}
\affiliation{%
  \institution{Programa de Pós-Graduação em Informática (PPGIa), Pontifícia Universidade Católica do Paraná (PUCPR)}
  \city{Curitiba}
  \country{Brazil}}
\email{jean.barddal@ppgia.pucpr.br}

\renewcommand{\shortauthors}{Garcia et al.}

\begin{abstract}
{The society produces} textual data online in several ways, \eg, {via} reviews and social media posts. Therefore, numerous researchers have been working on discovering patterns in textual data {that} can indicate peoples' opinions, interests, \etc. Most tasks regarding natural language processing are addressed using traditional machine learning methods and static datasets. This setting can lead to several problems, \eg, outdated datasets and models, which degrade {in} performance over time. 
{This is particularly true regarding concept drift, in which the data distribution changes over time. Furthermore, text streaming scenarios also exhibit further challenges, such as the high speed at which data arrives over time}. Models for stream scenarios must adhere to the aforementioned constraints while learning from the stream, {thus} storing texts for limited periods and consuming low memory. This study presents a systematic literature review regarding concept drift adaptation in text stream scenarios. Considering well-defined criteria, we selected {48} papers {published between 2018 and August 2024} to unravel aspects such as text drift categories, detection types, model update mechanisms, stream mining tasks addressed, and text representation methods and their update mechanisms. Furthermore, we discussed drift visualization and simulation and listed real-world datasets used in the selected papers. {Finally, we brought forward a discussion on existing works in the area, also highlighting open challenges and future research directions for the community.}
%Therefore, this manuscript reviews concept drift adaptation in text stream mining scenarios.
\end{abstract}




\received{20 February 2024}
\received[revised]{10 October 2024}
\received[accepted]{XX XXXX 2024}

\maketitle

\section{Introduction}
{Intelligent systems (IS) based on} machine learning (ML) have been increasingly researched as processing power has increased and storage capacity has been cheapened. The development of frameworks and libraries, such as Weka \citep{hall2009weka} and Scikit-Learn \citep{scikit-learn}, has enabled the rapid development and deployment of ML models and their applications. Moreover, Tensorflow \citep{tensorflow2015-whitepaper}, Keras \citep{chollet2015keras}, PyTorch \citep{paszke2019pytorch} and HuggingFace \citep{wolf2019huggingface} are more contemporary enablers that are related to deep learning models and generally rely on graphic processing units (GPUs) to expedite the training process. Therefore, there has been an increase in the development of ML applications, such as credit scoring \citep{barddal2020lessons}, emotion recognition \citep{delazeri2022evaluation}, and cryptocurrency pricing prediction \citep{garcia2019evolvable}. 

%Data, the main raw resource for generating good machine learning models, is being produced by software, sensors, processes, and people. 
Software, sensors, processes, and humans generate data, the primary raw resource for developing ML models.
Humans, in particular, produce a considerable amount of unstructured data on the Internet, especially on social media, where users upload pictures and post opinions regarding anything, including products, artists, and politicians. Therefore, social networks have been considered a low-cost, rapid source of information, with the collected data utilized for election prediction \citep{dwi2015twitter,brito2023machine,tsai2019machine}, stance analysis \citep{p33}, event detection \citep{p12}, \etc. 

Texts are unstructured data. Most ML approaches expect numbers as input parameters, so texts cannot be directly used as input for ML methods. To overcome the aforementioned limitation, text must be processed, cleaned, sometimes standardized, and converted to fixed-size numerical vector representations. The conversion from unstructured to structured data is also known as feature extraction \citep{ahuja2019impact,thuma2023benchmarking}.
%\citep{ahuja2019impact,thuma2023benchmarking}
%It is necessary to process the texts, clean them, and convert them to numerical representations to obtain knowledge from texts using ML because most ML approaches expect numbers as input parameters. Studies related to this activity have increased recently and can be categorized as natural language processing (NLP) studies. 
Recent {advances in} natural language processing (NLP) advances have simplified text-based real-life applications. It is worth mentioning Word2Vec \citep{mikolov2013distributed}, which is a neural network-based approach for generating word embeddings (vector representation), and BERT \citep{devlin2018bert}, a bidirectional transformer-based modeling architecture, that can be applied in tasks such as {sentiment analysis, and spam detection.} 
%sequence-to-sequence learning, \eg, language translation, text generation, and text classification. 
One advantage of the aforementioned methods is their reuse capability. 
%\akst{There are several pre-trained models available} \ak{Several pre-trained models are available} 
Several pre-trained models are available on the Internet in specialized hubs such as HuggingFace\footnote{https://huggingface.co/models}. A pre-trained model can aid in extracting features from text and use them as input for a classifier, \eg, a sentiment classifier. The time necessary to develop the final ML model can be drastically reduced if {tailoring} a representation-learning model from scratch is {not required}. {For instance, using pre-trained models is a common approach when the target application is aligned with the context in which the pre-trained model was built. Additionally, in the case of using the pre-trained model in a transfer learning fashion, it has been shown possible to fine-tune the ML model, the representation model, or both, depending on the computational resources available and the expected outcomes of the intelligent system}.

Although the aforementioned approaches were initially designed for batch learning, it is possible to use pre-trained models to extract features in data stream scenarios. Data streams are considered a collection of sequential data that comes consecutively, or in small batches, in a timely order \citep{bifet2018machine}. Thus, for ML models in data streams, there are challenges such as learning from the data the instant it arrives, adapting the model in case of pattern change, and keeping it concise. %Text streams are a specialization of data streams, respecting all the aforementioned characteristics.
\textit{Text streams} represent a continuous flow of textual data, such as social media updates, news articles, customer reviews, or online discussions. % In the context of text streams, concept drift occurs when the underlying patterns and relationships within the textual data shift, making previously learned models or approaches ineffective. Concept drift in text streams arises from the dynamic nature of language, evolving language, and trends, changing context and sentiment, and diversity of data sources. Therefore, understanding and addressing concept drift is crucial for maintaining the accuracy, relevance, and ethical integrity of ML models for text stream processing.
Several social networks and news agencies provide application programming interfaces (API) that function as a text stream. {X \footnote{https://x.com/} (former Twitter)} is an example of a social media platform that offers API access {to its data}. {Conversely, Massive Online Analysis (MOA) \citep{bifet2010moa} and RiverML \citep{montiel2021river} have been enablers of experimentation and development of methods for stream mining, despite not targeting textual data specifically.}

%In the case of pattern change, it is commonly referred to as \textit{concept drift} in the literature. 
{A data pattern change is commonly referred to as \textit{concept drift} in the scientific machine learning literature}. 
\textit{Concept drift} is a phenomenon that occurs in data subject to non-stationary processes \citep{bifet2018machine,gama2014survey}. 
In real life, for example, changes may occur in temperature or customer purchasing patterns across given analyzed periods.
%These changes will also be reflected in the data. In the data context, this kind of change in pattern is called concept drift. 
Concept drift imposes several difficulties for ML models, \eg, if concept drifts are not captured and managed by the model, its performance will degrade over time. It can be even more challenging for ML approaches that require the processing of text streams due to the constraints inherent to streaming learning settings, such as the speed of the stream.
%In the context of text streams, concept drift occurs when the underlying patterns and relationships within the textual data shift, making previously learned models or approaches ineffective. Concept drift in text streams arises from the dynamic nature of language, evolving language, and trends, changing context and sentiment, and diversity of data sources. Therefore, understanding and addressing concept drift is crucial for maintaining the accuracy, relevance, and ethical integrity of ML models for text stream processing.
In text streams, concept drift occurs when the underlying patterns and relationships within the textual data shift, making previously learned models or approaches ineffective. Concept drift in text streams arises from the dynamic {and} evolving nature of language and its data sources, {where trends, contexts, and sentiments change over time}. Therefore, understanding and addressing concept drift is crucial for maintaining the accuracy, relevance, and ethical integrity of ML models for text stream processing.

{In addition to concept drift, a specialized type of drift can emerge in texts: \textit{semantic shift}. Semantic shift also referred to as \textit{semantic change} \cite{de2024survey}, regards changes in word meanings over time \cite{bloomberg33}. These changes can affect not only the words themselves but also their entire context, which can influence the performance in downstream tasks such as classification, for example. Another interesting aspect regarding text is that they cannot be treated {in its raw form, thus requiring processing to be represented in a numeric format} so that the drifts/semantic shifts {are to} be detected. {Even though} some authors argue the existence of different types of drifts/semantic shifts in real-world datasets, \eg, \citet{p18}, these drifts are difficult to label. This is supported by one of the findings reported in this paper, in which only one dataset had drifts labeled \cite{p41} and corresponded to sentiment drift events identified during a soccer match.}

%--- Falar das dificuldades de processar texto em tempo real.
Processing text and learning in stream scenarios {is} challenging due to the requirements for ML models to function effectively in such scenarios.
%the constraints of streaming scenarios
The requirements include: (i) learning from the data as it arrives; (ii) discarding the data after learning from it; (iii) performing all operations in a single-pass fashion \citep{gama2004learning,bifet2018machine}. 
In addition, NLP-related activities can be challenging in stream scenarios, such as maintaining an updated and concise vocabulary and updating representations when possible. 
{Therefore, text stream scenarios are even more restrictive since the NLP-related activities must also be designed to ideally perform one-pass operations.}
%Therefore, the constraints of streaming scenarios are more restrictive when handling text streams.
%\JPBCOMMENT{sao mais restritivas ou na verdade são mais restrições? parecem ser coisas diferentes}
%\CGCOMMENT{Reescrevi.}
% --- Falar sobre text streams
%\textit{Data streams} are an abstraction where the data arrives continuously and sequentially and is temporally ordered \citep{MOA-Book-2018}. A \textit{text stream} is a stream that respects the concepts of data streams; however, it receives a stream of texts instead. Nowadays, several social networks provide application programming interfaces (API) that function as a text stream. Twitter\footnote{https://twitter.com/} is an example of a social media platform that offers API access. 

\begin{figure}[!htp]
\includegraphics[width=0.4\textwidth]{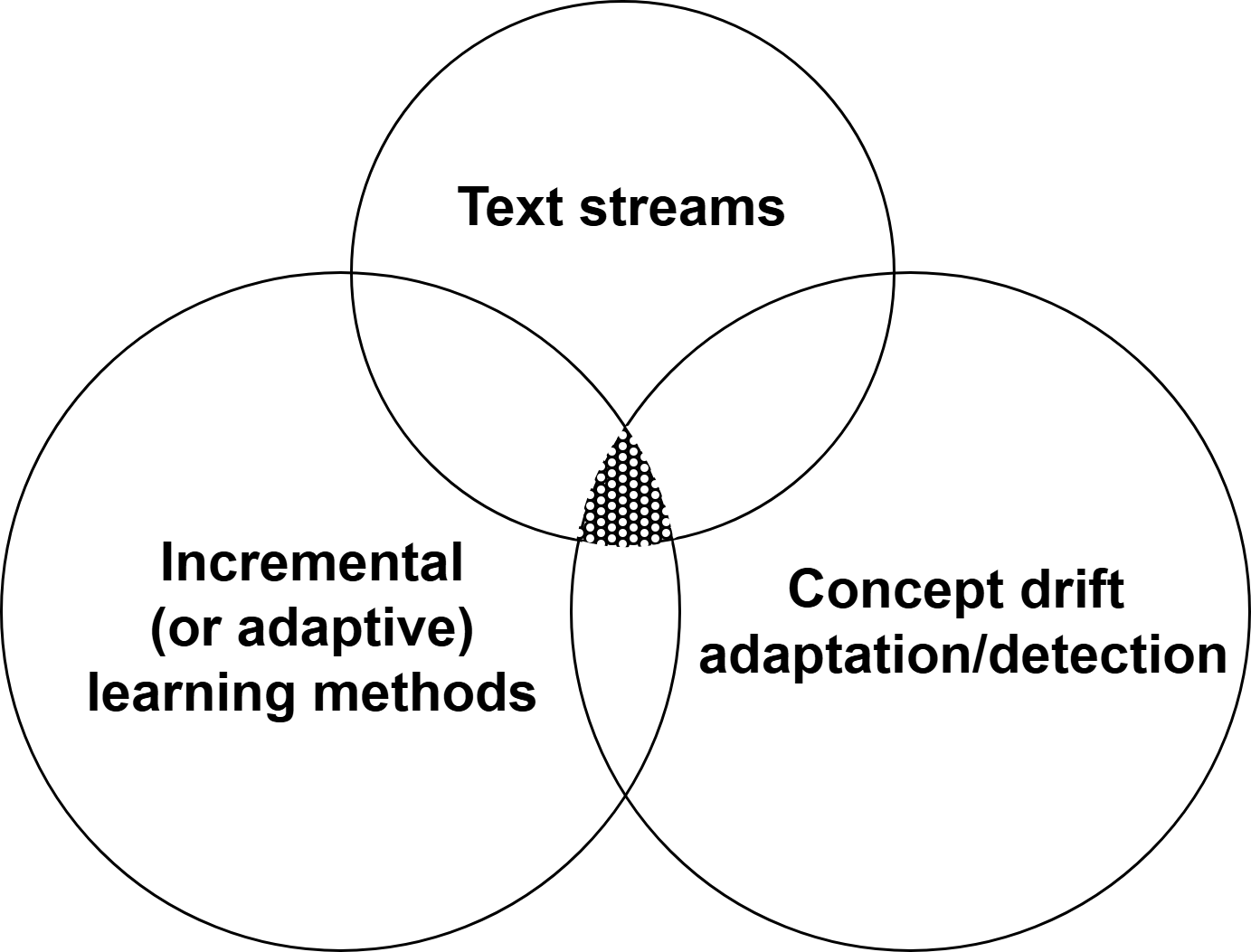}
\centering

\caption{{Intersection of subjects of interest in this review. We are mainly interested in the papers on the intersection (hatched area) of these three subjects.}}
\label{fig:intersection}
\end{figure}

%--- Objetivos da Revisão Sistemática
{Motivated by the challenges and constraints of text streams, the existence of concept drift, and the characteristic of intelligent systems to learn incrementally in these scenarios,} this study {offers} a systematic review regarding concept drift {adaptation} in text streams. {Fig.~\ref{fig:intersection} provides our scope for this review, in which we target the intersection of text streams, concept drift detection/adaptation, and works that introduce novel incremental and adaptive learning methods for such scenarios}. {In other words, this} systematic review {unravels} the most common approaches to managing concept drift, {updating the model to recover from concept drifts, text representation methods,} datasets, and applications in challenging scenarios such as text streams. 
This work is organized as follows: Section \ref{sec:dsm} introduces data stream mining and presents the aspects of concept drift, semantic shift, and concept drift detectors. Section \ref{sec:rsp} details the protocol for this systematic review. Section \ref{sec:res} presents and discusses the results. Section \ref{sec:datasets} lists and describes the available real-world datasets. Section \ref{sec:on-drift} discusses concept drift visualization and drift simulation settings. Section \ref{sec:conclusion} concludes the study and emphasizes {open challenges and} future directions. \review{To facilitate the reader to follow the acronyms, we added Section \ref{appendix:acronyms} to list and explain the acronyms present in this manuscript, functioning as a glossary}.

\section{Background}
\label{sec:dsm}

According to \citet{bifet2018machine}, ``data streams are an algorithmic abstraction to support real-time analytics''. Data streams are data items arriving continuously and are temporally ordered. In traditional data mining, it is compulsory to have data collection so that the ML model can learn patterns from it and perform the desired task. However, there are several constraints in Data Stream Mining (DSM). {Because the data arrives continuously {and streams are potentially infinite}, storing the data {to posteriorly learn from} can become unfeasible{.}}

Thus, the ML model must learn from the data and discard it within a short period \citep{bifet2018machine}. In addition, \citet{bifet2018machine} mentioned that there are two main challenges for ML models when handling data streams: (a) learning from the data the instant it arrives and (b) being able to adapt in case the data evolves. {Since} these challenges must be addressed quickly and consume minimal processing, the outcome is an approximate model rather than a precise model.
{Furthermore, the same authors highlighted that since} data streams are continuously arriving rapidly and can be infinite, the data generation process may undergo significant changes over time, reflecting the data distribution. These changes, namely \textit{concept drift}, increase the challenges of managing data and text streams.

{In this paper, we define a text $\boldsymbol{X}$ as a sequence of arbitrary length composed of tokens $\boldsymbol{t}$. These tokens can include words (lexical units), punctuation marks, subwords, and other elements. Thus, we represent a text as $\boldsymbol{X} = (\langle \boldsymbol{t}_i\rangle~|~i = 1, ..., n)$, where $n$ denotes the total number of tokens. Typically, these tokens are organized in a specific order that adheres to the rules of natural language, allowing them to convey meaningful information. Initially, texts were primarily used for communication between humans. More recently, they have also served as logs for communication from systems to humans. Furthermore, in the last developments, text facilitates interactions from humans to systems, exemplified by chatbots and large language models like ChatGPT. }

%{Concept drift in text streams is formally described as follows. 
%Let a text stream $T = \{X_1, X_2, X_3, ...\}$ be a potentially infinite input text $X_i$ sequence, where $i$ is the text index.

{Concept drift in text streams can be formally defined as follows. 
Let a text stream $\boldsymbol{T} = (\langle \boldsymbol{X}_j\rangle~|~j = 1, ...)$ represent a potentially infinite sequence of input texts $\boldsymbol{X}_j$, where $j$ denotes the text index.}
{In the context of a classification task involving textual data streams, each text may be associated with a label $y$, resulting in a sequence of pairs $(\boldsymbol{X}, y)$, or more formally, $\boldsymbol{T} = (\langle \boldsymbol{X}_j,~ y_j\rangle~|~j = 1, ...)$. According to \citet{gama2014survey} concept drift is said to occur if}

\begin{equation}
    \exists{\boldsymbol{X}}: p_{t_0}(\boldsymbol{X},y) \neq p_{t_1}(\boldsymbol{X},y),
\end{equation}

\noindent {where $p_{t_0} (\boldsymbol{X},y)$ represents the joint distribution of $\boldsymbol{X}$ and the label $y$ at time $t_0$.} {It is important to note that $\boldsymbol{X}$ can be represented numerically as a dense vector or through word frequencies and co-occurrences over time. Such numerical representations facilitate the extraction of statistics that are essential for detecting concept drift and semantic shifts.}

%{\review{Concept drift in text streams is formally described as follows. Let a text stream $T = (\langle X_i\rangle~|~i = 1, ...)$ be a potentially infinite input text $X_i$ sequence, where $i$ is the text index.} Assuming a textual data stream in a classification task, each text may be accompanied by its label $y$, thus becoming a sequence of pairs $(X, y)$\review{, \ie, $T = (\langle X_i,~ y_i\rangle~|~i = 1, ...)$}. Therefore, according \citet{gama2014survey}, a concept drift is deemed to have occurred if

%\begin{equation}
%    \exists{X}: p_{t_0}(X,y) \neq p_{t_1}(X,y),
%\end{equation}

%\noindent in which $p_{t_0} (X,y)$ is the joint distribution between $X$ and the label $y$ in a time $t_0$.} \review{We highlight that $X$ is a numerical representation, a dense vector, or even word frequencies or word co-occurrences over time. Thus, such numerical representations allow for the extraction of statistics that enable concept drift/semantic shift detection.}

\begin{figure}[!htp]
\includegraphics[width=0.9\textwidth]{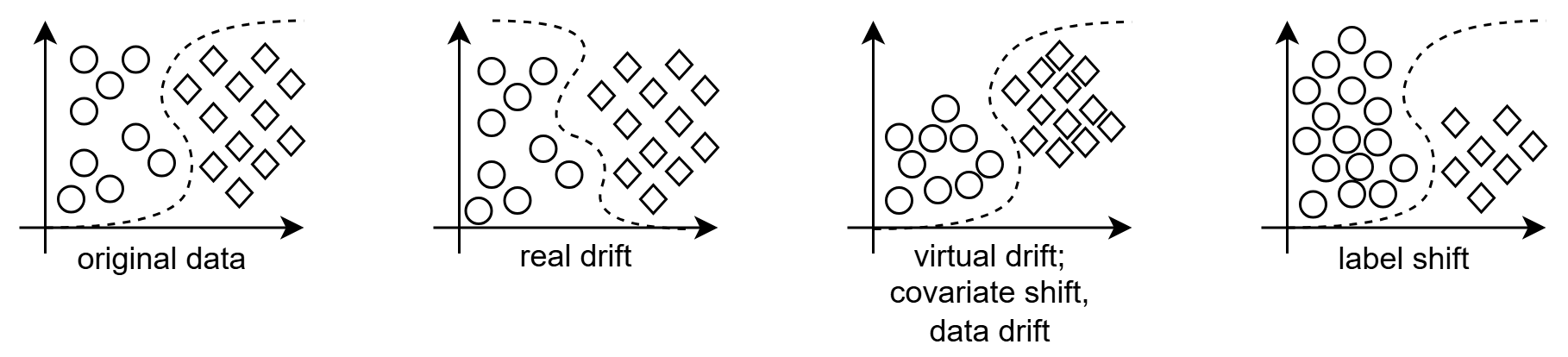}
\centering

\caption{{Types of concept drift. Adapted from \citet{gama2014survey}. Each marker, \ie, circle, and diamond, represents an arbitrary class/label. Dashed lines correspond to the border between regions of classes.}}
\label{fig:drift-types}
\end{figure}

According to Gama et al.~\cite{gama2014survey}, ``data is expected to evolve''. Thus, the data distribution can change as time passes. These changes are referred to as \textit{concept drift}. 
%Formally, a drift happens when $p_{t_{i-1}}(y|X) \neq p_{t_{i}}(y|X) $, meaning that the data distribution in the time $t_{i_{i-1}}$ is different from the data distribution in $t_{i_{i}}$. 
{The machine learning literature highlights} two primary types of drifts in data distribution: (i) \textit{Real concept drift}, where the relationship between $\boldsymbol{X}$ (input data) and $y$ (class) changes, and (ii) \textit{{Virtual} concept drift}, where the data distribution in $\boldsymbol{X}$ changes, but $p(y|\boldsymbol{X})$ does not change, meaning that the boundaries are unchanged. Real concept drift can occur even if the data distribution in $\boldsymbol{X}$ does not change. {Across scientific and industry communities, virtual drifts may also be referred to as \textit{covariate shift} \citep{feng2024towards}, or \textit{data drift} \cite{p47}. Another type of drift is the \textit{label shift}, which corresponds to changes in label distribution, compared to reference data \citep{zhao2021active}. Fig.~\ref{fig:drift-types} shows the aforementioned types of drifts.}

\begin{figure}[!htp]
\includegraphics[width=0.8\textwidth]{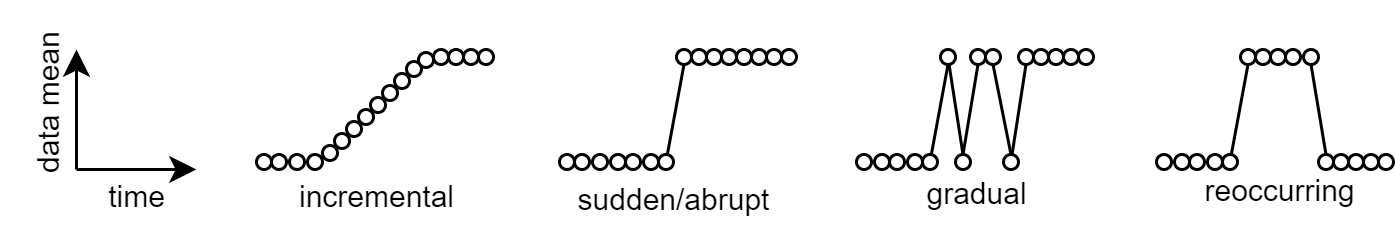}
\centering

\caption{{Dynamics of concept drift over time. Adapted from \citet{gama2014survey}.}}
\label{fig:cd-over-time}
\end{figure}

%Also, concerning the dynamics of concept drift over time, 
In addition, \citet{gama2014survey} highlighted four different types of concept drift dynamics over time. The four categories are as follows: (a) \textit{abrupt}, where the data distribution changes from $t_i$ to $t_{i+1}$; (b) \textit{incremental}, where the data distribution changes from $t_i$ to $t_{i+\Delta}$, where $\Delta > 1$; (c) \textit{gradual}, where the data distribution switches between different means until remaining in the last distribution; and lastly (d) \textit{reoccurring}, where the data distribution changes and later, switches back to the first data distribution observed. {Fig.~\ref{fig:cd-over-time} depicts the concept drift types concerning the dynamics over time.}

\begin{figure}[!htp]
\includegraphics[width=\textwidth]{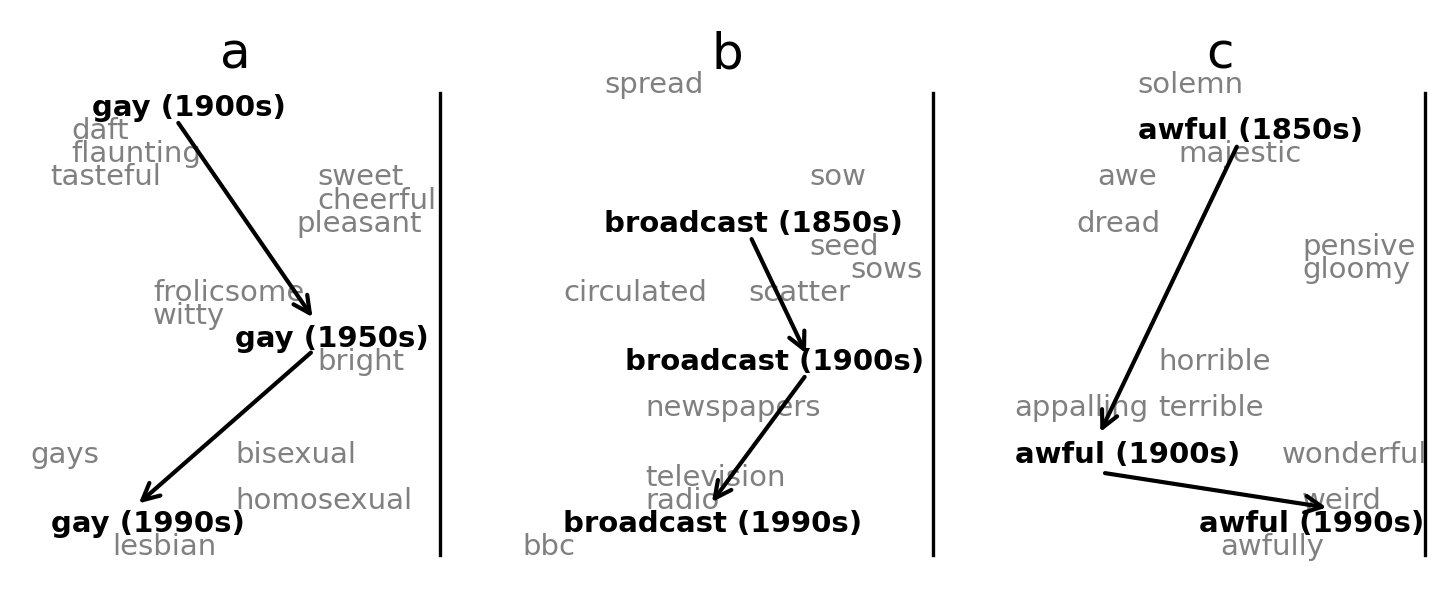}
\centering

\caption{Semantic shift across several decades or centuries. Adapted from \cite{hamilton2016diachronic}.}
\label{fig:semantic-shift}
\end{figure}

When it comes to text, different aspects of drifts may emerge, such as a word gaining or losing meanings over time, known as \textit{semantic shift} {\cite{bloomberg33}, sometimes referred to as \textit{semantic change} \cite{de2024survey}. In this paper, we use only the terminology \textit{semantic shift} for the sake of simplicity}. %In the next section, we approach semantic shift.
%\section{Semantic shift}
%\label{sec:semantic-shift}
According to \citet{kutuzov2018diachronic}, \textit{semantic shift} constitutes ``the evolution of word meaning over time''. Fig.~\ref{fig:semantic-shift} depicts examples of semantic shifts that occurred across decades and centuries \citep{hamilton2016diachronic}. %According to the authors in \cite{hamilton2016diachronic}, 
Fig.~\ref{fig:semantic-shift} was generated using Word2Vec representations \citep{mikolov2013distributed} and t-SNE \citep{tsne} for dimensionality reduction, according to \citet{hamilton2016diachronic}. In the 1850s, \textit{awful} had a positive connotation, as depicted in Fig.~\ref{fig:semantic-shift} (c). The surrounding words, \eg, \textit{majestic} and \textit{solemn}, corroborated the previous statement. However, in the 1900s, the word \textit{awful} shifted to a negative connotation due to its proximity to the words \textit{terrible} and \textit{horrible}. %According to the authors, the visualization was generated using Word2Vec \citep{mikolov2013distributed} and t-SNE \citep{tsne} for dimensionality reduction.  
{More precisely, \textit{semantic shift} has been studied across the years. \citet{de2024survey} overviewed the subject and characterized semantic changes considering the aspects of \textit{dimension}, \textit{relation}, and \textit{orientation}. In the case of dimension, \citet{de2024survey} considered broadening, \ie, gaining new meanings, and narrowing, \ie, becoming more specific or losing previous meanings. Considering the relation, \citet{de2024survey} mentioned metaphorization and metonymization, which occurs, according to the authors, ``when a word takes on a new meaning that, to some extent, inherits qualities from its original meaning through a figurative relationship the speaker aims to convey''. Finally, changes in orientation regard the connotation of a new meaning, \ie, towards positive (amelioration) or negative (pejoration). }

%In order to measure the evolution of a word meaning over time, 
Several works have been proposed to measure the evolution of a word's meaning over time \citep{di2019training,belotti2020unimib,ryzhova2021detection}. Some papers provide semantic shift detection methods that measure the cosine distance between word embeddings in a period and the word embeddings from the same words in a previous period \citep{p22}. If the distance exceeds a certain threshold, it is deemed a semantic shift to have occurred. Other approaches may use embedding alignment across time slices, such as orthogonal Procrustes \citep{hamilton2016cultural} and compass alignment \citep{belotti2020unimib}. However, traditional ML methods mostly address semantic shift detection, \ie, outside of the streaming context. It means that for most of the approaches, there are no constraints on processing and storage.

%In a world with voluminous data being generated each second, 
Approaches capable of handling text streaming become relevant in a world where enormous quantities of data are generated each second. Therefore, this review focuses exclusively on approaches applied to text stream scenarios. In addition, despite works that depict semantic shifts over long periods, works such as \citet{stewart2017measuring,p41} demonstrated that semantic shifts may occur not only in decades or centuries but also in a shorter period, \eg, weeks {or even a few minutes/hours}.

%Representation shift

%\section{Concept Drift Detectors}
%\label{sec:cdd}
Concept drift detectors are methods used for detecting changes in data distribution, and they can be beneficial in performing
both concept drift and semantic shift detection. 
These types of detectors were initially developed in statistics. However, there is no guarantee that such methods would work specifically in streaming scenarios because some may not work in a one-pass fashion \citep{bifet2018machine}. \citet{gama2014survey} categorized concept drift detection methods into four classes: (i) \textit{sequential analysis}; (ii) \textit{control charts}; (iii) \textit{monitoring two distributions}; and (iv) \textit{context-based methods}, which are also called \textit{heuristic methods}. \textit{Sequential analysis} corresponds to a scenario in which two subsets of data are generated sequentially by processes bound to different unknown distributions, \eg, $P_0$ and $P_1$. According to \citet{gama2014survey}, ``when the underlying distribution changes from $P_0$ to $P_1$ at point \textit{w}, the probability of observing certain subsequences under $P_1$ is expected to be \textit{significantly} higher than that under $P_0$''. It signifies that a statistical test, for example, can be used to detect this change. Two primary representatives of this category are the cumulative sum (CUSUM) test \citep{page1954continuous} and the Page-Hinkley test \citep{page1954continuous}, which is a variant of the CUSUM test \citep{gama2014survey,bifet2018machine}. 

The second category proposed by \citet{gama2014survey} is \textit{control charts}, also known as \textit{statistical process control} (SPC). Control charts correspond to ``standard statistical techniques to monitor and control the quality of a product during continuous manufacturing'' \citep{gama2014survey}. 
%In this case, a sequence of pairs $(X_i, y_i)$ 
In this case, the data are received over time and are input to the model, and the model's error is used to determine the states of the system. The system states are as follows: (i) \textit{in-control}, which indicates that the system is stable; (ii) \textit{drift detection}, which signifies the error increased significantly, compared to the historical error; and (iii) \textit{warning}, which indicates the error increased but was insufficient to raise a detection. Drift and warning are generally associated with a statistical confidence of 99\% and 95\%, respectively. An example of this category is the exponentially weighted moving average (EWMA) \citep{ross2012exponentially}.
The third category regards \textit{monitoring two distributions}. 
Methods in this category, according to \citet{gama2014survey}, ``typically use a fixed reference window that summarizes the past information and a sliding detection window over the most recent examples''. In this scenario, a drift is considered to have occurred if the distributions of the windows are statistically different. An example of a method that embeds a concept drift detector from this category is the Very Fast Decision Tree (VFDT) \citep{gama2006decision}. An actual concept drift detector that fits in this category is the Adaptive Windowing (ADWIN) \citep{bifet2007learning}. ADWIN is a distribution-free concept drift detector suited for detecting drifts in real-valued or bits streams \citep{bifet2018machine}. It maintains a window with the most recent items, from which subwindows are compared. If these subwindows exhibit different means above a threshold based on Hoeffding's bounds, a drift is flagged \citep{gama2014survey}. 
ADWIN is computationally more expensive in time and memory than \textit{sequential analysis} detectors; however, it is simpler to use because the user does not need to specify a cutoff parameter \citep{gama2014survey,bifet2018machine}. In addition, ADWIN provides more precise change points \citep{gama2014survey}. 

%Two examples of approaches in this category are the Very Fast Decision Tree (VFDT) \citep{gama2006decision} and Adaptive Windowing (ADWIN) \citep{bifet2007learning}. 

The last category, \ie, \textit{context-based}, regards specific approaches that use characteristics intrinsic to ML methods to perform drift detection or adaptation. For instance, \citet{leite2012evolving,soares2019evolving,garcia2019incremental} proposed a method that balances incremental learning and forgetting using fuzzy granular computation. Whenever a new instance is inputted, the existing granules, \ie, groups that share similar properties, have their (either complete or partial, whenever there are missing attributes in the new instance) similarity with the newly seen instance calculated. 
The new instance is assigned to the chosen granule if the similarity exceeds a certain threshold. However, if no granule can match the newly seen instance, \ie, a drift occurs, and a new granule is created to accommodate the new instance. In addition, a pre-defined parameter controls the periods of verifying stale granules, which can be deleted to maintain the model's conciseness.

%According to \cite{MOA-Book-2018},
The common metrics used to evaluate and compare concept drift detection methods, according to \citet{bifet2018machine}, are as follows: (i) mean time between false alarms (\textit{MTFA}), which assesses the frequency with which a method raises false alarms; (ii) false alarms rate (\textit{FAR}), which is the inverse of \textit{MTFA};
%\akst{calculated as $FAR = \frac{1}{MTFA}$};
(iii) mean time to detection (\textit{MTD}), which assesses how quickly the method detects and responds to drift once it occurs; (iv) missing detection rate (\textit{MDR}), which determines how frequently the method fails to warn when drift occurs; and (v) average run length (\textit{ARL}), which is the time it takes to raise the alarm once a drift occurs \citep{bifet2018machine}. \textit{ARL} integrates \textit{MTD} and \textit{MTFA} \citep{bifet2018machine}. Additional metrics, such as Mean Time Rate (\textit{MTR}) \citep{wares2019data,bifet2017classifier}, may emerge in the literature; however, the{} primary focus is on missing drifts, hits, time/iterations until detecting an actual drift, or a combination of such factors. \textit{MTR}, for instance, is analogous to \textit{ARL} \citep{wares2019data}.

Typically, concept drift detectors are coupled to traditional or online ML systems by receiving the hits and errors of prediction. These concept drift detectors have two levels of alarms: \textit{warning} and \textit{drift}. The most straightforward use is when a warning alarm is issued. Either the input data are buffered, or a new ML model is trained such that when the drift alert occurs, a new model (trained using data from the buffer) replaces the outdated one.
This learning strategy is called \textit{background learning} \citep{gomes2017adaptive}. Thus, the idea is to maintain an updated model based on the most recent/frequent data.

\section{Systematic Review Protocol}
\label{sec:rsp}

This review followed the guideline proposed by \citet{kitchenham2007guidelines}, which comprises three steps: (i) planning the review, (ii) conducting the review, and (iii) reporting the review. \textit{Planning the review} includes identifying the need for the review and formulating the research questions. In \textit{conducting the review}, we select primary studies and perform data extraction and synthesis. Finally, in \textit{reporting the review}, it is expected to disclose the results and findings.
In this work, we used %Google Scholar as primary source of studies. %This choice took into consideration that, although Google Scholar can return much more noise, it democratizes the search, returning papers from several venues, that might have not been indexed by the most popular paper indexers. As a side effect, we could obtain a more thorough view from the area. 
five sources of studies: IEEEXplore\footnote{https://ieeexplore.ieee.org/}, Science Direct\footnote{https://www.sciencedirect.com/}, ACM Digital Library\footnote{https://dl.acm.org/}, Springer Link\footnote{https://link.springer.com/}, and Scopus\footnote{https://www.scopus.com/}. We devised a series of four questions to guide our research. The primary question, $RQ1$, takes precedence, while the remaining questions are derived from $RQ1$. Table \ref{tab:rqs} displays our research questions for reference.

%To guide our research, we developed a set of four questions. The primary question, $RQ1$, takes precedence while the remaining questions are derived from $RQ1$. Table \ref{tab:rqs} displays our research questions for reference.

%%Backup
%five sources of studies: IEEEXplore\footnote{https://ieeexplore.ieee.org/}, Science Direct\footnote{https://www.sciencedirect.com/}, ACM Digital Library\footnote{https://dl.acm.org/}, Springer Link\footnote{https://link.springer.com/}, and Scopus\footnote{https://www.scopus.com/}. We developed four research questions to guide the information extraction from the papers. The research questions are shown in Table \ref{tab:rqs}. The first question, namely $RQ1$, is the main while the other questions are ramifications from $RQ1$. 

\begin{table}[!htp]
\caption{Research questions used in this work.}
\label{tab:rqs}
%\resizebox{\textwidth}{!}{
\begin{tabular}{ll}
\hline
\textbf{ID} & \textbf{Research Questions}                                                                       \\ \hline
RQ1         & ``How to handle concept drift using ML approaches having as source  \\
            &    text streams?''\\
RQ2         & ``Which type of application is addressed?''                                                     \\
RQ3         & ``Which type of token/word/sentence representation is used in the study?''                \\
RQ4         & ``Which datasets were used to evaluate the proposed approach(es)?''                             \\\hline         
\end{tabular}
%}
\end{table}

% \begin{enumerate}
%     \item Research Question 1 ($RQ1$): ``How to handle concept drift using machine learning approaches having as source text streams?'' 
%     \item Research Question 2 ($RQ2$): ``Does the concept drift detection happen implicitly or explicitly?''
%     \item Research Question 3 ($RQ3$): ``Which type of application is approached?''
%     \item Research Question 4 ($RQ4$): ``Which type of token/word/sentence representation is used in this application?''
%     \item Research Question 5 ($RQ5$): ``Which datasets were used to evaluate the approach(es)?''
% \end{enumerate}

The search query was developed considering $RQ1$. We also used a few synonyms to aid in developing a broadening query. 
The reader can discover additional information on the terms and synonyms in Table \ref{tab:synonyms}.
%Regarding the terms and their respective synonyms, the reader can resort to Table \ref{tab:synonyms}. %Besides, all the research questions were unraveled in order to extract more fine-grained information from the studied papers.
$RQ2$ focuses on the applications the papers addressed when handling concept drift in textual streams. This question is crucial because it can illustrate various scenarios, the potential, and increased interest in specific problems. Besides the application, we wanted to know which ML methods are employed and how these models are updated, \eg, incrementally or regularly retrained. %, and the programming languages, or frameworks used in the selected works. 
With $RQ3$, we intended to uncover the most common approaches to representing texts (or smaller parts, such as tokens, words, and sentences). Finally, $RQ4$ pursues insights into the existence of consolidated datasets for the field and their aspects, such as the level of labeling in the dataset, \eg, instance or token, the data mining task employed, \eg, clustering, classification, whether the dataset contains real-world data or it is synthesized, metrics used in those data mining tasks, and whether drifts are labeled in the dataset.

%-- Falar sobre as subquestões

\begin{table}[!htp]
\caption{Table containing keywords and respective synonyms.}
\label{tab:synonyms}
%\resizebox{\textwidth}{!}{
\begin{tabular}{ll}
\hline
\textbf{Keyword} & \textbf{Synonyms}\\ \hline
concept drift    & semantic shift, representation shift, semantic change \\ %\hline
machine learning & -\\ %\hline
text streams & textual streams, social network streams, Twitter streams, diachronic, \\ 
 & text streaming\\ %\hline
detection        & -\\ \hline
\end{tabular}
%}

\end{table}

We developed the query presented below using Table \ref{tab:synonyms}. The terminologies \textit{semantic shift} and \textit{representation drift} are closely related to \textit{concept drift}, especially in the textual context. Semantic shift (or semantic change), according to \citet{bloomberg33}, refers to ``innovations which change the lexical meaning rather than the grammatical function of a form''. However, according to \citet{fu2022adapterbias}, the representation shift in NLP relates to changes in the vector representation{, when using semantic vectors as representations for word meaning}. We included \textit{social network streams} because they are the notable source of text streams produced directly by humans nowadays. We also used the terminology \textit{Twitter streams}, because Twitter{, \eg, currently named $X$\footnote{\url{https://x.com/}},} is a microblog (one of the most popular) and generated around 500 million tweets (posts) per day, in 2022\footnote{https://www.dsayce.com/social-media/tweets-day/}. Furthermore, we included the term \textit{diachronic}. When serving as an adjective for a dataset, \textit{diachronic} refers to a dataset that contains data produced over time. The term \textit{machine learning} was withdrawn because concept drift is mostly addressed by or in processes that use ML techniques.
The query used in the search is:
\texttt{(``concept drift'' OR ``semantic shift'' OR ``representation shift'' OR ``semantic change'') AND (``text streams'' OR ``textual streams'' OR ``textual streaming'' OR ``social network streams'' OR ``twitter streams'' OR ``diachronic'') AND (``detection'')}. Each source has its parameters, but we prioritized full-text search in all of them.

\subsection{Inclusion and Exclusion Criteria}
\label{subsec:inc-exc-criteria}

The inclusion and exclusion criteria used in this review are described below. It is crucial to note that we limited this review to papers published after 2018 because other previous secondary studies tackle similar problems {\citep{kutuzov2018diachronic,tahmasebia2021survey,patil2021concept,montanelli2023survey}}. {\citet{kutuzov2018diachronic} evaluated several papers regarding diachronic word embeddings and semantic shifts. The authors approached several aspects, such as diachronic semantic relations and the sources of diachronic data for training and testing. \citet{tahmasebia2021survey} developed a survey on computational approaches for lexical semantic change detection. They approached aspects such as the semantic change types and computational modeling of diachronic semantics. \citet{patil2021concept} also developed a survey on concept drift detection for social media. The authors provided information on datasets and the evolution of techniques over time. \citet{montanelli2023survey} %presented a survey on contextualized semantic shift detection 
{presented a
survey on modeling lexical semantic change through modern, deep language models, including large language models,}
regarding aspects such as time awareness, learning scheme, language model, training language, and corpus language.}
{The surveys/reviews from  \citet{kutuzov2018diachronic,tahmasebia2021survey,montanelli2023survey} evaluated semantic shift and diachronic aspects without concerning specifically streams and methods that respect the streaming processing constraints. \citet{patil2021concept} approached a similar aspect as ours; however, we provided deeper analysis on several characteristics, such as model update scheme, text representation methods, and their update schemes when available, datasets, and so on.}
{Thus,} {a substantial difference between our systematic review and the aforementioned works is that we focus on papers that approach the problem of concept drift{/semantic shift} using text streams as a data source. {Using streams as data sources requires specific approaches to overcome the stream processing constraints, as seen in Section \ref{sec:dsm}}}. Therefore, according to Table \ref{tab:ic-and-ec}, we considered the following inclusion and exclusion criteria. It is also essential to note that this review protocol was {last} executed {on August 20, 2024.}

\begin{table}[!htp] 
\caption{Inclusion and Exclusion criteria used in this study.} 
\label{tab:ic-and-ec} 
\resizebox{\linewidth}{!}{ 
\begin{tabular}{lllll} 
\cline{1-2} \cline{4-5} 
\multicolumn{1}{l}{\textbf{Ref}} & \multicolumn{1}{l}{\textbf{Inclusion criteria}} & \textbf{} & \textbf{Ref} & \textbf{Exclusion criteria} \\ 
\cline{1-2} \cline{4-5} 
\multicolumn{1}{l}{IC1} & \multicolumn{1}{l}{The study is published in journals or } & & EC1 & The study is not primary \\ 
\multicolumn{1}{l}{} & \multicolumn{1}{l}{conference proceedings} & & EC2 & The study is not written in English \\ 
\multicolumn{1}{l}{IC2} & \multicolumn{1}{l}{The study is published from 2018 (inclusive)} & & EC3 & The study is incomplete \\ 
%\multicolumn{1}{l}{IC3} & \multicolumn{1}{l}{The study must be accessible in electronic way} & & EC3 & The study is incomplete or not accessible \\
\multicolumn{1}{l}{IC3} & \multicolumn{1}{l}{The study presents a method for handling} & & EC4 & The study is not an article \\ 
\multicolumn{1}{l}{} & \multicolumn{1}{l}{concept drift} & & EC5 & The study is duplicated \\ 
\multicolumn{1}{l}{IC4} & \multicolumn{1}{l}{The study uses text streams as data source} & & EC6 & The study does not meet  \\ 
\cline{1-2} 
& & & & the inclusion criteria \\ \cline{4-5} \end{tabular} 
} 
\end{table}

After gathering the returned papers, each researcher screened their abstracts to flag the inclusion or exclusion of each study. Concerning divergences, the researchers agreed to read the divergent papers carefully to have confidence in their decision. We used Cohen's Kappa coefficient \citep{mchugh2012interrater} to measure the agreement level between the researchers.

\section{Results and Discussion}
\label{sec:res}

\begin{figure}[!htp]
\includegraphics[width=0.85\textwidth]{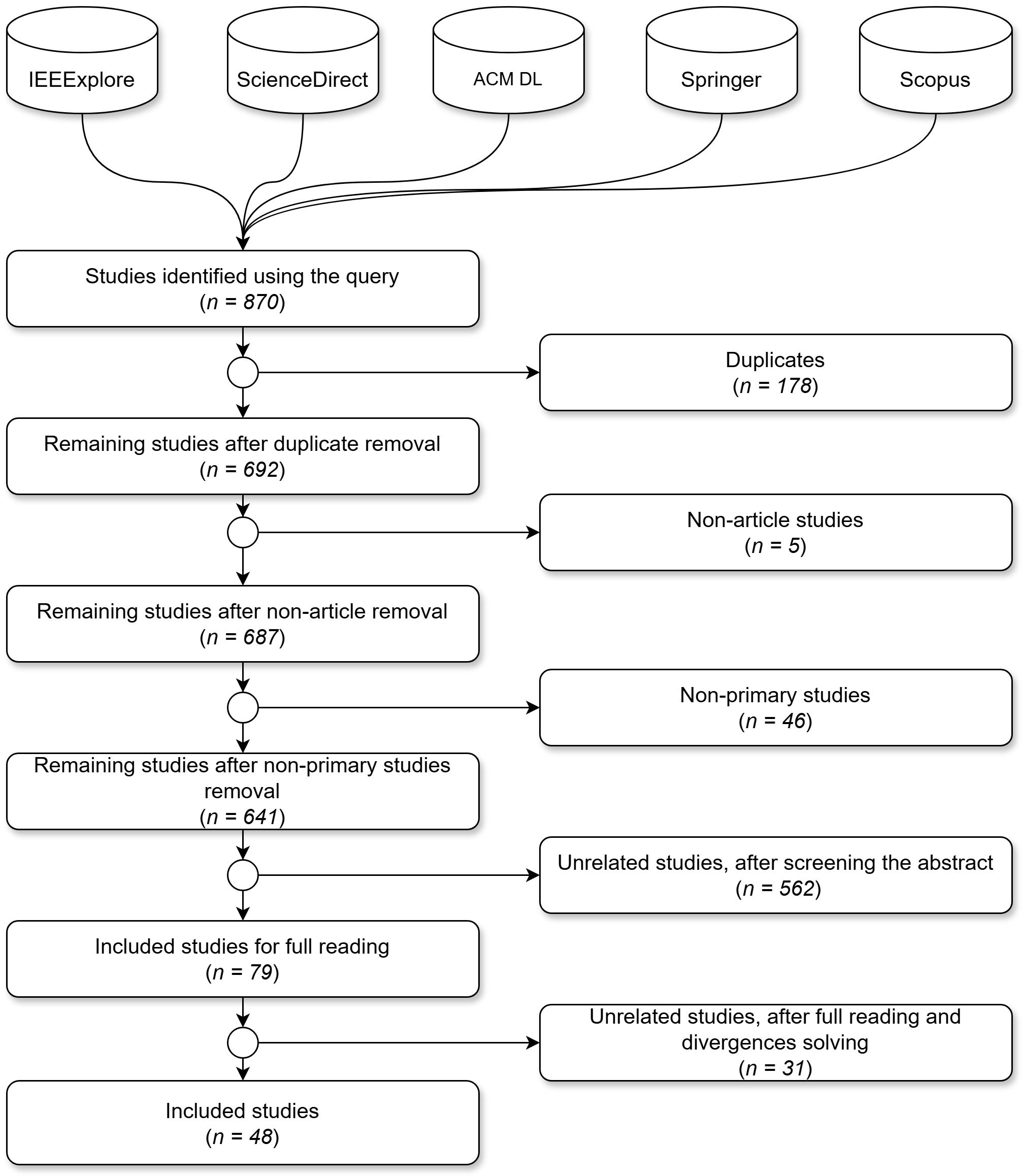}
\centering

\caption{{Process of papers selection. Each rounded-corner rectangle on the right side corresponds to an exclusion criterion. The numbers of remaining studies after each elimination are presented on the left side.}}
\label{fig:filter}
\end{figure}

Fig.~\ref{fig:filter} overviews the paper selection process. We collected 
%494 papers, 
%{662 papers,}
{870 papers,}
considering the research query. The final calculated Cohen's Kappa coefficient reached 
%94.8, 
%{87.25,}
{84.61\%,}
which indicates a high level of agreement between the researchers. In addition, the divergences were discussed after a thorough reading of the divergent papers, and a decision was reached on their inclusion or exclusion. 
%After removing duplicates (n=49), non-article studies (n=4), non-primary studies (n=35), and unrelated studies (n=371), we retained 35 articles for a full reading and analysis. 
%{After removing duplicates (n=132), non-article studies (n=4), non-primary studies (n=38), and unrelated studies (n=448), we retained 40 articles for a full reading and analysis. }
{After removing duplicates (n=178), non-article studies (n=5), non-primary studies (n=46), and unrelated studies (n=562+31=593), we retained 48 articles for a full reading and analysis. }

Considering the process depicted in Fig.~\ref{fig:filter}, the reader's attention may be drawn by the high number of unrelated studies after screening the abstract. It occurred due to the query term \textit{diachronic}, which relates to something that evolves, especially concerning language. Most approaches that handle language evolution cannot work in streaming environments (about 60\% of the papers {in our initial identification using the query}). 
Therefore, we excluded those studies from our paper selection. {In addition, we highlight that we are interested in approaches that handle \emph{text streams} as data sources. It means that to be considered for our selection, the approaches must process the datasets seeking to respect the text stream constraints (see Section \ref{sec:dsm}). This characteristic filtered out several papers from our selection. Furthermore, around 16\% of the papers did not handle/mention drift, although the terminology was included in the keywords, as shown in Table \ref{tab:synonyms}.}
%Table \ref{tab:selected-papers-characteristics} shows the selected papers ordered by year, categorized in each aspect selected to study: (C) categories of text drift; (D) type of detection; (MU) model update; (TR) text representation; and (TRUS) text representation update scheme.

\begin{figure}[!htp]
\includegraphics[width=.9\textwidth]{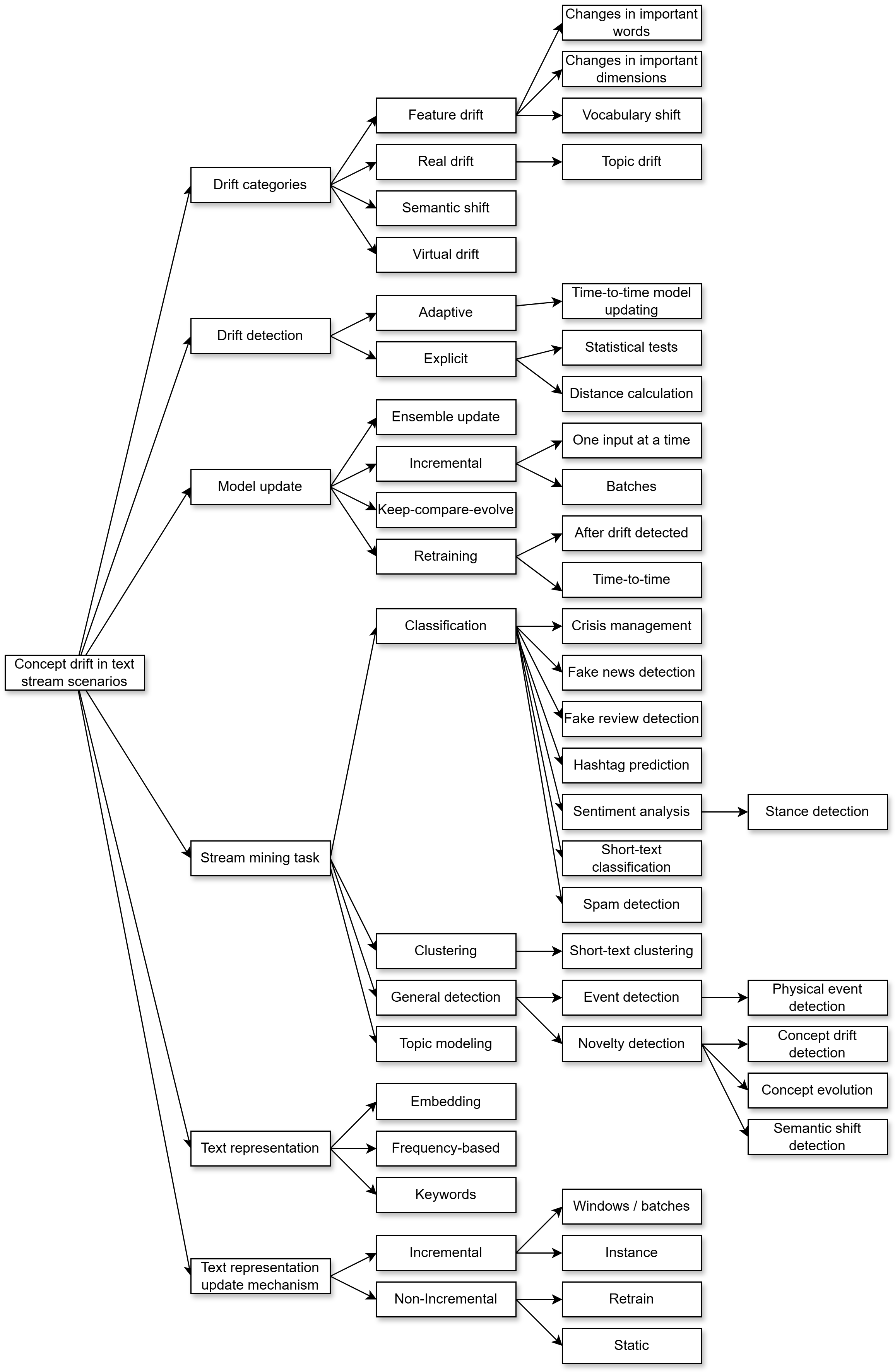}
\centering
%\caption{A taxonomy regarding \akst{concept drift detection in text stream settings} \ak{concept drift in text stream scenarios}.}
\caption{{A taxonomy regarding concept drift in text stream scenarios.}}
\label{fig:taxonomy}
\end{figure}

Based on the information extracted from the selected papers using the research questions, we categorized the approaches for handling text drifts presented according to the following characteristics: (DC) text drift categories; (DD) drift detection types; (MU) model update; (TR) text representation; and (TRUS) text representation update scheme. Our proposed taxonomy is depicted in Fig.~\ref{fig:taxonomy}. In addition, Table \ref{tab:selected-papers-characteristics} shows the selected papers according to our proposed taxonomy. Subsection \ref{subsec:overall-statistics} describes the main statistics of the selected papers. %We unravel the overall statistics and each aspect of the papers in the next subsections.

The selected papers were studied in detail considering the taxonomy presented in Fig.~\ref{fig:taxonomy}. Section \ref{subsec:td-categories} describes and categorizes the types of concept drift handled in the selected papers, \ie, \textit{Drift categories}. Section \ref{subsec:td-detection} analyzes how the text-related concept drift detection is performed, \ie, in a model-adaptive way or explicitly, regarding the \textit{Drift detection} in our proposed taxonomy. 
%Regarding the taxonomy presented in Fig. \ref{fig:taxonomy}, in \textit{Text drift categories}, we categorize the types of concept drift handled in 
%the selected papers. Section \ref{subsec:td-categories} describes each category in detail. 
%Regarding \textit{Text drift detection}, we analyze how the text-related concept drift detection is performed, i.e., in a model-adaptive way or explicitly. Section \ref{subsec:td-detection} details the papers on each detection method. 
%Section \ref{subsec:td-model-update} 
Section \ref{subsec:td-model-update} describes how the ML models used in the papers are updated when handling a text stream, \ie, \textit{Model update} in the taxonomy. % Section \ref{subsec:td-model-update} details the model update schemes developed in the selected papers.
%In \textit{Model update}, we describe how the ML models used in the papers are updated when handling a text stream. Section \ref{subsec:td-model-update} details the model update schemes developed in the selected papers. 
We categorized the approaches according to the related \textit{Stream mining tasks}, in addition to the applications and related metrics. Section \ref{subsec:td-stream-mining-tasks} expands the information on the stream mining tasks presented in the papers. 
In \textit{Text representation}, we uncovered the text representation methods used in the papers, considering embeddings, frequency-based methods, and words directly. Section \ref{subsec:td-text-representation} describes 
%In Section \ref{subsec:td-text-representation} we describe 
the text representation methods used in the papers. 
For \textit{Text representation update mechanism}, we analyzed whether and how the text representations are updated over time. Section \ref{subsec:text-representation-upd-scheme} explores the update scheme of the text representation methods.
All selected methods were studied under the taxonomy's second level, \ie, text drift categories, text drift detection, model update, stream mining task, text representation, and text representation update mechanism. In addition, the methods can fit more than one characteristic below the second level. %For instance, a given method could be categorized into classification and concept evolution, in which the former is a subcategory of novelty detection.
{Recalling the Research Questions presented in Section \ref{subsec:inc-exc-criteria}, RQ1, \ie, \textit{``How to handle concept drift using ML approaches having as source text streams?''}, is addressed in Sections \ref{subsec:td-categories}, \ref{subsec:td-detection}, and \ref{subsec:td-model-update}; RQ2, \ie, \textit{``Which type of application is addressed?''} is addressed in Section \ref{subsec:td-stream-mining-tasks}; RQ3, \ie, \textit{``Which type of token/word/sentence representation is used in the study?''} is approached in Section \ref{subsec:td-text-representation}; and finally, RQ4, \ie, \textit{``Which datasets were used to evaluate the proposed approach(es)?''}, is conveyed in Section \ref{sec:datasets}}.

% We started having as a return from Google Scholar with 417 works. All these papers were screened in order to determine if they should be included or excluded in this study. We removed duplicated papers, remaining 409 papers. After removing non-article papers, papers not written in English, papers outside the scope, non-primary studies, unavailable studies and unpublished studies, it remained 69 papers.

%\begin{landscape}
\begin{table}[!htp]
\rowcolors{2}{gray!25}{white}
%\centering
\caption{Selected papers ordered by year. {Acronyms are explained in the legend.}}
\resizebox{.93\linewidth}{!}{
\begin{tabular}{lllllll}
\hline
\textbf{Method} & \textbf{(DC)} & \textbf{(DD)} & \textbf{(MU)} & \textbf{(SMT)} & \textbf{(TR)} & \textbf{(TRUS)}  \\ \hline
AWILDA \citep{p1} & r $>$ td & e $>$ st & i $>$ o & tm & {kw} & n $>$ s\\ 
OBAL \citep{p2} & r & a & i $>$ b & class $>$ cm & fb & n $>$ s \\ 
CRQA \citep{p-25} & r & e $>$ st & - & class $>$ sa, nd $>$ cdd & - & n $>$ s\\ 
AIS-Clus \citep{p3}  & r, fd $>$ ciw & a & i $>$ b, i $>$ o & clust, class, gd $>$ ed, nd $>$ ce & {kw} & n $>$ s \\ 
- \citep{p4} & r $>$ td & e $>$ dc & i $>$ b & class $>$ stclass & fb & n $>$ s \\ 
- \citep{p5} & fd $>$ ciw & a, e $>$ st¹ & i $>$ b & class $>$ s, class $>$ sa & fb & n $>$ s \\ 
MStream \citep{p6} & r $>$ td & a & i $>$ b & clust $>$ stclust & fb & n $>$ s \\ 
OurE.Drift \citep{p7} & r $>$ td & e $>$ dc & eu & class $>$ stclass & fb & n $>$ s \\ 
- \citep{p8} & r $>$ td & e $>$ st & i $>$ o & class $>$ tc & {kw} & n $>$ s \\
- \citep{p9} & r & a & i $>$ b & class & e & n $>$ s \\ 
AIS-Clus \citep{p10} & r, fd $>$ ciw & a & i $>$ b, i $>$ o & clust, class, gd $>$ ed, nd $>$ ce & {kw}  & n $>$ s\\ 
LITMUS-ASSED \citep{p11} &  r & a & i $>$ b & gd $>$ ed $>$ ped & e & n $>$ s\\ 
LITMUS \citep{p12} & r & a & i $>$ b & gd $>$ ed $>$ ped & e & n $>$ s \\ 
DCFS \citep{p13} & fd $>$ cid & e $>$ st & r $>$ ad & class $>$ s, nd $>$ ce & fb & n $>$ s \\ 
LITMUS \citep{p14} & r, v & e $>$ dc & eu & gd $>$ ed $>$ ped & e & n $>$ s \\
ESACOD \citep{p15} & r & e $>$ st & r $>$ ad & class, nd $>$ ce & e & n $>$ s \\ 
- \citep{p16} & r & a & r $>$ t & class $>$ sa $>$ sd & fb, e & n $>$ r \\ 
- \citep{p17} & r & e $>$ st & i $>$ o & class $>$ frd & fb &  n $>$ r \\ 
- \citep{p18} & r & a & i $>$ o & class $>$ ht & fb, e & n $>$ s  \\ 
OFSER \citep{p19} & r & a & i $>$ o & class $>$ s & fb & n $>$ s \\ 
- \citep{p20} & r & a & i $>$ b & class $>$ sa $>$ sd & fb, e & n $>$ r \\ 
- \citep{p21} & r & e $>$ dc & eu & class $>$ stclass & fb & n $>$ s\\ 
- \citep{p22} & r, s, fd $>$ v & a & kce & class $>$ ht & e & i $>$ b   \\ 
EStream \citep{p23} & r $>$ td & a & i $>$ o & clust $>$ stclust & fb, e & n $>$ s  \\ 
EWNStream+ \citep{p24} & r $>$ td & a & i $>$ b & clust $>$ stclust  & fb & n $>$ s\\ 
GCTM \citep{p26} & r $>$ td & a & i $>$ b & tm & e & n $>$ s \\ 
BSP \citep{p27} & r $>$ td & a & i $>$ b & tm & e & n $>$ s \\ 
- \citep{p28} & r & e $>$ st & i $>$ b & class $>$ ht & e, fb & n $>$ s\\ 
DDAW \citep{p29} & r, v & e $>$ dc & eu & class $>$ sa & - & -\\ 
GOWSeqStream \citep{p30} & r $>$ td & a & i $>$ b & clust $>$ stclust & e & n $>$ s \\ 
GDWE \citep{p31} & r, s & a & i $>$ b & class & e & i $>$ b \\ 
- \citep{p32} & r & a & i $>$ o & class $>$ sa & fb & i $>$ inst   \\ 
- \citep{p33} & r & a & i $>$ b, r $>$ t & class $>$ stclass & fb & n $>$ r \\ 
SMAFED \citep{p34} & r & a & i $>$ b & class, clust, gd $>$ ed & e & n $>$ s \\ 
WIDID \citep{p35} & s & e $>$ dc & i $>$ b & nd $>$ ssd & e & n $>$ s\\ 
{- \citep{p36}} & r $>$ td & e $>$ dc & r $>$ td & class $>$ stclass & e & n $>$ s\\ 
{FFCA index \citep{p37}} & r & e $>$ dc & - & class $>$ fnd & fb & n $>$ s\\ 
{TSDA-BERT \citep{p38}} & r & e $>$ dc & r $>$ ad & class $>$ sa & e & n $>$ r\\ 
{DDAW \citep{p39}} & r, v & e $>$ dc & eu & class $>$ sa & f & -\\ 
{textClust \citep{p40}} & r & a & i $>$ b, i $>$ o & clust & fb & i $>$ b\\ 
{- \citep{p41}} & r & e & - & stclass & {kw}, fb & i $>$ inst\\ 
{- \citep{p42}} & r $>$ td & dc & i & hd & {kw} & n $>$ s\\ 
{OSMTS \citep{p43}} & r & dc & i & class $>$ ml & {kw}, e & n $>$ s\\ 
{TCR-M \citep{p44}} & r $>$ td & dc & r $>$ t & class & fb & n $>$ s\\ 
{- \citep{p45}} & r, fd $>$ ciw & dc & r $>$ t & clust & e & i $>$ inst\\ 
{- \citep{p46}} & r & a & i, r & class & fb & n $>$ r\\ 
{AE \citep{p47}} & v & e & - &  cdd & a & n $>$ s\\ 
{AdaNEN \citep{p48}} & r & a & i & class & e & n $>$ s\\ 
\hline
\end{tabular}
}

\label{tab:selected-papers-characteristics}
%\vspace{-3pt}

\begin{flushleft}
%\caption*{
{\scriptsize	
\textit{Legends}: $>$ : a level down in the taxonomy. \textbf{(DC) Drift category} $\rightarrow$ \textit{r}: real drift; \textit{td}: topic drift; \textit{fd}: feature drift; \textit{cid}: changes in important dimensions;  \textit{ciw}: changes in important words; \textit{v}: vocabulary shift; \textit{vd}: virtual drift; \textit{s}: semantic shift. 
\textbf{(DD) Drift detection method} $\rightarrow$ \textit{a}: adaptive; \textit{e}: explicit; \textit{st}: statistical tests; \textit{dc}: distance calculation. 
\textbf{(MU) Model update} $\rightarrow$ \textit{eu}: ensemble update; \textit{i}: incremental; \textit{o}: one input at a time; \textit{b}: batches; \textit{kce}: keep-compare-evolve; \textit{r}: retraining; \textit{ad}: after drift detected; \textit{t}: time-to-time. 
\textbf{(SMT) Stream mining task} $\rightarrow$ \textit{class}: classification; \textit{cm}: crisis management; {\textit{fnd}: fake news detection;} \textit{frd}: fake review detection; \textit{ht}: hashtag prediction; \textit{sa}: sentiment analysis; \textit{sd}: stance detection; \textit{s}: spam detection; \textit{stclass}: short-text classification; \textit{clust}: clustering; \textit{stclust}: short-text clustering; \textit{gd}: general detection; \textit{ed}: event detection; \textit{ped}: physical event detection; \textit{nd}: novelty detection; \textit{cdd}: concept drift detection; \textit{ce}: concept evolution; \textit{ssd}: semantic shift detection; \textit{tm}: topic modeling.
\textbf{(TR) Text representation} $\rightarrow$ \textit{e}: embedding; \textit{fb}: frequency-based; \textit{{kw}}: {key}words. \textbf{(TRUS) Text representation update scheme} $\rightarrow$ \textit{i}: incremental; \textit{b}: batch; \textit{inst}: instance; \textit{n}: none; \textit{r}: retrain; \textit{s}: static.\\
{\scriptsize ¹: one version uses ADWIN to explicitly detect feature drift.}

}
\end{flushleft}
\end{table}    

\subsection{Main Statistics}
\label{subsec:overall-statistics}

We unraveled statistics on the selected papers regarding (i) the sources, (ii) years of publication, and (iii) venues of publication. Table \ref{tab:number-selected-papers-source} shows the number of selected papers by source. Scopus provided {37.5\%} of the selected papers for this work. We noted {a steady interest across the} years in streaming text applications susceptible to concept drift in its various possibilities. Considering the limited time range in our search, \ie, between 2018 and %2022 (inclusive), 
%{September 2023,}
{August 2024,} we collected the respective number of papers: (2018) 10 papers; (2019) seven papers; (2020) two papers; (2021) six papers; 
%and (2022) 10 papers. 
{(2022) 11 papers; (2023) {eight} papers; }
{and (2024) four papers.}
Considering the characteristics of the papers across the years, we cannot infer a trend. We hypothesized that this behavior occurred because the research area is still incipient.

%Considering the characteristics of the papers across the years, the last two years show an increasing trend in relevant papers for the research topic of this paper.

\begin{table}[!htp]
\centering
%\caption{Number of selected papers \ak{according to the source.} \akst{, by source.}}
\caption{Number of selected papers according to the source.}
\begin{tabular}{ll}
\hline
Source & Selected papers \\\hline
ACM Digital Library	& {5}\\ %4
IEEE Xplore & {8}\\ %5\\ %6
Science Direct & {8}\\ %5\\ %6
Scopus & {18}\\ %12\\ %15
Springer Link & 9\\\hline
Total & {48}\\\hline %35 %40\\\hline

\hline
\end{tabular}
\label{tab:number-selected-papers-source}
\end{table}

Table \ref{tab:venues} shows the venues that contributed the most to our search. The journal Expert Systems with Applications published {four} papers, followed by IEEE \review{International} Conference on Evolving and Adaptive Intelligent Systems (EAIS) {with three papers}, ACM SIGKDD, Evolving Systems, International Joint Conference on Artificial Intelligence (IJCAI), International Joint Conference on Neural Networks (IJCNN), and Neurocomputing, each with two papers. 

\begin{table}[ht!]
\rowcolors{2}{gray!25}{white}
%\caption{Venues \ak{where} the selected papers \ak{were published.}}
\caption{Venues where the selected papers were published.}
\resizebox{\linewidth}{!}{ 
\begin{tabular}{lr}
\hline
\textbf{Venues}                                                                      & \multicolumn{1}{l}{\textbf{Appearances}} \\ \hline
Expert Systems with Applications                                                     & {4}                               \\
IEEE \review{International} Conference on Evolving and Adaptive Intelligent Systems (EAIS)                  & {3}                                        \\
ACM SIGKDD International Conference on Knowledge Discovery and Data Mining           & 2                                        \\
Evolving Systems                                                                     & 2                                        \\
International Joint Conference on Artificial Intelligence (IJCAI)                    & 2                                        \\
International Joint Conference on Neural Networks (IJCNN)                            & 2                                        \\
Neurocomputing                                                                       & 2                                        \\
ACM International Conference on Distributed and Event-Based Systems                  & 1                                        \\
ACM Symposium on Document Engineering                                                & 1                                        \\
{ACM Transactions on Knowledge Discovery from Data}	                         & {1}\\
\review{Annual Meeting of the Association for Computational Linguistics: Industry Track} & 1 \\
Applied Intelligence                                                                 & 1                                        \\
\review{Asian Conference on Intelligent Information and Database Systems}           & 1 \\
Brazilian Conference on Intelligent Systems (BRACIS)                                 & 1                                        \\
Chaos\review{: An Interdisciplinary Journal of Nonlinear Science}                                                                                & 1                                        \\
Cognitive Computation                                                                & 1                                        \\

{Computer Systems Science and Engineering}                                           & {1} \\ 
{Computers, Materials and Continua}                                                  & {1} \\ 
IEEE Access                                                                          & 1                                        \\
{IEEE Transactions on Big Data}                                                         & {1} \\ 
IEEE Transactions on Cybernetics                                                     & 1                                        \\
{IEEE Transactions on Systems, Man, and Cybernetics: Systems}	             & {1}\\
International Conference of Reliable Information and Communication Technology        & 1                                        \\
International Conference on Collaboration and Internet Computing (CIC)               & 1                                        \\
\review{International Conference on }Computational Collective Intelligence                                                & 1                                        \\
\review{International Conference on Knowledge-Based and Intelligent Information \& Engineering Systems} & 1 \\
International Conference on Information and Knowledge Management                     & 1                                        \\
{International Conference on Machine Learning and Applications (ICMLA)	}    & {1} \\
International Journal of Computer Science (IAENG)                                    & 1                                        \\
International Journal of Information Technology and Decision Making                  & 1                                        \\
{International Workshop on Computational Approaches to Historical Language Change }    & {1}                     \\
Journal of Big Data                                                                  & 1                                        \\
{Knowledge and Information Systems}	                                         & {1}                               \\
%{Lecture Notes in Computer Science}	                                         & {1}                               \\
Neural Computing and Applications                                                    & 1                                        \\
Pattern Recognition Letters                                                          & 1                                        \\
%{Procedia Computer Science}	                                         & {1}                               \\
%{Proc. of the Annual Meeting of the Association for Computational Linguistics}	        & {1}                               \\
{Technological Forecasting and Social Change}      & {1} \\ 

Vietnam Journal of Computer Science                                                  & 1                                        \\
World Congress on Services                                                           & 1                                        \\ \hline
\end{tabular}
}
\label{tab:venues}
\end{table}

\subsection{ Drift Categories}
\label{subsec:td-categories}

\begin{figure}[!htp]
\includegraphics[width=0.6\textwidth]{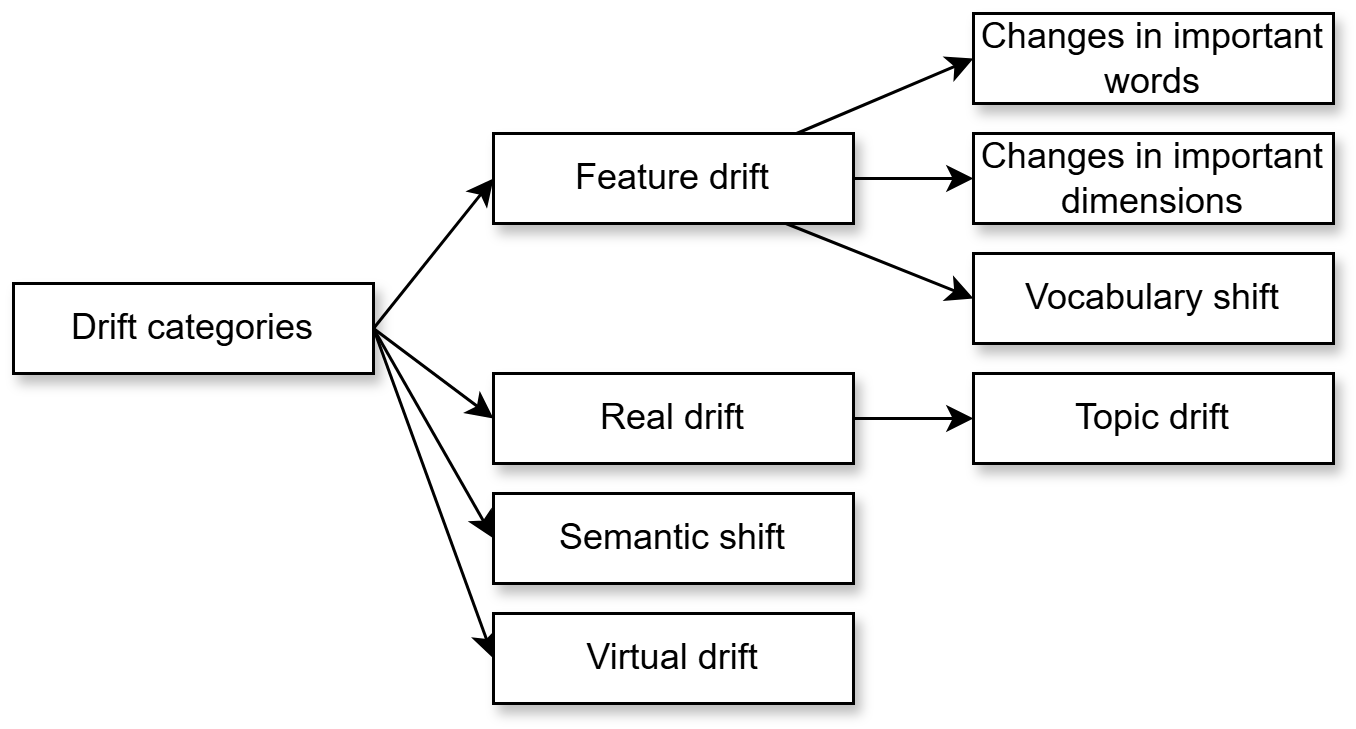}
\centering

\caption{Drift categories.}
\label{fig:text-drift-categories}
\end{figure}

Considering the categories of concept drift in text stream settings, we arranged them into (i) \textit{Feature drift}; (ii) \textit{Real drif}t; (iii) \textit{Semantic shift}; and (iv) \textit{Virtual drift}. Fig.~\ref{fig:text-drift-categories} depicts the arrangement.

\subsubsection{Feature drift}
Feature drift considers the changes in the importance of features, signifying that, over time, a subset of features may become necessary for an ML model, while other subsets may become obsolete \citep{barddal2017survey}. It constitutes a challenge for an ML model because incrementally defining the best feature set over time can be complex. In addition, the dependent ML model can have its performance degraded over time if the selected feature set is inadequate. 

Considering the subcategories of \textit{feature drift} depicted in Fig.~\ref{fig:text-drift-categories}, we describe the \textit{Changes in important words}, \textit{Changes in important dimensions}, and \textit{Vocabulary shift}. Different types of features regarding text-related tasks may be considered in ML approaches. For instance, one approach is to consider the texts split into tokens or use direct techniques such as bag-of-words or TF-IDF. {When considering 1-gram, \eg, a single word/token, t}hese techniques resort to counting tokens and measuring their overall importance, respectively. {However, these techniques can be used in n-grams, leveraging subsequent words. In the papers studied in this work, two papers leverage 2-grams (bigrams): \citet{p23} and \citet{p40}.}

%\JPBCOMMENT{Uma coisa que senti falta é a questão de n-grams. BOW e TFIDF podem ser usados nesse formato e não apenas em 1-grams.}
%\CGCOMMENT{Adicionei um pequeno pedaço de texto no parágrafo anterior e especifiquei no posterior.}
Other approaches regarded the numerical transformation of texts, such as Word2Vec \citep{mikolov2013efficient}. {Since methods such as bag-of-words and TF-IDF, in 1-gram fashion,} can be directly related to specific words (or tokens), and the changes in those words are regarded in this study as \textit{Changes in important words}. However, changes in numerical representations without direct relation to words or tokens are regarded as \textit{Changes in important dimensions}. {For instance, in the case of bag-of-words and TF-IDF, each column represents a given word/token. If a word from an arbitrary point in the text stream starts appearing more (or less) than a previous point, this change would be considered \textit{Changes in important words}. On the other hand, if we leverage a Word2Vec representation, we cannot link directly to a word since the representation learning process is based on semantic connections, using surrounding words to predict a target word, \ie, Continuous bag-of-words (CBOW), or vice-versa (Skip-gram). In the case of Word2Vec, any change in dimensions would correspond to \textit{Changes in important dimensions}.}

Finally, we considered \textit{vocabulary shift} as one type of feature drift. Vocabulary shift \citep{p22} ponders the changes of words in a vocabulary maintained by the approach as a type of text drift. Different from the aforementioned subtypes of feature drift, vocabulary shift considers the changes, \ie, addition or removal of items, in the internal structure that stores the tokens.
{\citet{p22}} compared vocabularies in year-timed slices, measuring changes between vocabularies from different years.

\citet{p22}, \citet{p13}, and \citet{p5} addressed one of these aforementioned categories of feature drift directly. \citet{p5} proposed an ensemble-based method for predicting feature values in the next time point. Considering this case, the work was categorized as \textit{Changes in important words} because their method used a sketching mechanism to retain essential words in a fixed-size feature space, according to their occurrence count. In one version the authors presented, they utilized ADWIN \citep{bifet2007learning} to evaluate a significant decrease in word usage to decide when to remove it from the sketch.

\citet{p13} proposed a feature selection method based on correlation suitable for data streams, categorized as \textit{Changes in important dimensions}. Although the method was not developed specifically for use on text streams, the authors demonstrated its use on a text-related dataset, \ie, a spam dataset. Their method retained a covariance matrix coupled to a concept drift detector. Whenever it received a warning signal, the covariance matrix was incrementally updated. When the concept drift detector triggered a drift signal, a one-pass algorithm computed feature-feature and feature-class correlations. Subsequently, a new Naive Bayes model was trained based on the new feature subset, which was chosen according to the merit of each feature subset from the correlation-based feature selection method (CFS) \citep{hall1999correlation}.

Unlike prior works, \citet{p22} used \textit{vocabulary shift} to estimate the changes in the usage of tokens across several years, \ie, between 2013 and 2019. The authors proposed sampling methods for updating BERT models \citep{devlin2018bert} to maintain the models' usefulness in text-streaming scenarios. Initially, the authors emphasized that ``vocabulary is the foundation of language models''. However, vocabularies can contain different types of representation, such as complete words and sub-word segments, \eg, wordpiece \citep{devlin2018bert}. The authors analyzed the vocabulary shift considering the 40,000 most frequent tokens, {accounting for} hashtags and wordpieces. Regarding hashtags in 2013 and 2019, the vocabulary shift was 78.31\%, while for wordpieces in the same period, the shift was 38.47\%. The authors argued that these results and their analysis justify the development of such an incremental method proposed by them. Furthermore, the authors stated that although larger vocabularies may lessen the vocabulary shift, they were more computationally costly and, therefore, potentially infeasible for real-world scenarios.

\subsubsection{Real drift}
\label{subsubsec:real-drift}
We considered \textit{real drift} according to the definition in \cite{gama2014survey}, which is changes in $p(y|\boldsymbol{X})$ that can occur with or without changes in $p(\boldsymbol{X})$. Considering this case, $\boldsymbol{X}$ regards the input features, while $y$ corresponds to the class, and $p$ is the probability. Real drift in a classification task refers to the change in the classes' boundaries, which may be accompanied by changes in the data distribution in $\boldsymbol{X}$.
%. In addition, the data distribution may happen together with this change in boundaries. 
In this work, few papers handle different types of real concept drift, \eg, \textit{sentiment drift}. However, because they regarded changes in $p(y|\boldsymbol{X})$, these papers were categorized as real drift.

This study considered \textit{topic drifts} as an extension of \textit{real drifts}. In the literature, topic drifts are encountered in applications regarding topic modeling, topic labeling, and short-text classification. Thus, a topic could drift by the change of either text labeled as a particular topic, \ie, $p(y|\boldsymbol{X})$), or by the change of a topic distribution in the stream, \ie, $p(\boldsymbol{X})$, or both simultaneously. In addition, it is common to use methods based on Latent Dirichlet Allocation (LDA) in short-text-related applications.

A significant number of papers regarded exclusively \textit{real drifts} \citep{p2,p3,p9,p11,p12,p15,p16,p17,p18,p19,p20,p21,p22,p-25,p28,p29,p32,p33,p34,p37,p38,p40,p41,p42,p43,p45,p46,p48}. Most commonly, methods in this category either: (i) used concept drift detectors to detect drift and trigger the model update or (ii) updated the model regularly.  

\citet{p11}, \citet{p12}, and \citet{p14} presented from multiple perspectives a system for detecting physical events with emphasis on landslides, \ie, the sudden mass of rock and earth movements downwards steep slopes. %\footnote{According to the Cambridge dictionary: ``a mass of rock and earth moving suddenly and quickly down a steep slope''. Available at: https://dictionary.cambridge.org/us/dictionary/english/landslide.}.
They combined data from social media (which is voluminous but not so trustworthy) and governmental reports (scarce but trustworthy) to train a model for landslide detection. The authors argued that the terminology \textit{landslide} can suffer concept drift because of its use in different contexts, such as politics. In their case, the model was updated regularly, using the governmental reports as ground truth. However, \citet{p17} and \citet{p28} utilized concept drift detectors to detect drifts explicitly. \citet{p17} used ADWIN \citep{bifet2007learning}, DDM \citep{gama2004learning}, EDDM \citep{baena2006early}, and Page Hinkley \citep{page1954continuous,sebastiao2017supporting}, while evaluating fake reviews detection. The authors claimed that fake reviews could lead customers to make poor decisions. Also, it is an adversarial problem: once {models} become better at detecting fake reviews, the unlawful reviewers change patterns over time to overcome the models. The adversarial aspect of this problem results in concept drift, which can cause the models{' performance} to degrade over time. 

{\citet{p38} proposed a complete system for tweet collection, automated training data generation, and BERT (re)training for sentiment prediction and adaptation to sentiment drift, namely Twitter Sentiment Drift Analysis - BERT (TSDA-BERT). The authors used Apache Kafka\footnote{\url{https://kafka.apache.org/}} to simulate the Twitter stream. A BERT model had a three-layer dense network on top that performed the classification. Since the sentiment drift is verified using the predictions, we categorized this paper in the \textit{real} drift category.}

{\citet{p40} proposed a 2-phase online method for textual clustering, namely textClust. This method leveraged TF-IDF to decide the proximity of incoming text to microclusters. In addition, the authors took advantage of unigram and bigram representations and used cosine similarity to evaluate the most suitable cluster to include the incoming text when possible. Over time, in the offline phase, the method could maintain the model concisely by merging similar clusters and removing outdated ones. To define the outdated clusters, the authors used a fading factor for the cluster weights. The authors mentioned that the fading factor helps the model handle concept drift. }

{\citet{p41} performed an experiment to detect changes in sentiment in tweets regarding a soccer match using drift detectors and a lexicon-based classifier. The authors collected tweets during a soccer match between a Brazilian and a Chilean team during the Sudamericana Cup. The context comprised two legs: the first leg, Internacional (the Brazilian soccer team) lost the match by 2-0. During the week between the matches, online influencers created an atmosphere to encourage Internacional to reverse the score. However, Internacional conceded a goal for Colo-Colo (the Chilean soccer team). The system could detect the average sentiment regarding the Internacional's supporters. However, during the match, Internacional scored three goals, the average sentiment became positive, and the sentiment changes could also be detected by the system using drift detectors. 
}

{\citet{p43} presented an incremental semi-supervised method for multilabel text streams named OSMTS. Their method used the initial part of the stream to create the first micro-cluster structure, and from that, the incremental classification occurred. In addition, the method was capable of keeping itself concise by removing stale micro-clusters using an aging scheme and merging micro-clusters when they are similar enough. The micro-cluster used in this paper stores eight pieces of information, \eg, number of documents, word frequencies, the sum of word frequencies, label, decay weight, last update timestamp, and the timestamp of arriving words. An interesting part of the method was that it leveraged the relationship between labels, which made sense for a multilabel scenario. The authors implemented their approach on MOA and evaluated it using nine datasets. In addition, the authors compared it to 12 other methods, obtaining the best results in most datasets having only 20\% of the data. The authors evaluated their approach in terms of hamming loss, example-based accuracy, and micro-average recall. In addition, compared to other approaches, their method was conservative regarding memory.}

Another significant number of papers approached the \textit{Topic drift} problem {\citep{p1,p4,p6,p7,p8,p23,p24,p26,p27,p30,p36,p44}}. Topic drift primarily refers to short-text-related tasks, which commonly require additional steps to provide satisfying results, \eg, data enrichment step or use of statistical information of the application context. 
\citet{p4} proposed a method for short-text classification using feature space extension. Probase \citep{wu2012probase}, an open semantic network, was used for the extension. According to \citet{p4}, Probase was selected by the availability of several super-concepts. It means that, in order to enrich a short text, they could obtain more information from Probase, \eg, super-concepts(Apple) = [company, tech giant, large company, manufacturer], {and add it} to the short text. \citet{p23} developed EStream, a method for efficient short-text clustering. Their approach used lexical, \eg, bigrams, unigrams, biterms, and semantic information from GloVe \citep{pennington2014glove} to define the clusters. Changes in proximity between text and clusters over time were used to determine whether a concept drift occurred. 

Both \citet{p1} and \citet{p7} used LDA \citep{blei2003latent} to address their challenges (short-text classification and topic modeling, respectively). As \citet{p4}, \citet{p7} enriched data using external sources. They employed LDA to mine hidden information from these external sources to add the top representative words in the short texts. Drifts were flagged by calculating the semantic distance between each short text in the current and subsequent chunks. Similarly to \citet{p7}, \citet{p1} used LDA for topic modeling in document streams. In this case, the authors integrated an ADWIN to LDA to detect topic drifts.

{\citet{p36} presented a method for short-text classification in text stream scenarios. The authors enriched short texts by using representations from BERT and Word2Vec. Both were trained using massive corpora, which, according to the authors, should be highly consistent with the topics related to the datasets the authors evaluated. In addition, the authors proposed a distributed LSTM-based ensemble method that includes a concept drift factor. The concept drift factor was used to determine the importance of an LSTM layer in the final result.}

{\citet{p44} provided a method called TCR-M for topic change detection and adaptation in textual data streams, which leveraged an ensemble and an extra classifier for error corrections. The topic change recognition process, according to the authors, not only detects the changes but also scores the severity of the change. The authors first used LDA for topic extraction, which was limited to ten. To detect drifts, the authors measured changes in topic probability between the current, the previous, and the next chunks. A potential change was scored by the statistical test with $0.1$ of significance. If the p-value was below $0.1$, the change was considered severe. The degree of severity defined whether the extra classifier should be retrained. The authors evaluated their method against a bagging model, Learn++.NSE \cite{elwell2011incremental}, and LeverageBagging. Their method was presented in two versions: TCR-M, which reconstructed a bagging model at each time point, and TCR-M (retrain), in which the extra classifier was retrained based on the results. It is not clear if TCR-M had its bagging model fully reconstructed at each time point. The authors used an Amazon review dataset but split it into six subsets related to categories. Their method, mainly TCR-M (retrain), obtained the best accuracy values in four out of six subsets. However, in terms of F1-Score, the same method performed best only in one subset. The discrepancy of results across the metrics is not discussed, although the authors mentioned that the method was ``only a preliminary attempt for text stream learning''.}

%% ALEKOE: \cite -> \citet (19-9) Parei aqui.

%%% PAREI AQUI (17/05). Página 20 do documento word

\subsubsection{Semantic shift}
\textit{Semantic shift} regards changes in the meaning of tokens over time. It is most commonly handled in papers that study linguistic changes over several years, decades, or even centuries. Generally, the datasets that support these tasks are entitled \textit{diachronic}. However, semantic changes can also occur within a short time, such as in weeks \citep{stewart2017measuring} {or minutes/hours \cite{p41}}. The semantic shift was briefly introduced and discussed in Section \ref{sec:dsm}.

\citet{p22}, \citet{p31}, and \citet{p35} approached the problem of semantic shift. \citet{p22} discussed the semantic shift as an analysis of whether it occurred. In the specific task of hashtag prediction, the authors evaluated the shift in top contextual words of the hashtags \#china, \#uk, and \#usa, considering the years 2014 and 2017. The authors agreed that, in 2014, the contextual words related to \#usa were related to the World Cup, while in 2017, the words were related to US politics. However, \citet{p35} aimed at detecting semantic shifts incrementally. In this case, the authors applied clustering methods, such as affinity propagation, to generate clusters in time slices. The authors determined a semantic shift by measuring the distance between embedding sets using metrics such as Jensen-Shannon divergence \citep{nielsen2019jensen} and the distance between prototype embeddings.
%The authors used metrics such as Jensen-Shannon divergence and distance between prototype embeddings to measure the distance between sets of embeddings, and thus determine a \textit{semantic shift}.
\citet{p31} presented a word-level graph-based method to generate dynamic word embeddings. The fundamental concepts were around maintaining long-term and short-term word-level knowledge graphs. These graphs preserved the co-occurrence between words. The relations between words helped define the occurrence of \textit{semantic shifts}. For semantic shift detection, the authors evaluated the closest words to \textit{apple} (in the New York Times dataset) and \textit{network} (in the Arxiv dataset). In addition, the authors evaluated their method by considering trend detection and text stream classification. {Although the aforementioned papers selected a small number of words to evaluate, there are shared tasks that monitored an entire vocabulary over time, \eg,  \citet{zamora2022lscdiscovery}, allowing participants of the shared task to develop their solutions, either considering the text streaming constraints or not.}
%In both \citep{p31} and \citep{p35}, the words of interest in detecting semantic shifts must be known in advance.   

\subsubsection{Virtual drift} 
\label{subsubsec:virtual-drift}
According to \citet{gama2014survey}, \textit{virtual drift} regards changes in data distribution without changing the boundaries between classes. Using a similar notation as in Section \ref{subsubsec:real-drift}, virtual drift happens when $p(\boldsymbol{X})$ changes but $p(y|\boldsymbol{X})$ does not. In addition, \citet{gama2014survey} stated that different definitions exist for virtual drift in the literature. \citet{p14} and \citet{p29}\citep{p39} illustrated {the \textit{virtual drift}} category. Virtual drifts must be tracked, particularly in cases where no classes or clusters' labels $y$ are available.

\citet{p14} proposed a method for landslide detection. The method relied on social media data and governmental reports. Section \ref{subsubsec:real-drift} already cited this paper together with \cite{p11} and \cite{p12}. However, \citet{p14} explicitly emphasized their concern about handling the \textit{virtual drift} problem. They highlighted that model fine-tuning is sufficient in this case, compared to model re-creation. {Nonetheless}, no reason for their concern about virtual drifts was provided. \citet{p29} presented a two-component method for concept drift detection applied to sentiment analysis and opinion mining. Similar to \citet{p14}, \citet{p29}\citep{p39} handled virtual drift. Although it is not explicit in the papers, the drift detection method used two windows to evaluate possible concept drift based on a distance metric to be selected. Different from most works that coupled a concept drift detector with a classifier to utilize the classification errors as a proxy for the detector, 
\citet{p29}\citep{p39} used the input data, thereby using the concept drift detector to check $p(\boldsymbol{X})$.

\subsection{Drift Detection Methods}
\label{subsec:td-detection}

\begin{figure}[!htp]
\includegraphics[width=0.6\textwidth]{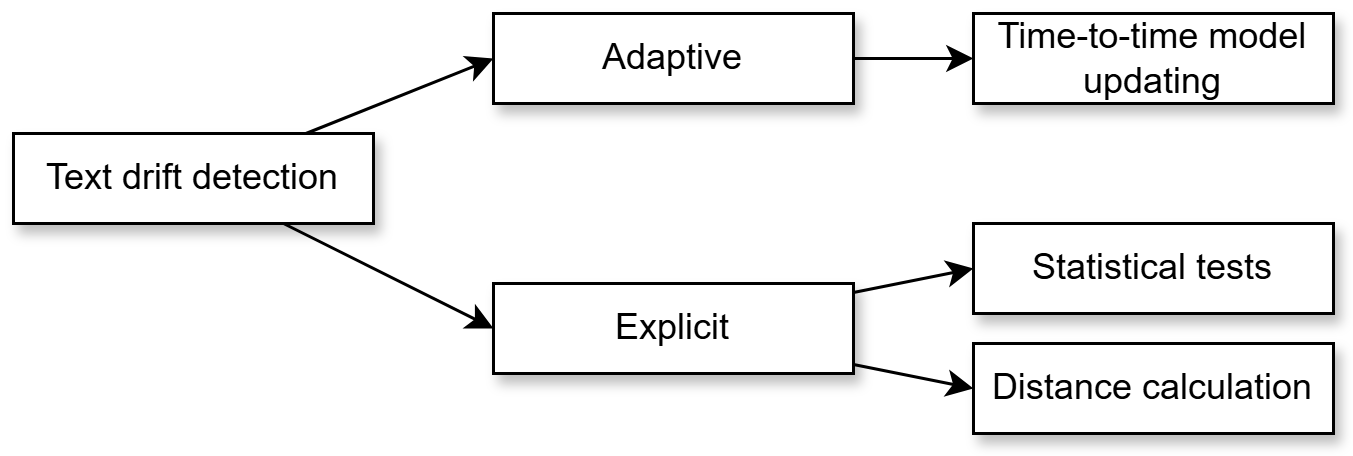}
\centering

\caption{Drift detection categories.}
\label{fig:text-drift-detection}
\end{figure}

We considered two categories for drift detection methods: \textit{Adaptive} and \textit{Explicit}. Fig.~\ref{fig:text-drift-detection} depicts the categorization regarding the type of drift detection. In subsequent subsections, we describe 
%\akst{some} 
selected papers from each drift detection scheme. 

\subsubsection{Adaptive} Adaptive corresponds to a self-updating model without explicitly detecting drift but rather from time to time. This category was called \textit{blind adaptation} in \cite{gama2014survey}. A substantial number of papers considered \textit{adaptive} approaches {\citep{p2,p3,p5,p6,p9,p10,p11,p12,p16,p18,p19,p20,p22,p23,p24,p26,p27,p30,p31,p33,p34,p40}}. \citet{p2} proposed a batch-based method with an application for crisis management in social media. The method was based on active learning, and a user was queried whenever the classifier 
%is not confident enough to output whether an input text is relevant or not for the task. 
failed to confidently determine whether the input text was relevant to the task. The authors selected two events corresponding to two subsets from a more extensive dataset, \ie, Colorado floods and Australian bushfires, and 1000 data points were labeled via crowd-sourcing. In addition, the authors mentioned that labeling data was particularly costly in streaming scenarios; however, it still required a human in the loop in a task such as crisis management. According to the authors, this model can adapt itself in the case of concept drift using the characteristics of the ML technique. For example, although they applied their scheme using k-nearest neighbors (k-NN) and support vector machines (SVM), it could be any other classifier. For instance, the authors claimed that when using k-NN and SVM, the continuous calculation of the boundaries results in drift adaptation.

\citet{p22} split the social media data by considering the years of publication. The work considered two datasets, corresponding to three different tasks: (i) 2014 Country hashtag prediction, (ii) 2017 Country hashtag prediction, and (iii) OffensEval 2019. The authors compared seven methods for each scenario: two static BERT models and five dynamic BERT models. Considering the static BERT models, one was trained with data from the previous year, and the other used data from the current year. For example, considering the 2014 Country Hashtag prediction task, one model was trained with tweets from 2013, and the other (a model checkpoint from the first model) was updated with an amount of data from 2014. The dynamic BERT models were fine-tuned using sampled tweets from the current year using different sampling methods, \eg, uniform random, weighted random, token embedding, sentence embedding, and token masked language modeling (MLM) loss. The sampling methods defined different strategies for the model to overcome drifts/semantic shifts over time. 
The uniform random sampling was regarded as a sampling method in which the tweets from the current year were sampled randomly. In addition, the weighted random sampling method was used to sample the tweets from the current year randomly, considering the number of wordpieces generated by the tokens in the current year's tweets.
%The uniform random sampling method regarded sampling methods in which the tweets from the current year were sampled randomly. 
%In addition, the weighted random regarded a sampling method in which the tweets from the current year were sampled randomly, considering the number of wordpieces generated by the tokens in the current year's tweets.
However, token embedding, sentence embedding, and token MLM loss differ. The token embedding method assigned higher weights to tweets that contained new tokens and random samples from the current year's tweets. The sentence embedding method calculated the cosine distance between the updated and the current models. Both cosine distance and tweet length were used to determine a score, and then the sampling was performed. The token MLM loss method considered the last layer from the BERT model, masked out 15\% of the tokens, and used the surrounding words to predict the masked ones. A high loss value may indicate drifts.

\citet{p6} proposed two algorithms for short-text stream clustering: MStream and MStreamF, a concise version that deletes outdated clusters. The algorithms receive document batches and are one-pass, in which the first document creates a new cluster, and the subsequent either selects one of the clusters to be assigned to or creates a new cluster. This assignment occurs after the batch is processed. The authors argued that concept drift is handled by assuming that the documents were generated by a Dirichlet Process Multinomial Mixture (DPMM) \citep{antoniak1974mixtures} and thus derived the probabilities of documents belonging to existing clusters.

\citet{p16}, \citet{p20}, \citet{p33}, and \citet{p46} tackled {three} different problems similarly: stance detection about vaccination, the Green Pass (as the EU Digital COVID Certificate is known), {and body shaming detection}, {all} in Italy. The authors in {the first} work categorized the application into \textit{stance detection}, a branch of sentiment analysis. In these cases, the tweets were classified in a three-class fashion as either (i) in favor, (ii) neutral, or (iii) not in favor. {\citet{p46} addressed the task of binary classification regarding the use of body shaming language.} \citet{p16} and \citet{p20} analyzed public opinion about vaccines in Italy based on tweets. \citet{p16} addressed concept drift by incrementally retraining the model, such as an SVM model. However, they emphasized that considering their dataset, incremental retraining could not outperform a static SVM in terms of accuracy.
%provide better performance in terms of accuracy than the static SVM. 
\citet{p20} handled concept drifts similarly to \citet{p16}. However, the tweets from the new batch were semantically weighted according to previous events. Thus, the authors reached better values than other approaches, \eg, static model, regular retrain, DARK \citep{costa2017adaptive}, and the proposed semantic scheme. Although \citep{p20} was published in 2021, it was applied to regular vaccinations unrelated to COVID-19. However, \citet{p33} covered the opinion about the Green Pass concerning COVID-19. The authors evaluated different schemes to handle concept drift, including retraining with sliding windows and an ensemble of classifiers. The complete retraining led to the best average accuracy. Still, the highest feature space was reached due to the data accumulation and the utilization of TF-IDF as a text encoding method that generates a very high-dimensional representation. {\citet{p46} presented an approach for body shaming detection in Twitter posts between 2021 and 2022. Interestingly, the authors evaluated approaches considering the concept drift problem. However, the authors leveraged a ``regularly retraining'' approach rather than explicitly detecting concept drifts. The authors used TF-IDF representation to test the Complement Naive Bayes (CNB), Logistic Regression, and SVM. 
%The authors tested the Complement Naive Bayes (CNB), Logistic Regression and SVM, and used TF-IDF representation. 
In the experiments, considering static, incremental, and sliding approaches, the best results were obtained by the CNB using the sliding approach. The sliding approach used a queue to manage the storage of new and past data.}

{\citet{p40} presented an online method for textual clustering, \ie, textClust. In order to overcome concept drifts, the method leveraged a fading factor. It helped the model to exclude stale clusters. In addition, there was another parameter $tr$ that dynamically determined the distance limit for a cluster to merge with another. This was also used to help determine whether a new input instance should be incorporated into a given cluster. }

\subsubsection{Explicit} 
\label{subsubsec:explicit}
\textit{Explicit} approaches directly detect the drift via statistical tests or distance calculation. As examples of statistical tests used in the selected papers, we mention the Page-Hinkley test \citep{page1954continuous,sebastiao2017supporting}, and ADWIN \citep{bifet2007learning}. As examples of distance calculation metrics, we cite the Jensen-Shannon divergence \citep{nielsen2019jensen}, the Kullback-Leibler divergence test \citep{kullback1951information}, and the cosine distance. Several approaches explicitly handled concept drift \cite{p1,p-25,p4,p7,p8,p13,p14,p15,p17,p21,p28,p29,p35,p36,p37,p38,p39,p41,p44,p47}.

Concerning explicit detection using statistical tests, \citet{p17} tested four concept drift detectors: ADWIN \citep{bifet2007learning}, DDM \citep{gama2004learning}, EDDM \citep{baena2006early}, and Page-Hinkley test \citep{page1954continuous,sebastiao2017supporting}. We considered ADWIN a statistical test because, in the original paper, the authors indicated that their statistical test verifies whether the observed average in subwindows is above a defined threshold \citep{bifet2007learning}. In addition, DDM \citep{gama2004learning} and EDDM \citep{baena2006early} performed evaluations based on the statistical properties of a stream and thus were considered in this work a \textit{statistical test}. \citet{p17}, simulated concept drift by splitting the temporally ordered dataset into five chunks and rearranging them. The concept drift detectors used the calculated accuracy over the most recent input data as a proxy, \ie, a window size of 200. ADWIN and EDDM had the best accuracy (coupled with a classifier) among the scenarios tested in the study.

\citet{p28} proposed a method that uses random projection for dimensionality reduction using text streams as input. In their experiments, preprocessing was done offline for the whole dataset to generate TF-IDF and embedding representations. 
%the authors state that ``preprocessing was done in offline mode after all tweets were crawled because, for the embedding and TF-IDF measure, all occurring words must be present at the time of processing'' \citep{p28}. 
Thus, their process was not fully incremental, except for the dimensionality reduction method, which was incremental (in batches). 
Considering the real-world dataset, \ie, NSDQ, proposed in the same paper, the authors obtained a vector representation of 3442 dimensions using TF-IDF. Using their online dimensionality reduction method, NSDQ was projected onto 200 dimensions. The authors concluded that random projection could reduce the run time, even considering the offline preprocessing time. To detect concept drift, the authors used KSWIN \citep{raab2020reactive}, based on the Kolmogorov-Smirnov test \citep{kolmogorov1933sulla,smirnov1948table}. In this case, KSWIN monitored every dimension of the vector representation. In addition, the authors mentioned that different types of concept drift might be present because NSDQ is a real-world dataset \citep{p28}. Their assessment of concept drift detection relied on true positives and false positives. However, it is unclear how both metrics were calculated due to the absence of labeled drifts in the dataset. The results indicated more concept drifts detected in the original space, an expected outcome because KSWIN monitors each dimension separately. Finally, the authors mentioned that models trained with original and projected feature spaces maintained the same level of accuracy. Both \citet{p14} and \citet{p28} used t-SNE \citep{tsne} plots to support the existence of concept drift in the datasets on which they applied their proposed methods.

{\citet{p41} evaluated the use of drift detectors for sentiment drift detection, using collected data during a soccer match in South America. The authors used the Incremental Word Context (IWC) \cite{p32} to trace back the events that generated the sentiment drifts. Using IWC, it was possible to determine which events generated the drift, who participated in them, and the atmosphere of the moment. The authors evaluated three drift detectors: ADWIN \cite{bifet2007learning}, EDDM \cite{baena2006early}, and HDDM \cite{frias2014online} (in the averaged and weighted versions). ADWIN was the most precise method, having a delay of around 2 minutes and raising only one false alarm.}
%\JPBCOMMENT{adicionar referencias para detectores, mesmo q elas ja tenham aparecido antes}
%\CGCOMMENT{Referências adicionadas.}

%\textbf{Explicar explicit + distance calculation}
Considering the \textit{Explicit} detection with the aid of distance metrics, \citet{p4} developed a method for short-text classification in the presence of topic drifts. As explained in Section \ref{subsec:td-categories}, the approach automatically enriched the short texts using Probase. The topic drift detection was performed as follows: the short-text stream was received in chunks, and after they were clustered, the label distribution could be evaluated using the clusters. Subsequently, the distance between the cluster centers in sequential chunks was calculated using the cosine distance. According to the value obtained, the method categorized it either into: (a) no drift, (b) noisy impact, or (c) topic drift. In addition, the authors simulated topic drifts by generating datasets with topic changes after fixed periods. Their detection method was compared to nine drift detectors. Regarding false alarms, missing drifts, and delay, the proposed method obtained high average rankings, which were statistically equivalent (using the Bonferroni-Dunn test) to the best drift detectors in each metric.

\citet{p29, p39} developed an ensemble classifier coupled to a novel mechanism for drift detection-based adaptive windows (DDAW). Their method suited text streams, especially users' sentiments and opinions. Their approach can be divided into two components: (i) drift detection and (ii) classification. In many applications, classification errors are used as a proxy for the drift detector. However, the drift detection component compared the data distribution considering two windows. Thus, it was possible to measure drift by evaluating the dissimilarity between the windows. An intriguing aspect of this approach was that it allowed for distance metrics and statistical tests. In the paper, the authors compared the Hellinger distance \citep{hellinger1909neue}, Kullback-Leibler divergence \citep{kullback1951information}, Total Variation distance, and the Kolmogorov-Smirnov test \citep{kolmogorov1933sulla,smirnov1948table}. Their approach, coupled with the Hellinger distance, obtained the best values regarding false alarms, detection rate, and accuracy, even compared to other drift detection methods, \ie, AEE \citep{kolter2005using}, RDDM \citep{barros2017rddm}, and Page-Hinkley \citep{page1954continuous,sebastiao2017supporting}. It was unmentioned how the drifts were labeled or whether the data was rearranged to simulate drifts. 

\citet{p14} developed a system for landslide detection, a physical event that causes destruction and for which there are no physical sensors to detect. The authors combined data from social media and governmental agencies to perform the detection. Concept drift was detected using the Kullback-Leibler divergence test \citep{kullback1951information} to evaluate the distribution of two batches. The model was updated by generating or updating the classifiers to handle the concept drift.

{\citet{p36} presented a distributed long short-term memory (LSTM)-based ensemble method for short-text classification in text stream scenarios. The short texts were enriched by using BERT and Word2Vec models. The LSTM-based method included a concept drift factor used as a threshold to compare the distance between the LSTM layer trained with the previous batch and the layer trained with the current batch. If the concept drift factor was above the threshold, the weight of the current layer would be bigger to generate the combined final output.}

{\citet{p37} proposed a fuzzy-formal-concept-analysis-based index for concept drift detection and applied the method to a fake news classification problem. Although the concept drift detection was not directly approached, the authors calculated the correlation between the classifier's performance and the proposed index. The index was calculated from a fuzzy lattice, \ie, a fuzzy hierarchical knowledge structure, while the classifier's performance was calculated using F-Score and accuracy. Their results demonstrated a high (Sperman's and Pearson's) correlation, between 69\% and 87\%. The authors claimed that the method had the potential to be used as a proxy for the model update process. In addition, the fuzzy lattice seemed never to be updated, which may hamper the model from properly working over a long time.}  
{\citet{p38} proposed a sentiment drift analysis system based on BERT models, namely TSDA-BERT. According to the authors, the system receives data in a sliding window fashion corresponding to four days. The authors calculated the positive and negative scores per window based on the proportion of them in the window. From these values, a sentiment drift measure was calculated by simply subtracting the number of negative from the number of positive tweets. This measure was used for sentiment drift detection by calculating it between time periods; if the score was negative and later went positive or vice-versa, it indicated a drift.}

{\citet{p47} proposed a model-agnostic framework for drift detection. More specifically, the authors focused on data drift, \ie, virtual drift. This paper presents a dataset comprising texts used as requests to a virtual assistant. In this case, drifts may occur due to novel topics and deviation from previous topics, and these may result from real problems such as external trends, new features/services introduced by a company, \etc. In addition, the authors mentioned the difficulty of obtaining datasets regarding this scenario, and therefore, they created a novel dataset, mimicking user requests and then introducing drifts. The introduction of drifts was performed using the Parrot paraphrasing framework \cite{prithivida2021parrot} and LAMBADA \cite{anaby2020not}. Although the authors frequently mentioned in the paper the terminologies \textit{stream, text stream, }and \textit{short-text stream}, their approach was not fully incremental. Their approach consisted of training an autoencoder to learn the data distribution of a dataset of interest. From this point, their approach was able to compute the similarity between data chunks and the original distribution, learned by the autoencoder. Later, the drift and change point detection was performed. Additionally, their method contains a module for drift interpretation based on a clustering algorithm. Their method contained a single parameter corresponding to a threshold for the cosine distance between the original representation and the reconstructed (through the autoencoder) to detect drifts.}

\subsection{Model Update Method in Text Stream Settings}
\label{subsec:td-model-update}

\begin{figure}[!htp]
\includegraphics[width=0.75\textwidth]{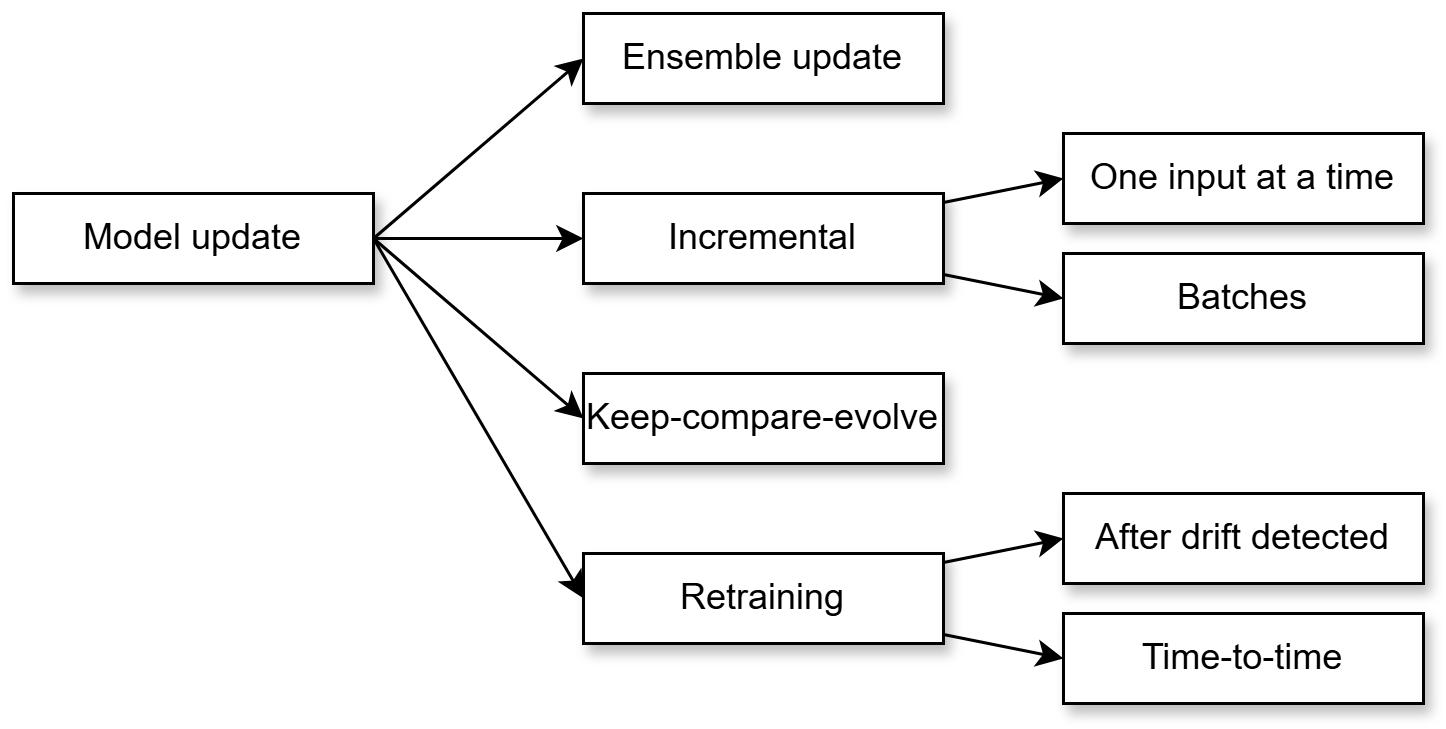}
\centering

\caption{Model update methods used when handling text streams bound to concept drift.}
\label{fig:text-drift-model-update}
\end{figure}

We also looked closely for information regarding the model update scheme from the analyzed papers. Fig.~\ref{fig:text-drift-model-update} depicts the organization. We found four mechanisms: (i) \textit{Ensemble update}, in which the base learners are substituted or removed over time; (ii) \textit{Incremental}, which corresponds to the model incrementally learning new data without a retraining process, splitting regarding the amount of data used to learn: one input at time or batches; (iii) \textit{Keep-compare-evolve}, which corresponds to methods that generate and evolve new models to adapt to drifts and uses the old model to measure the similarity between information from both models; and (iv) \textit{Retraining}, which can occur after detecting a concept drift, or time-to-time, which does not detect drifts but adapts to them.

\subsubsection{Ensemble update}
In this category, the works proposed techniques that create, update, and combine multiple models, the so-called ensembles.
%take advantage of several models in order to join them in an ensemble fashion. 
Over time, an ensemble can be updated by removing outdated base learners while adding new base learners trained on newly arrived data. \citet{p12}, the presented system for landslide detection used batches to update the model. The landslide detector used a classifier, which was an ensemble. The authors mentioned that they used two approaches for selecting base learners: relevancy and recency. When using relevancy, a k-NN search was performed to discover the most relevant base learners from a pool of trained base learners, considering the centroid of the data used to train these learners. However, the recency scheme returned the most recent base learners used to compound the ensemble. In addition, the weighting scheme can be configured as an unweighted, weighted, or model-weighted average. The unweighted average considered the base learners equally to provide an output. The weighted average considered weights provided by domain experts, and the model-weighted scheme considered the base learners' prior performance to weigh them.

\citet{p21} described an ensemble classification model for short text classification in environments bound to concept drift. The paper emphasized three main aspects: a feature extension based on the short text features, a concept drift detection method, and an ensemble model. 
Considering the ensemble model, the authors used SVM as a base classifier. A new classifier was added when concept drift was detected. If the classifier pool is complete, the oldest {classifier} was removed to add the current {one} after being trained on the new batch.

\citet{p7} proposed a short text stream classification method based on content expansion coupled with a concept drift detector. The expansion was performed by adding information from external sources, and 100 Wikipedia pages related to 50 keywords were selected, totaling 60,600 pages. The classification task in this study was performed using an ensemble of SVMs, in which each base learner was trained per chunk using the expanded texts. 
%Their approach works by training an SVM model per chunk, using the expanded texts. Thus, an ensemble is built. 
The number of base learners was limited to a specific parameter $H$: when this number is met, the oldest learner is replaced. In specific situations, the latest learner can replace an older learner trained using semantically similar chunks.

\citet{p29, p39} presented an ensemble method for classification. Particularly, \citet{p39} tackled the sentiment classification problem. The ensemble model was updated over time by removing the worst base learner from the ensemble when it reached the maximum number of base learners. To determine the worst base learner, a weighting calculation is performed by leveraging the base learner's mean squared error on the new input data, \ie, $\textrm{MSE}_i$, and the base learner's mean square error on the data from the previous batch (reference data), \ie, MSE$_r$. The complete weight calculation for a base learner was performed as $weight = \frac{1}{\textrm{MSE}_r + \textrm{MSE}_i + \alpha}$, where $\alpha$ is a non-zero factor to avoid division by zero.

\subsubsection{Incremental}
The \textit{Incremental} update scheme regards models capable of learning from new pieces of data without completely retraining the model. 
In our selection, several papers employed incremental models to approach their applications \citep{p1,p3,p4,p5,p6,p8,p9,p10,p11,p14,p17,p18,p19,p23,p24,p26,p27,p28,p30,p31,p32,p34,p35,p40}. However, we distinguished between the manners in which the data were inputted into the model: (i) One input at a time and (ii) In batches.

\citet{p18} proposed a method for dimensionality reduction using random projection. As already cited in Section \ref{subsubsec:explicit}, the process was not fully incremental. In this study, the authors utilized three classifiers: (i) Adaptive Robust Soft Learning Vector Quantization \citep{heusinger2020passive}, (ii) Adaptive Random Forest \citep{gomes2017adaptive}, and (iii) Self-adjusting Memory k-NN \citep{losing2017self}. The dimensionality reduction method uses a window of size 1000. However, when applied to the classification methods, the process {in} incremental \textit{One input at time}, except for the Self-adjusting Memory k-NN, which \citet{p18} cited that they used as parameters five neighbors and a window size of 1000 to match the window size of the random projection. \citet{p17} incrementally updated the models. In their case, they used Stochastic Gradient Descent for SVM, Perceptron, and Logistic Regression algorithms incrementally. However, similar to \citet{p18}, the process was not fully incremental because it used TF-IDF and principal component analysis (PCA) for dimensionality reduction.

\citet{p3, p10} presented a method for text stream clustering called AIS-Clus, based on the artificial immune system \citep{kephart1994biologically}. This system had online and offline phases. The offline phase comprised receiving historical data to generate the first clusters. In the online phase, new data were divided into equal blocks, \ie, it worked in batches. Concurrently, each instance was evaluated alone, being also capable of handling novel classes. Thus, this work could be categorized as \textit{Incremental in Batches} or \textit{Incremental with One input at a time}, depending on the point-of-view. Although it worked in a clustering fashion, the method performed classification tasks. 

\citet{p1} presented the adaptive window-based incremental LDA (AWILDA), a method for topic modeling in document streams. This method contained two LDA models, one for topic modeling and another for drift detection, with the help of ADWIN. It received the data in batches, making it possible for the approach to use ADWIN as a drift detector and to resort to LDA over the batch. 

{\citet{p40} presented a stream text clustering method. The use of online and offline phases for algorithms that perform stream clustering is well known. The offline phase generally performs adjustments in the model, such as the stale cluster removal and merging of similar clusters. In the online phase, the method received input data and verified the most similar cluster to assign the new input data to the most similar cluster. However, a new cluster is created to accommodate the incoming text if no cluster is sufficiently similar. Due to these characteristics, this method could be categorized as \textit{Incremental with One input at a time}. Interestingly, this {method outperformed} other batch-based methods in the evaluation considered in the paper.
}

\subsubsection{Keep-compare-evolve}

\citet{p22} is the single representative of this model update category. As aforementioned, this study proposed three methods for sampling to update the language models. The three methods, \ie, the Token Embedding Shift method, Sentence Embedding Shift method, and Token MLM Loss method, used both current and previous models to evaluate changes to sample new data to fine-tune the current model. Thus, more significant differences between a given text representation and the representations provided by the old and current models generate higher chances for a given text to be selected for fine-tuning. Thus, in this specific case, it is costly to fine-tune using all the data because of the size of the BERT models. In addition, GPUs are necessary to speed up the training/update of these models.

\subsubsection{Retraining}
Some papers resorted to the complete retraining of models. The retraining can be triggered by drift detection or periodically, typically after batch processing. As noted, \citet{p13} proposed a dynamic feature selection method to handle feature drift, namely Dynamic Correlation-based Feature Selection (DCFS). This method used concept drift detectors, such as ADWIN. Concept drift detectors generally provide two levels of signaling: warning and drift. Whenever a warning signal was outputted, DCFS updated the covariance matrix incrementally. The feature-feature and feature-class correlations were calculated when a drift signal was emitted. Thus, a new Naive Bayes model was trained from scratch using the feature subset selected according to the correlation-based feature selection (CFS).

Other works also utilized the retraining scheme \cite{p16,p20,p33,p46}. All these papers compared approaches that resorted to the retraining scheme. Retraining occurs regularly and considers data from events. However, the dataset was increased incrementally to be used by the methods during the training step. For example, when event \#10 concluded, the data related to this event were appended to the data regarding previous events. Thus, a new model can be trained based on the dataset, now containing the data about event \#10. Concerning these four works, only \citet{p33} used an incremental approach, \ie, Complement Naive Bayes \citep{rennie2003tackling} with the partial fit. For this approach, however, the authors used TF-IDF for vectorization, 
%which demands previous knowledge of the texts.
which was not updated during the online monitoring after the first event. 
Thus, the process was not fully incremental. In addition, the authors did not mention any strategy for maintaining a dataset in a feasible size after several incremental additions of batches. {\citet{p46} used the retraining scheme even with the so-called \textit{sliding} and \textit{incremental} strategies. In their paper, \textit{sliding} added new data and removed old data in a data structure for model retraining, and \textit{incremental} accumulated data over time, which directly impacts the dimension number of the TF-IDF representation.}

{The system proposed by \citet{p38}, \ie, TSDA-BERT, also considered periodic retraining to overcome sentiment drift. Whenever a sentiment drift happens, the system uses a domain impact score, which calculates the impact of a tweet in the domain. The calculation considers the intersection of a tweet's words and the domain-specific impact words. According to the authors, if the impact was above 0.5, it indicates adherence to the domain. However, the authors did not explain how the domain-specific words were selected. Compared to \citet{p16}, \citet{p20} and \citet{p33}, \citet{p38} provided a strategy to maintain the training set in a feasible size. The tweets with higher adherence to the domain were included in the training set, and the same number of tweets were removed from the training set. It means that the training set is always the same size. The authors mentioned the utilization of at most 324,685 tweets in the training set. This training set was used for fine-tuning over time.}

\subsection{Stream Mining Tasks applied in Text Stream Settings}
\label{subsec:td-stream-mining-tasks}

\begin{figure}[!htp]
\includegraphics[width=0.9\textwidth]{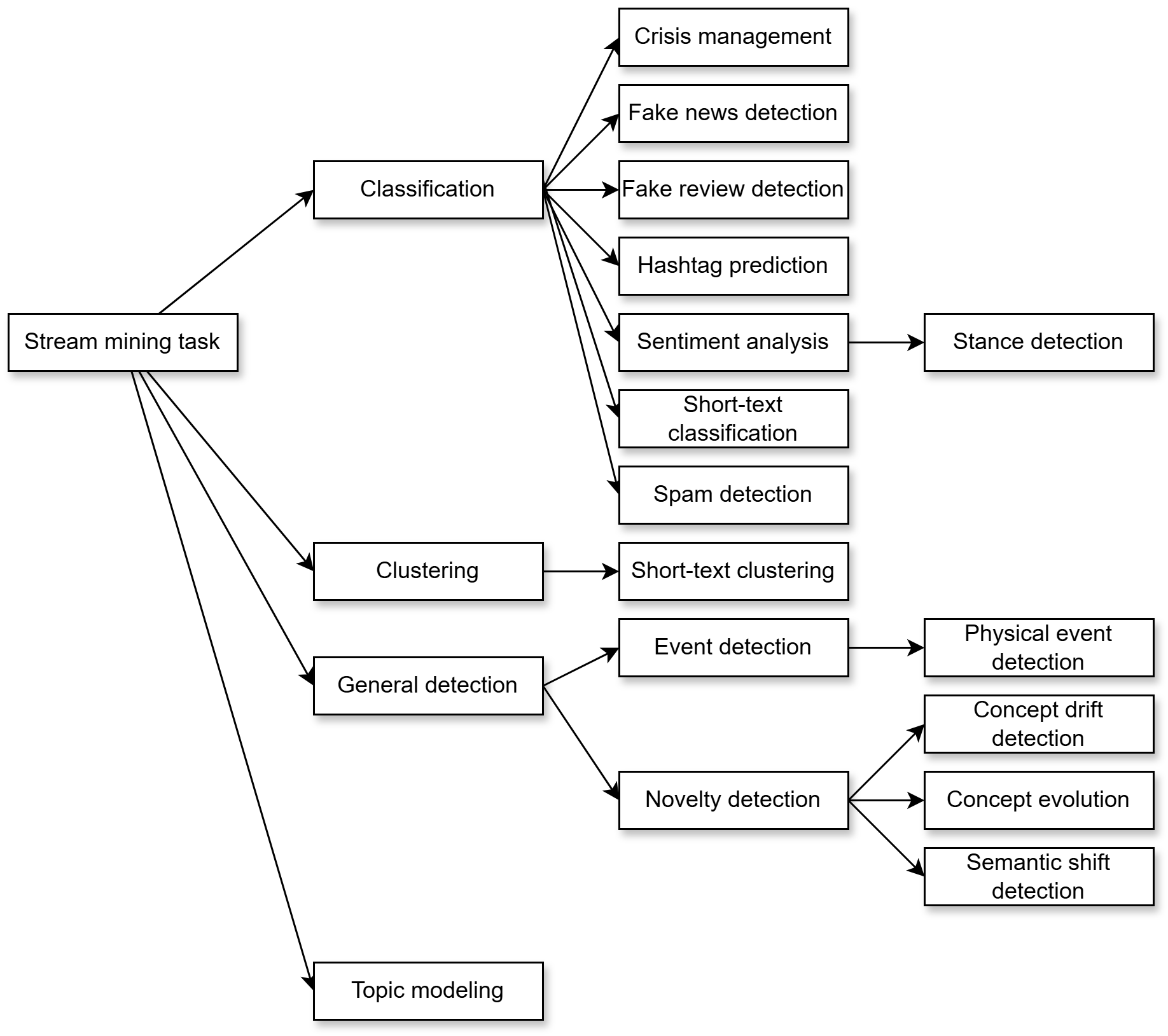}
\centering

\caption{{Text stream mining tasks and applications found in the selected papers.}}
\label{fig:text-drift-stream-mining}
\end{figure}

In Fig.~\ref{fig:text-drift-stream-mining}, we organized the stream mining tasks addressed and {the} respective applications in the analyzed papers, considering the information obtained from the selected papers. {This subsection addresses the Research Question 2 (RQ2), \ie, \textit{``Which type of application is addressed?''}.} In this study, we considered Stream mining tasks: (a) \textit{Classification}; (b) \textit{Clustering}; (c) \textit{General detection}; and (d) \textit{Topic modeling}.

\subsubsection{Classification}
Classification is among the most common stream mining tasks. In the general classification, the objective is to predict, with arbitrary accuracy, a unique class from a small set of values from a given input. Some applications found in the papers addressing the classification task include (i) \textit{crisis management}; {(ii) \textit{fake news detection};} (iii) \textit{fake review detection}; (iv) \textit{hashtag prediction}; (v) \textit{sentiment analysis}; (vi) \textit{short-text classification}; and (vii) \textit{spam detection}. %Fig. \ref{fig:text-drift-stream-mining} depicts the aforementioned stream mining tasks and applications.

Regarding \textit{crisis management}, \citet{p2} aimed to identify the relevant tweets about two environmental disasters: the Colorado floods and the Australian bushfires. It is considered a binary classification task because the model assesses whether or not a tweet is relevant, sometimes with a human in the loop. Their approach was evaluated regarding the average error and the number of queries. {Because the method presented in \cite{p2} employed active learning strategies, the label uncertainty determines whether the system should query a user. } Only \citet{p2} represented this application in the classification task. 

{\textit{Fake news detection} was addressed by \citet{p37}. The authors proposed an index based on fuzzy formal concept analysis, which correlates with the classifier's performance. According to the authors, the fake news detection problem is generally tackled as a binary classification, where a model should classify news as fake or real. The authors evaluated three ML methods: Random Forest, Naive Bayes, and Passive-Aggressive \citep{crammer2006online}. Although the authors proposed the method, they did not couple the index to the methods to trigger retraining. Three datasets containing news articles between 2018 and 2020 were used, \ie, NELA-GT-2018, NELA-GT-2019, and NELA-GT-2020 (see Section \ref{sec:datasets}). According to the authors, only the Passive-Agressive algorithm was tested online, and the news between February and August 2018 were used as the training set, considering also the fuzzy lattice structure. The classifiers' evaluation was performed using accuracy and F1-score. The evaluation of the proposed index happens through visual analysis, Pearson's and Spearman's correlation, and cosine similarity. The authors argued that their method would allow early drift detection but did not provide experiments or evidence. }

Considering \textit{fake review detection}, \citet{p17} tackled this task using three ML methods: SVM, logistic regression, and perceptron. The authors used four Yelp datasets, only one containing fake and genuine reviews. \citet{p17} noted that the datasets ``were built based on an unknown filtering algorithm and web-scraper techniques to label each review as fake or genuine''. Because the idea was to determine whether or not a review is fake, it corresponds to a binary classification task. The ML methods were evaluated using accuracy and statistically assessed using the Nemenyi test \citep{nemenyi1963distribution}. The authors claimed that their work is the first to address concept drift in the fake review detection problem. Considering the selected papers, this was the only method that approached fake review detection.

\textit{Hashtag prediction} was addressed in \cite{p22}, \cite{p18}, and \cite{p28}. Both \citet{p18} and \citet{p28} used random projection as a dimensionality reduction method for text streams. Also, a dataset, \ie, NSDQ, was proposed for the problem because it generated high-dimensional data and could be reduced in real-time by random projection. Furthermore, this is the only real-world textual dataset addressed in these papers, while the others are synthetic. This dataset contains 15 classes that make the stream mining task approached by them a multiclass classification. The evaluation was performed in terms of accuracy, Cohen's Kappa, and run time. \citet{p22} tested the sampling approaches for updating BERT using two datasets: OffensEval 2019 and Country Hashtag Prediction. Approaching OffensEval constituted a binary classification task; thus, the authors used the Area Under Curve (AUC) of the Receiving Operating Characteristic (ROC) curve and F1 score. However, addressing Country Hashtag Prediction corresponded to a multiclass classification task and was evaluated using micro-F1 score, macro-F1 score, and accuracy. 

In \textit{sentiment analysis}, the objective is to develop a model capable of inferring a user sentiment from text. According to \citet{medhat2014sentiment}, ``sentiment analysis (SA) or opinion mining (OM) is the computational study of people’s opinions, attitudes and emotions toward an entity''. Similarly to sentiment analysis, \textit{stance detection} regards the position of a given text's author about a specific topic, considering the labels \textit{in favor}, \textit{neutral/neither}, and \textit{against}, sometimes expressed in literature with different labels but with similar meanings \citep{p20,kuccuk2020stance}. \citet{p20}, \citet{p16}, and \citet{p33} approached \textit{stance detection}, with \cite{p20} and \cite{p16} related to vaccination, and \cite{p33} regarded the stance about the green pass, as mentioned in previous sections. The authors in these three works collected the dataset that they needed to utilize primarily from Twitter. As aforementioned, \textit{stance detection} classifies texts in three labels, indicating that it is a multiclass classification task. \citet{p16} used F1 score, precision, recall, AUC, and accuracy to evaluate the method. \citet{p20} evaluated models using accuracy and F1 score, and \citet{p33} used F1 score, accuracy, and the number of features in each model. 

\citet{p32} proposed a sentiment lexicon inductor for time-evolving environments in a sentiment analysis context. The authors claimed that sentiments could change over time, while new words in different sentiments can emerge. In addition, the lexicon would be static in a fully incremental system without sentiment induction. In this case, from a seed lexicon, the authors processed the dataset in a stream fashion and, at the same time, inferred sentiment from tokens absent in the lexicon. Although, in practice, the system outputs a value limited by a logistic function, we presented this paper in the classification section because of the sentiment analysis application. In addition, the authors tested their approach by deliberately changing lexicon sentiment scores and measuring how long the system would take to recognize the new sentiments. Finally, the authors used accuracy and Cohen's Kappa to evaluate the classifiers applied together with their method.

{\citet{p41} leveraged a lexicon-based classifier for sentiment analysis in a text stream environment regarding a soccer match. The authors observed that the sentiment changes very quickly, derived from the events in the soccer match, supporting the statement that sentiments could change over time \cite{p32}. Using ADWIN, HDDM, and DDM, the authors observed that ADWIN obtained the best results in terms of missing drifts, delay (regarding time and posts), and false alarms. This work also fits the category \textit{Concept drift detection}, because the sentiment stream classification was not the objective, but a means of evaluating the sentiment drift detections. }

%\JPBCOMMENT{o que vc quer dizer com performance? }
%\CGCOMMENT{Ajustado}
{Aiming at improving classification performance using fewer data, some papers, \eg, \citet{roychowdhury2024tackling}, adapted text stream mining tasks. Originally working on regular classification tasks, the authors proposed converting to entailment-style modeling. The method generates augmented data by creating multiple pairs of text and label hypotheses, where only one pair is true, and the others serve as negative examples. This approach enables the model to adapt to new concepts with significantly less labeled data, particularly in few-shot learning scenarios. The proposed technique was evaluated on both real-world and synthetic datasets, reaching the best values regarding macro F1-score. The authors claimed a 75\% reduction in labeling costs compared to regular fine-tuning methods.}

\textit{Short-text classification} is addressed in {\cite{p4,p21,p36} and \cite{p47}}. \citet{p4} proposed a method for short text streams bound to concept drift. This method took advantage of Probase for short text enrichment. The approach was evaluated in terms of time and accuracy. \citet{p21} described a method for text stream classification based on feature extension and ensembles formed by ensembles. This method can handle concept drifts by calculating the distance between each short text in the previous and new batches. {\citet{p36} proposed a method for short text classification in text stream scenarios. This method enriches text by using representations from BERT and Word2Vec. In addition, the method uses a Convolutional Neural Network (CNN) to extract high-level features. This method handled concept drift by resorting to a concept drift factor used in the systems.} Both approaches in {\cite{p4,p21} and \cite{p36}} were applied to the same datasets, {\ie, Tweets, TagMyNews, and Snippets,} considering text classification {as} topics. %Thus, \textit{Topic classification/labeling} can be very similar to the applications that approach short-text classification.

{\citet{p47} addressed short stream classification, using as dataset user requests to virtual assistants. In addition, the authors mentioned that drifts emerge in this scenario due to the deployment of new features/services, external trends, or service interruption, for example. One of the challenges in this specific work was the shortness of the texts, in which most had less than five words. Therefore, to have a significant number of samples, the authors employed a generation method \citep{rahamim2023text}, Parrot, and LAMBADA, reaching 600,000 samples. \citet{p47} proposed four drift scenarios for evaluation: (a) gradual drift, (b) abrupt drift, (c) no drift, and (d) short-lived anomaly. In the drift scenarios, the authors introduced drifts by positioning a subset in specific points of the stream, \ie, uniformly for the gradual drift, and at the timestep \textit{t}=15 for abrupt drift. Their method contained a single parameter, which corresponded to a threshold for the cosine distance between the original representation and the reconstructed (through the autoencoder) to detect drifts. Interestingly, the proposed method is the only one agnostic to the model. However, the method was not fully incremental due to the autoencoder training using the anchor dataset. }

Some papers addressed \textit{Spam detection} as experiments \citep{p5,p13,p19}. Because the goal is to classify a piece of text into either non-spam or spam, the task is considered a binary classification task. \citet{p5} provided an ensemble-based mechanism for predicting a feature's probability of association with a given class by considering that words might be subject to temporal trends and a sketch-based feature space maintenance mechanism that allows for memory-bounded feature space maintenance. The approach utilized an ensemble compounded by statistical techniques to account for feature periodicities. The ensemble consisted of a Poisson model \citep{p5}, a Seasonal Poisson model \citep{holt2004forecasting}, an Auto-regressive Integrated Moving Average (ARIMA) model \citep{box2015time}, and an Exponential Weighted Moving Average (EWMA) model \citep{nishida2012improving}, to capture regular, seasonal, auto-correlated, and sudden trends. A sketch-based approach was designed to maintain a concise feature space. The authors tested three versions: a baseline sketch that retains only word and occurrence counts, a fading sketch that considers the importance of frequent words, and a drift-detector-based sketch, which uses ADWIN to detect the decrease in word usage. The approaches were compared in terms of accuracy, Cohen's Kappa, and run time. 

\citet{p13} proposed a method for feature selection based on correlations to handle feature drifts in data stream scenarios. The method is not exclusive to spam detection, but the spam dataset was the only text-based dataset used by the authors. The method is evaluated in terms of accuracy. 
\citet{p19} proposed a method for feature selection in binary text stream classification tasks, namely OFSER. The proposed method leverages adaptive regularization and weighs the input for each new data. The regularization, according to the authors, decreases the impact of the feature drift. Despite being fast and having decent overall performance, their method depends on a parameter to define the number of features to be selected from the original set. The method runs on top of a Naive Bayes classifier, chosen due to its simplicity and naive assumption of independence among the features. The approach was evaluated using F1 score, accuracy, memory consumption, and run time. Furthermore, due to ``an undesired conservativeness of the Friedman test'' \citep{p19}, it was statistically assessed using the Iman-Davenport test \citep{iman1980approximations} instead of the Friedman test \citep{friedman1937use,friedman1940comparison}, and the Bergmann-Hommels' procedure \citep{garcia2008extension} instead of the Nemenyi test \citep{nemenyi1963distribution}. OFSER ranked among the three best approaches.

As expected, the most frequent metrics in this stream mining task were accuracy, Cohen's Kappa, F1 score, AUC, and run time. Although not all methods were assessed regarding run time, it is crucial to have values for this metric due to its use in streaming scenarios, where time and memory consumption are constrained.

\subsubsection{Clustering}
%\akst{Some} \ak{Two} 
%Two works approach the stream clustering task \citep{p3,p10}. Both papers presented similar approaches that use the artificial immune system (AIS) for text clustering. 
{\textit{Clustering} is a stream mining task in which the aim is to find intrinsic clusters, according to their features \citep{bifet2018machine}. The general idea is to minimize the similarity between different clusters and maximize the intra-cluster similarity \citep{bezerra2015data}. Differently from classification, in the clustering task, the labels are not available before the learning process. Therefore, alternative metrics are necessary, and since there is no ground truth, the learning process is named unsupervised \citep{bifet2018machine}. 

Three works approached the stream clustering task \citep{p3,p10,p40}. The first two papers presented similar approaches that use the artificial immune system (AIS) for text clustering, while the third presents textClust, a stream clustering method.} \citet{p3} developed a method for text stream clustering based on the AIS called AIS-Clus. It used heuristics based on the AIS to cluster data efficiently and, by discovering these clusters, can also detect concept drift and feature evolution. The authors could also recognize new classes corresponding to the concept evolution task in the experiments. According to the authors, the AIS is analogous to the biological immune system because it receives an intruder, clones specific cells, and handles the intruder until it dies. In their approach, for each new input (analogized as antigen), a scoring function calculates its adherence to each cluster (analogized as a B-cell). The clonal selection makes copies of clusters that undergo a mutation process. Later, the negative selection mechanism makes it possible to detect noisy data. Their method does not start from scratch, having an initial static phase for preprocessed historical data clustering. The other phase is online stream processing, which receives the clusters from the first phase as input. The authors used a survival factor for each word in an aging-like scheme. Although it works in a clustering scheme, the method is evaluated in terms of F1 score, accuracy, recall, and precision. More information is provided in \cite{p10}, which expands on \cite{p3}, and new experiments are executed. For example, to test the approach's capacity to handle new classes, the authors arranged data in three datasets to simulate the emergence of new classes/events, one of which included texts in Arabic. When AIS-Clus is compared to other methods, \ie, CluStream and DenStream, it achieves the best results regarding the precision, recall, and number of clusters, functioning as a classifier as described in \cite{p3}.

{\citet{p40} presented an online method for textual clustering, namely textClust. The algorithm is available within RiverML Python library \citep{montiel2021river}\footnote{https://riverml.xyz/0.19.0/api/cluster/TextClust/}. Over time, in the offline phase, the model was maintained concisely by merging similar clusters and removing outdated ones. A fading factor for the cluster weighting is used to determine cluster staleness. The method was evaluated in terms of homogeneity, completeness, and normalized mutual information (NMI). Homogeneity evaluates how well a clustering method assigns the data points to the clusters. Reaching 1 for homogeneity means that each cluster contains data points of a single class. On the other hand, completeness measures whether the data points of a given class were assigned to the same cluster. Reaching the value 1 for completeness means that the data points of each class were assigned to a single cluster. The authors support these statements by mentioning that ``completeness scores tend to be lower than the homogeneity scores'', and that it ``indicates that online clusters are quite pure with low entropy, but the topics are distributed over multiple clusters'' \citep{p40}.}

Most selected works that addressed a stream clustering task focused on short-text clustering. \citet{p23} proposed an efficient method for similarity-based short-text stream clustering called EStream. The method's efficiency comes from utilizing an inverted index to find the most similar clusters. The authors tested lexical (unigram, bigram, and biterm) and semantic text representations (using a pre-trained GloVe {\citep{pennington2014glove}}). Their method has two steps: the online and the offline phases. First, each cluster is lexically represented as a cluster feature 4-sized vector consisting of the features (in unigram, bigram, or biterm), their frequencies in the cluster, the number of texts in the cluster, and the cluster identifier. The semantic representation consists of the cluster vector and the cluster center, calculated from the average of the GloVe representation of the texts. EStream was compared in terms of NMI, homogeneity, and V-measure. EStream had the best performance in 50\% of the datasets used for evaluation. The authors highlighted that EStream requires less running time and that it stores more information than the other approaches, but that would be an acceptable trade-off \citep{p23}. They also highlighted that EStream might perform inadequately in more extensive texts.

\citet{p30} proposed a new method called GOWSeqStream, for short text stream clustering, using deep sequential methods, graph-of-words representation, and pre-trained word-embedding models. It uses subgraph mining to extract semantic information from the texts, although it lacks information on how to use it, considering even the number of sliding windows and the support. The method also utilized Word2Vec representations to generate embeddings to serve as input for other deep encoders, such as GRU. The author also experimented using bidirectional LSTM, Doc2Vec, and BERT representations. These representations were utilized as input for a DPMM. The method was compared with five approaches using three datasets; the proposed approach achieved the best values for two. The author also compared the representation generation; the best combination was with BERT and Bi-LSTM. In addition to English, the author used a Vietnamese text dataset as a final test. In this scenario, the proposed approach achieved the best results among the competitors. As in \cite{p23}, the authors used the NMI as the primary evaluation metric.

\citet{p24} proposed a new short text stream clustering method using an incremental word relation network. The authors highlighted their primary contribution as (a) a new method for real-time short text clustering using a bi-weighted relation: term frequency and co-occurrences, to overcome sparsity; (b) a fast method to locate core terms that represent text clusters sufficiently; (c) the mechanism to overcome topic drift, removing outdated relations and incrementally adding new terms and relations. In addition, the authors proposed a new data structure to represent the clusters, which they named \textit{cluster abstract}. This data structure had five fields: an index, the number of short texts in clusters, the sum of timestamps, the squared timestamps sums, and a new attribute compared to EWNStream (their previous approach) called \textit{pd}, containing a core term set. The method used data windows and specific calculations to update the model to add new data, exclude outdated data, and merge clusters. Besides, the method had a decay scheme to control the forgetfulness of old clusters. In essence, the method develops a graph containing terms and relations, and the clusters were obtained from groups of closely related words. The method searches for a cluster abstract with the most intersection of words considering the input data to predict a cluster to newly inputted data. Using a dataset crawled by themselves, the authors compared their proposed method against EWNStream, MStream, Sumblr, and Dynamic Topic Model. EWNStream+ outperformed its previous version (achieving roughly 86\% of NMI accuracy) and was approximately 30 percentage points better than MStream, the third in the ranking. In addition, the run time was very modest across different stream lengths.

\citet{p6} proposed two text stream clustering algorithms: (a) MStream, a one-pass clustering method that utilizes Dirichlet Multinomial Mixture Model (DPMM) and an update process per batch; and (b) MStreamF, which deletes outdated clusters, maintaining a concise model. Considering the clustering process of the MStream algorithm, there is the assumption that the new documents arrive sequentially, and each is processed only once. The initial document generates a new cluster, and subsequent documents choose one of the existing clusters or create a new one. The authors' updating process proves beneficial in the batch processing of text streams. The process was designed such that each document gets assigned and then temporarily deleted from the cluster so that the similarity of the other documents in the same batch is not impacted. After completing the batch process, all documents are assigned to their original cluster. For MStreamF, the authors developed a deleting scheme that works for batch processing by adding a new parameter $B_s$, which accounts for the number of batches. When the number of processed batches meets the $B_s$ parameter, the new batches are processed after the documents related to the oldest batch are deleted. As the iterations go by, it is expected that some clusters will become empty, indicating that they are outdated and could be deleted. The approaches were assessed in terms of NMI, run time, and number of clusters. They concluded that MStreamF is faster than MStream due to the conciseness of the former model. Comparing the proposed and the state-of-the-art models, MStream and MStreamF outperformed their competitors. MStreamF performed best with temporally ordered datasets, whereas MStream performed best with unordered datasets. The run time of all algorithms increased linearly with the size of the datasets, while the single-pass algorithms were faster.

In summary, the NMI, run time, and the number of clusters were the most often used metrics for stream clustering and short-text stream clustering. The latter may be considered a measure of conciseness, which directly corresponds to one of the constraints of streaming scenarios, \ie, memory consumption, and may indirectly impact run time. NMI, a Shannon-entropy-based metric, measures the similarity of two sets and, concerning clustering, the similarity of the ground-truth and the model-generated clusters \citep{p6,emmons2016analysis}. {Other metrics may appear, such as completeness and homogeneity. Those metrics vary between 0 and 1, where the higher, the better. Homogeneity evaluates how well a clustering method assigns the data points to the clusters. A perfect homogeneity, \ie, 1, indicates that each cluster contains data points of a single class. As aforementioned, completeness evaluates whether the data points of a given class were assigned to the same cluster. A perfect completeness value suggests that the data points of each class were assigned to a single cluster.}

\subsubsection{General detection}
In this category, we grouped papers that tackled \textit{event detection} and \textit{novelty detection}. According to \cite{faria2016novelty}, novelty detection is ``the ability to identify an unlabeled instance (...) that differs significantly from the known concepts''.
As suggested in \cite{faria2016novelty}, we considered \textit{concept drift detection}, \textit{semantic shift detection}, and \textit{concept evolution} as sub-categories of \textit{novelty detection}. {We separated this section to encompass approaches that focused on detection rather than incorporated detection methods in classifiers or clustering methods, for instance. } 

We also considered \textit{physical event detection} a sub-category of \textit{event detection}. As mentioned previously, \citet{p11}, \citet{p12}, and \citet{p14} described distinct aspects of a system for landslide detection. They utilized governmental reports as trustworthy sources and social media posts as social sensors (also named strong and weak signals, respectively). The system was described as fully autonomous and continuously evolving, becoming unnecessary human intervention. Although the works were similar in several aspects, there were minor variations in the evaluation metrics. \citet{p11} selected precision and F1 score metrics. The event detection was assessed using false positives and false negatives, where the original variant of the system was used as ground truth. \citet{p12} used F1 score, precision, recall, and the number of events detected as metrics. There was no ground truth regarding the number of events: only the events counted. \citet{p14} used accuracy to evaluate classifiers' {performance} across data windows.

\citet{p34} proposed a framework for real-time event detection using social media as a data source. The interesting highlights in this paper regard the tweets' enrichment for slang, abbreviations, and acronyms based on external sources. The method creates a local vocabulary using data from various external sources. In addition, the authors utilized spelling correction and emoticon replacement. The authors used an incremental clustering algorithm to cluster events and then rank these events based on important words for each event. %The framework itself seems not to be fully incremental, due to the processes that involve multiple passes over the data (e.g. )
The authors evaluated their method using two experiments: (a) comparing it to the General Social Media Feed Preprocessing Method (GSMFPM) to determine if the enrichment layer performs effectively; and (b) event detection from social media. In experiment (a), the authors represented the tweets using unigrams and bigrams, supposedly later converted to GloVe (unclear in the paper). Later, the vectors are applied as input to a Feedforward Neural Network (FNN) and a CNN. These approaches are not incremental, thus presenting concerns about the process' timeliness regarding real-time events. In this experiment, they measured the cross-entropy loss across the training epochs for both Twitter Sentiment Analysis and Naija datasets. Their method outperforms GSMFPM.
The second experiment measures accuracy over events in social media, using precision, recall, and F1 score. The authors used a dataset called Event2012, which contains annotations about events. The proposed method obtained a higher F1 score than the other approaches.

Regarding \textit{novelty detection} and its subdivision in this {study}, only one paper exclusively considers \textit{concept drift detection} \citep{p-25}. Three included the \textit{concept evolution} problem \citep{p10,p13,p15}, and another mentioned the \textit{semantic shift detection} \citep{p35}. Considering the \textit{concept drift detection},
\citet{p-25} used a cross-recurrence quantification analysis (CRQA) to detect concept drifts. The author's idea was to highlight the most significant hashtag-related events. Cross-recurrence quantification analysis was used to compare the changes in trajectory. This outcome is achieved by assessing the longest diagonal line of two consecutive windows and whether they follow the same generating process over time. All operations occurred inside a system called TSViz. The experiments discussed in the paper were on drift detection related to hashtags from Brazilian politics. The authors concluded that the drifts detected directly trace back facts from the news. According to the authors, recurrence analysis ``characterizes the behavior of dynamical systems by reconstructing produced data in phase spaces''. The authors used Normalized Compression Distance (NCD) to compute the similarity among texts and Naive Bayes to perform sentiment analysis; however, the authors did not detail the classification process. The results were visually assessed.

{Instead of providing a concept drift detector, \citet{zhang2024addressing} provided a framework for concept drift prediction. Although the approach focused on time series, the proposed framework is model-agnostic and monitors loss distribution drift to predict drift occurrence, which could also be interesting for text streaming scenarios. The method was evaluated in a prequential manner, \ie, train-then-test. When receiving new data, their framework makes a prediction with the model, updates the model $f_{\delta}$, stores (temporarily) the respective data $(x, y)$, stores loss $\mathcal{L}$ in $B$, and updates the memory bank $\mathcal{M}$. If the length of the $B$ is bigger than a predefined window size, and the z-score considering $B$ and the last window is bigger than a threshold, the fine-tuning process is triggered. The proposal also encompasses a parameter to control the frequency of fine-tuning, even if the aforementioned conditions are not met. Their method obtained the best performances in several of the tested scenarios.}

The \textit{concept evolution} problem regards the increase in the number of classes over time. For a model to be updated, it must internally account for these novel classes \citep{faria2016novelty}. Traditional ML methods require prior knowledge of the number of classes. \citet{p3}\citep{p10} proposed a method for text stream clustering based on AIS. These papers were previously referenced in this work. They also managed concept evolution (under the name of novelty detection). These methods addressed the concept evolution problem by cloning and mutating existing clusters, a heuristic of the clonal selection principle. If the novel data do not fit into a cluster, they are sent to the outlier buffer, where they are examined periodically to detect novel classes. \citet{p3} evaluated the quality of concept evolution handling using the $M_{new}$ metric, which measures the rate of novel class instances misclassified as from an existing class. In addition, the authors plotted the F1 score, accuracy, and recall over time, demonstrating the emergence of new classes and how their method recovers from concept evolution. The run time was not measured. \citet{p10} employed a similar plot as \citet{p3} for two datasets. In addition, they plotted the number of existing classes and identified classes by the method over time. The metric $M_{new}$ is also used, and the number of missed classes is computed.

\citet{p15} proposed ESACOD, a framework for streaming classification with concept evolution and subject to concept drift. Their work aimed to learn satisfying parametric Mahalanobis-based metrics in real time. According to the authors, the objective was to identify a feature space projection in which its constraints generate properties of cohesion and separation \citep{p15}.
%``the key idea is to identify a projection of feature space (...) where the constraints impose cohesion and separation properties''. 
Cohesion is the ability of data points to occur close to others from the same class. In contrast, separation is the ability of data points to be distant from others from different classes \citep{p15}. Their method trains an open-world classifier with a small dataset with an initial metric established. When new data arrives from the stream, the metric is applied to it, generating data in a new feature space, and the prediction is made afterward. If the prediction indicates that the data does not belong to a novel class, the prediction remains unchanged.
On the contrary, if the classifier assumes the data are from a potentially novel class, the data are added to a buffer. When this buffer is filled, it is checked for concept evolution and concept drift. An arbitrary percentage (between 0 and 30\%) of data with their respective labels is required. Finally, the evolution class metric is computed using paired constraints based on this randomly selected data.
%The pairwise constraints based on these randomly selected data are calculated to reckon the evolution class metric. 
Later, a k-means algorithm \citep{lloyd1982least} is applied, and a label propagation \citep{zhu2002learning} method is performed apparently to the other data in the buffer. If a concept drift or concept evolution is detected, a new classifier is trained with the data to replace the older classifier.
The authors concluded that their approach could address the challenges of multiple novel class detection and stream classification bound to concept drift and with few labels available. The method was evaluated in terms of accuracy and run time. Concerning concept evolution, the metrics used were $M_{new}$ and $F_{new}$, which measure the instances of an existing class misclassified as a novel class, $A_{new}$, which is the accuracy of novel class classification, and $A_{known}$, which is the accuracy of known class classification.

Regarding \textit{semantic shift detection}, \citet{p35} addressed this problem in an incremental way. The authors used incremental clustering techniques (such as affinity propagation) to generate representation clusters in time slices. The word contexts in the past were clustered into several clusters, serving as a memory for posterior observations. To generate representations, the authors tested BERT and Doc2Vec. BERT provided contextual representation, whereas Doc2Vec provided pseudo-contextual embeddings. The approach selected documents in which target words emerged, fine-tuned the embedding model to add new arriving documents, extracted the embeddings, clustered the representations, and refined the clusters by removing clusters of single or old representations. The authors tested their approach using representations generated by BERT and Doc2Vec for two datasets from SemEval 2020: CCOHA and LatinISE. The authors evaluated alternatives based on affinity propagation. The incremental version of the affinity propagation (IAPNA) performed adequately on the LatinISE dataset using BERT representations and on the English dataset using Doc2Vec representations. In contrast, the affinity propagation a posteriori had satisfying results in the opposite situations. The authors were surprised that Doc2Vec obtained decent results and consumed less time than contextual models.
%``performed well while being smaller and faster than contextual models''. 
%It is important to notice that the target word must be known in advance to perform analyses.

{\citet{castano2024incremental} also proposed a variation of the affinity propagation algorithm named APP. The authors evaluated their method against affinity propagation, IAPNA, and used the Iris, Wine, Car, and KDD-CUP datasets. The methods were assessed in terms of purity and NMI, a frequent metric in clustering settings. For the semantic shift detection task, the authors provided a thorough case study based on a diachronic corpus of Vatican publications in Italian containing around 29,000 documents, split into six subcorpora. From the corpora, texts written by Pope John Paul II were removed due to the variety and richness of his documents, according to the authors. Tracking a previously selected word, \ie, \textit{novità} (novelty), the authors could find the use of this word in a negative sense (in the first subcorpora) to its use in the context of innovation in the Catholic Church. }

{Although not directed to text stream scenarios, \citet{ishihara2022semantic} proposed a metric for semantic shift named semantic shift stability, improving decision-making on when to fine-tune a model. This method consisted of creating word embeddings, setting anchor words, introducing a rotation matrix, and calculating the stability. The stability was calculated using the cosine similarity between words in two rotated matrices. \review{Also majorly unrelated to text stream scenarios, \citet{periti2024towards} extended the problem of semantic shift detection. According to the authors, frequently semantic shift detection in batch scenarios is addressed by considering two periods of reference. \citet{periti2024towards} proposed five methods for tracking semantic shift, considering consecutive time intervals, consecutive time periods, clustering over all time periods, incremental clustering over time periods, and scaling up form-based approaches. From the proposed methods, only incremental clustering over time seems to be suitable for text stream scenarios}. Other papers also developed methods for detecting semantic shifts or adapting to them in batch scenarios, \eg, \citet{kim2014temporal,hofmann2020dynamic,liu2021statistically}, to mention a few. Since we are interested in text stream scenarios, we mentioned these papers because they can inspire the development of incremental versions that are suitable for text stream learning, considering its constraints presented in Section \ref{sec:dsm}.}

\subsubsection{Topic modeling}
Topic modeling consists of statistical tools to examine textual data and identify the most relevant terms related to each theme. This approach facilitates the exploration of the interconnections among these themes and their temporal evolution \citep{blei2012probabilistic}. It is also considered a text mining task \citep{kherwa2019topic}. Four selected papers approach \textit{topic modeling}  \citep{p1,p7,p26,p27}.

\citet{p1} proposed an approach mixing LDA and ADWIN to overcome the problem of topic modeling in document streams, entitled AWILDA. LDA \citep{blei2003latent} is a common method for topic modeling. The authors mentioned that LDA had gained much attention, and it also has an online version. However, one problem with the online version is setting window sizes because drifts may happen in a smaller period than the window size. Thus, the authors defined the window with the aid of an ADWIN module, which can assist in determining topic drifts and the new window for LDA to consider. Two classes of algorithms were mentioned: the passive, which updates a model for each observation, and the active algorithms, which attempt to detect the drift and update the model only when the drift is detected. We can draw parallels between these classes of algorithms and the detection methods presented in Section \ref{subsec:td-detection}, \ie, \textit{adaptive} and \textit{explicit}, respectively. The author's idea was to separate the task of topic modeling and topic drift detection.
There are two LDA models inside AWILDA: one for language modeling ($LDA_m$) and the other for drift detection ($LDA_d$). In this approach, for each document received from the stream, AWILDA reckons the likelihood for $LDA_d$ and adds it to the ADWIN module. If a drift is detected, $LDA_m$ is trained on the subwindow ADWIN selects. $LDA_m$ is updated whenever a new document arrives from the stream. The authors evaluated their proposed method using the perplexity metric for document modeling and the latency between the actual current drift and the detection. According to the authors, perplexity is ``used by default in language modeling to measure the generalization capacity of a model on new data'' \citep{p1}. The authors concluded that AWILDA could recognize all drifts in the synthetic datasets and one version of the real-world dataset. In addition, the method can select the documents window to be used for updating. AWILDA can detect abrupt drifts and works sufficiently for gradual drifts. Compared to online LDA, it worked similarly until a drift occurred. When a drift occurs, AWILDA is retrained, which increases perplexity, but it ultimately outperforms the online LDA.

\citet{p7} proposed a short text stream classification method that uses content expansion and includes a concept drift detector. According to the paper, the external sources must satisfy two criteria: to be large and sufficiently rich to comprise most contents in the short text stream that will be classified and highly topic-consistent with the text stream. The method mines hidden information from the external corpus by using LDA because, according to the authors, LDA performs adequately on longer texts. From the LDA model, top representative words for the topics are selected to be added (once or several) times to a short text according to the topic distribution and word probability of belonging to a topic. The topic distribution represents each short text. The method was evaluated regarding accuracy (classification task) and the drifts, using false alarms, missing drifts, and delays. The datasets were arranged to simulate drift; however, the method was unspecified. The authors concluded that their approach surpassed the accuracy of all the competitors, demonstrating more stability. In addition, their approach could recover from drift earlier than other approaches and outperformed the competitors regarding delay and missing drifts.

\citet{p26} proposed a graph convolutional method for topic modeling, considering short and noisy text streams. The authors leveraged Word2Vec representations and Wordnet knowledge graph to improve the predictions of their method, called GCTM. The authors claimed that their method could balance the knowledge graph and the knowledge obtained from the previous data batch. This ability can be valuable when handling concept drift. GCTM integrates a graph convolutional network (GCN) into an LDA model to exploit a knowledge graph, and both are updated simultaneously in the streaming environment. 
%The paper is heavy in terms of statistics. 
The authors tested their approach using six short text datasets and two regular text datasets. Using previous knowledge allowed GCTM to output satisfying predictions and recover more quickly from concept drift. The authors simulated concept drift by rearranging the topics sequentially. The metrics selected for evaluation were the Log Predictive Probability (LPP) \citep{hoffman2013stochastic} and the Normalized Pointwise Mutual Information (NPMI) \citep{lau2014machine}. These methods measure the model generalization and the coherence of the topics, respectively. GCTM was evaluated in two ways: utilizing Word2Vec (GCTM-W2V) and the knowledge from the Wordnet graph (GCTM-WN). GCTM-WN and GCTM-W2V outperformed the competitors in LPP across all the datasets, even in the presence of concept drift. The authors also performed an ablation study.

\citet{p27} proposed an LDA-based topic modeling approach with mechanisms for balancing stability and plasticity, namely BSP. Stability-plasticity is a dilemma involving maintaining old knowledge (stability) and learning new knowledge (plasticity) \citep{p27,gama2014survey}. Balancing both prevents concept drift from impacting performance and catastrophic forgetting \citep{p27}. The authors used TPS and iDropout combined into an LDA-based topic modeling method. TPS \citep{tran2021dynamic} aided the model with external knowledge, \ie, Word2Vec representations.
%(later said it is Word2Vec representations). 
iDropout \citep{nguyen2019infinite} created variables $\beta^t$, updated whenever a new mini-batch is inputted. Because both are different mechanisms, the authors modified the calculation of $\beta$ to comprise information from both mechanisms. They performed experiments on eight datasets: one long, two regular, and five short-text. The authors compared their method to six different approaches. The hyperparameters were selected using a grid search. Similar to \citet{p26}, the authors contrasted LPP and NPMI. The authors tested using the datasets shuffled and ordered chronologically whenever possible. Their method achieved the best values for LPP in four out of six datasets tested. It is worth noting that their method achieved satisfactory results very rapidly at the highest levels. Their method maintained high levels of performance while using chronological datasets. The authors tested the stability and plasticity by simulating drifts by sorting the topics in order of classes, similarly to \citet{p26}. BSP could reach the best values when testing for catastrophic forgetting and maintained the highest levels when recovering from concept drift. As in \cite{p26}, the authors performed an ablation study to understand the impact of some parameters.

\subsection{Text Representation Methods}
\label{subsec:td-text-representation}

%\JPBCOMMENT{Segundo o que discutimos na reuniao, adicione aqui uma breve descrição do que é cada tipo de representação. Tente ser o mais formal para isso, preferencialmente usando formatações já existentes em livros/artigos}
%\CGCOMMENT{Coloquei informações logo abaixo da figura. Veja se atende, por favor.}

\begin{figure}[!htp]
\includegraphics[width=0.45\textwidth]{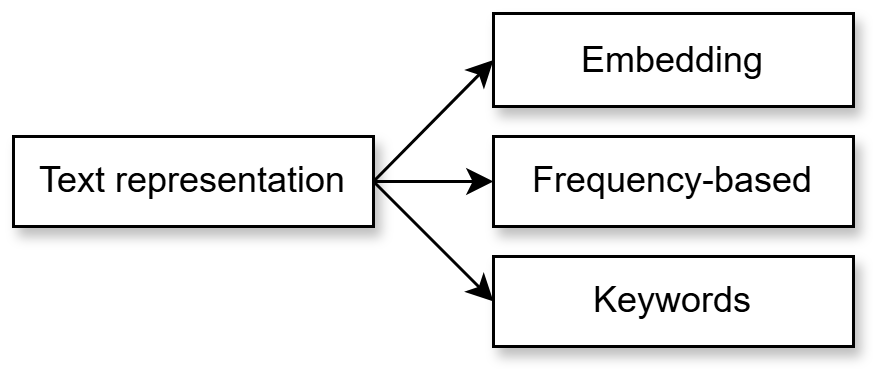}
\centering

\caption{Categories of text representation methods used in the papers.}
\label{fig:text-drift-text-representation}
\end{figure}

This subsection describes the text representation methods used in the 
selected papers{, aiming at answering the Research Question 3 (RQ3), \ie, \textit{``Which type of token/word/sentence representation is used in the study?''}. Besides collecting the aforementioned types, we also aimed to check how and if they are updated (see Section \ref{subsec:text-representation-upd-scheme})}. Fig.~\ref{fig:text-drift-text-representation} depicts the three main categories: (i) Embedding-based methods, such as Word2Vec and BERT; (ii) Frequency-based methods, which include Bag-of-Words, TF-IDF; and (iii) {Keywords}. 

{In our categorization, we considered \textit{embeddings} the dense vectors generally generated by neural-based approaches, such as Word2Vec, BERT, or even large language models; therefore, in this case, these language models are used as feature extractors \cite{tunstall2022natural}. These vectors are capable of representing word semantics and capturing the connotation of words \cite{tunstall2022natural}. \textit{Frequency-based} methods are those that resort to methods that leverage word counts and derive text representations. Sometimes, the word counts are used directly as a vector representation, \eg, Bag-of-words, or used as a means to calculate word importance, such as in TF-IDF \cite{hapke2019natural}. In both cases, they are generally used in a structured way because most machine learning methods are not able to handle variable-length input vectors. The category \textit{Keywords} regards the use of words themselves without resorting to vector representations, therefore maintaining a list of keywords to represent items. }

Table \ref{tab:text-rep} lists the text representation methods used across the papers. Seven approaches were categorized as \textit{frequency-based}, six as \textit{embedding}, and one as \textit{words}. Two papers have not provided the text representation method, while one provided \textit{file compression}, which cannot be directly classified among the categories but could adapt to \textit{words} because the compression is performed over a file containing a set of words. \citet{p-25} used \textit{file compression} and calculated text similarity by using a formula that considers the sizes of the zipped file containing the two texts and the zipped files containing each of the given texts separately.
%by measuring the difference between the sizes of the zipped files with the texts, separately. 
As mentioned, this method is named NCD \citep{cilibrasi2005clustering}. Although it appears reasonable for files and images, NCD is also used for texts in some works. For example, NCD is listed as a similarity metric in structured data \citep{ontanon2020overview}, and in texts \citep{pradhan2015review}, even in the presence of noise \citep{cebrian2007normalized}. At first sight, it may appear unreasonable because two different files containing distinct texts may result in similar file sizes. However, according to \cite{ontanon2020overview},
if two given files are similar, compressing them together results in an approximate file size to compressing only one. %considering two given files, if they are similar, compressing both together would result in an approximate file size to compressing only one of them. 
Thus, NCD calculation utilizes this aspect to determine the similarity between two files containing raw text.

%\JPBCOMMENT{Olhando essa taxonomia e a tabela, lembrei de Hashing Tricks, que não se encaixam em nenhuma dessas categorias. Além disso, existe uma ferramenta, Vowpal Wabbit, que oferece formas de tratar texto em formato streaming usando isso (\url{https://github.com/VowpalWabbit/vowpal_wabbit}). Também a gente pudesse comentar no texto que reconhecemos a existência dessas abordagens, mas elas não apareceram como resultado da query.}
%\CGCOMMENT{Adicionei o texto abaixo.}

{Although it did not appear across the studied papers, we acknowledge the existence of other incremental/online methods, such as Hashing Tricks \citep{attenberg2009collaborative}. Beginning from a zero-filled representation vector, Hashing Tricks leverage hash functions to convert tokens into hash values. Each hash value is then divided by the length of a pre-defined representation vector, and the result of the modulo operation defines the index of the representation vector to have its value increased. Although it has no learning at all, it performed competitively in stream scenarios \citep{thuma2023benchmarking}. This method is available in tools such as Vowpal Wabbit \cite{shi2009hash}\footnote{Available at \url{https://github.com/VowpalWabbit/}.}.}

\begin{table}[!htp]
\caption{Text representation used in the studied papers.}
%\rowcolors{2}{gray!25}{white}
\resizebox{.9\linewidth}{!}{ 
\begin{tabular}{lll}
\hline
\textbf{Text representation method} & \textbf{Papers}  & \textbf{Category} \\ \hline
Bag-of-words \citep{harris1954distributional}   & \citep{p4}\citep{p5}\citep{p13}\citep{p16}\citep{p19}\citep{p20}\citep{p21}\citep{p23}  & Frequency-based  \\
                                                & \citep{p28}\citep{p33}{\citep{p40}}{\citep{p44}}   &   \\
\rowcolor{gray!25} BERT \citep{devlin2018bert}                     & \citep{p20}\citep{p22}\citep{p30}\citep{p35}{\citep{p36}}{\citep{p38}}{\citep{p48}}  & Embedding  \\
Bigram                                          & \citep{p23}{\citep{p40}}  & Frequency-based  \\
%\rowcolor{gray!25}BiLSTM                                          & \citep{p30}  & Embedding  \\
\rowcolor{gray!25}Biterm                                          & \citep{p7}\citep{p23}  & Frequency-based  \\
Co-occurrences                                  & \citep{p24}  & Frequency-based  \\
\rowcolor{gray!25}Doc2Vec \citep{le2014distributed}               & \citep{p30}\citep{p35}  & Embedding  \\
FastText \citep{bojanowski2016enriching}        & \citep{p16}  & Embedding  \\
%File compression  & \citep{p-25}  & -  \\
\rowcolor{gray!25}GloVe \citep{pennington2014glove}               & \citep{p11}\citep{p16}\citep{p23}\citep{p27}\citep{p34}  & Embedding  \\
Graph-of-words                                  & \citep{p24}\citep{p26}\citep{p30}{\citep{p42}}  & -  \\
%GRU \citep{cho2014properties}                   & \citep{p30}  & Embedding  \\
\rowcolor{gray!25}Incremental Word Context                        & \citep{p32}{\citep{p41}}  & Frequency-based  \\
%N/A  & \citep{p12}, \cite{p29}  & -  \\
% Own embedding method  & \citep{p26}  & Embedding  \\
{PSDVec \citep{li2017psdvec}}  & {\citep{p45}}  & {Embedding}  \\
\rowcolor{gray!25}Sent2Vec \citep{moghadasi2020sent2vec}          & \citep{p34}  & Embedding  \\
TF-IDF \citep{salton1988term}                   & \citep{p2}\citep{p4}\citep{p18}\citep{p20}\citep{p28}\citep{p33}{\citep{p37}}{\citep{p39}} & Frequency-based  \\
                                                & {\citep{p40}}{\citep{p46}} &   \\
%TF-IDF + PCA  & \citep{p17}  & Frequency-based  \\
\rowcolor{gray!25}Word2Vec \citep{mikolov2013distributed}         & \citep{p9}\citep{p11}\citep{p14}\citep{p15}\citep{p16}\citep{p18}\citep{p26}\citep{p27} & Embedding  \\
\rowcolor{gray!25}                                                & \citep{p28}\citep{p30}{\citep{p36}}{\citep{p48}} &   \\
Word frequency                                  & \citep{p6}\citep{p24}  & Frequency-based  \\
\rowcolor{gray!25}Words                                           & \citep{p1}\citep{p3}\citep{p8}\citep{p10}{\citep{p41}\citep{p42}\citep{p43}}  & {Keyw}ords \\ \hline
\end{tabular}
}
\label{tab:text-rep}
\end{table}

Regarding the representations, several papers used more than one method, sometimes combined, \eg, Bag-of-words + TF-IDF. However, they were divided in Table \ref{tab:text-rep}. In addition, Word2Vec and Bag-of-words (BOW) were used in {12} papers and TF-IDF in {10} papers. Finally, words were used directly in {seven} papers as a representation method. We briefly described the methods as follows, considering the chronological order of each method.

\subsubsection{Bigram} Bigram adheres to the Bag-of-words concept, in which it is possible to organize texts in two dimensions: columns as words and rows corresponding to documents. The cells contain the count of a given word in a specific document. The difference is that a pair of sequential words is represented in each column instead of the words. For example, the sentence ``he has been here'' will generate three columns: (he, has), (has, been), and (been, here). The challenge incurred from utilizing bag-of-words in streaming scenarios also happens to bigrams, \ie, the dimensions regard fixed words and do not evolve. \citet{p23} used three representation methods while testing their proposed method for short-text stream clustering: unigram, \ie, bag-of-words, bigram, and biterm. {\citet{p40} also used both unigram and bigram for text representations.}

\subsubsection{Biterm}
According to \citet{p7}, a \textit{biterm} corresponds to unordered word-pair co-occurrences. Furthermore, \citet{p7} highlighted that biterms were more sparse than regular bag-of-words and utilized external sources to reduce the sparseness. Considering the biterm definition and using the same example as in a bigram, the biterms generated from the sentence ``he has been here'' would be (he, has), (he, been), (he, here), (has, been), (has, here), and (been, here). Considering the text stream scenario, it encounters challenges similar to those of bag-of-words and bigrams.
%As expected, it also incurs the same problems as bag-of-words, considering the text stream scenario. 
To overcome this, \citet{p7} developed an ensemble based on base learners trained using data chunks, each with its biterm topic model. \citet{p23} also used biterm as text representations. To evaluate their short-text stream clustering method, the authors used biterm, unigram, and bigrams. Biterms performed better than bigrams and unigrams, considering NMI values.

\subsubsection{Co-occurrences}
Co-occurrences count simultaneous occurrences of two particular words. \citet{p24} developed a bi-weighted word relation network that considers both the co-occurrences and the word frequencies. Although co-occurrences and word frequencies are not representations, we opted to include them as a single representation because they will be part of a graph, \ie, graph-of-words, which is an actual representation.

\subsubsection{Graph-of-words}
Graph-of-words (GOW) is a textual representation that transforms documents into graph-based structures
%``supports to transform given documents to graph-based structures''
\citep{p30}. According to the author, it can maintain long-term relationships between words. After generating the graphs regarding specific documents, frequent subgraph mining techniques were applied, and later, the mined frequent subgraphs were used as feature representations. In \cite{p30}, GOW had two parameters: sliding window and minimum support. GOW appears to have the capability of being updated in real time. However, its use with a pre-trained Word2Vec model (that can be outdated after an arbitrary period) made the process not fully incremental. Although they did not use the terminology \textit{graph-of-words}, \citet{p24} developed a corpus-level word relation network, namely EWNStream+, which retained the co-occurrence counts and word frequencies. According to the authors, EWNStream+ is incremental by receiving data batches. \citet{p26} 
proposed a novel graph convolutional topic model (GCTM) based on graph convolutional networks and LDA. The initial graph was formed using words and their relations. GCTM was tested using Word2Vec representations and WordNet. GCTM did not support incremental-fashioned training, implying that the text models could become obsolete.

\subsubsection{Word frequency}
Word frequency is the word count. 
\citet{p24} included word frequency as part of their word relation network, which also considered the word co-occurrences. This representation was also used to determine whether a word was outdated in the graph representation.

\subsubsection{{Key}words}
Several papers chose to use the words themselves rather than any text representation. {In this case, since the words are not structured as in a bag-of-words representation, for example, we named it \textit{keywords}.}
%It is common to choose to directly use words when using LDA-based methods.
\citet{p1} presented AWILDA, an LDA-based method integrated with ADWIN for topic drift detection. The authors used the {key}words lowercased and stemmed. \citet{p3}\citep{p10} described AIS-Clus, an incremental clustering method. Initially, the authors used DBSCAN \citep{ester1996density} to generate the cluster, and then sketches were developed to summarize each cluster. The sketches contained lists of {key}words and outliers present in a cluster. \citet{p8} presented a method to handle concept drift in an abruptly changing environment. The authors used {key}words to monitor probabilities in topics. Considering the updating scheme, {key}words could easily be added or removed from sketches. Therefore, we considered it possible to use it in streaming scenarios, although it could become complex and time-consuming to maintain a list of {keywords} in every sketch, as demonstrated in \citet{p1} and \citet{p10}, if not limited to respecting the constraints of data stream environments.

% ALEKOE 30-9

\subsubsection{Bag-of-words} Bag-of-words \citep{harris1954distributional} is probably one of the simplest methods for text vectorization, as it divides the text into tokens or words. Considering rows and columns, these tokens function as columns while the rows represent each text, such as tweets. There will be the counts of the tokens corresponding to a particular column in a text corresponding to a specified row in each cell. An evident characteristic is that bag-of-words representation in a unigram way does not represent the order of words, which can be leveraged in some applications. In streaming scenarios, it inhibits ML methods from performing properly. For example, suppose a bag-of-words representation is generated whenever each new text is inputted. In that case, the number of columns may increase, and most ML methods cannot handle dimension-changing inputs. Furthermore, even if the process runs in batches, the words of the bag-of-words may change. If the first batch defines the words for the bag-of-words representation, it may not recognize changes and new words, \ie, new dimensions, over time.

\subsubsection{TF-IDF}
Term-Frequency-Inverse Document Frequency (TF-IDF) is a statistic from the information retrieval area used for determining the importance of words to a document or a set of documents \citep{salton1988term}. The calculation considers the frequency of a term and the inverse document frequency, which defines how informative a term is across several documents. Generally, TF-IDF is used in the stream setting to encode data batches. It is worth noting that the term frequency calculation is remarkably similar to the bag-of-words procedure. Thus, it is common to discover the use of bag-of-words with TF-IDF. \citet{p2}, \citet{p20}, and \citet{p33} used TF-IDF after obtaining a data batch to encode the terms and the texts from the stream. \citet{p4} utilized TF-IDF to generate vector representations from the data batches so that a base learner could be trained and incorporated into the ensemble. \citet{p18} and \citet{p28} performed TF-IDF in an offline mode to generate a very high-dimensional vector so that they could test their dimensionality reduction strategy. \citet{p17} executed TF-IDF before all the processing. Later, the authors employed PCA to reduce the dimensionality of the datasets by selecting the 10,000 most meaningful components. Since TF-IDF works together with bag-of-words, it is impossible to update it {incrementally without changing the number of dimensions}. {\citet{p40} used TF-IDF to decide the proximity of incoming text to existing microclusters in the online phase. This calculation is also used in the offline phase, particularly when evaluating the merging of existing clusters. \citet{p37} used TF-IDF representations to generate the fuzzy lattice structure. \citet{p39} leveraged TF-IDF to compute the input vectors to train base learners. An interesting aspect regards preprocessing in \cite{p39}: the authors utilized the Stanford CoreNLP \citep{manning2014stanford} to segment words, part-of-speech tagging, and stemming. The authors used only the first three tags of noun, verb, and adjective. According to the authors, these tags ``carry the most valuable information regarding reviewed items''. However, no evidence is provided.}

\subsubsection{Word2Vec} Word2Vec \citep{mikolov2013efficient} corresponds to two distinct model architectures for learning distributed representations: Continuous Bag-of-words (CBOW) and Skip-gram. Both are neural network architectures, where the number of neurons is the same in the input and output layers, and the single hidden layer corresponds to the embedding size. Each neuron in the input and output layers can correlate to the words in the vocabulary. The representations, after training, are often obtained by taking the connection weights between a neuron (representing a word) in the output and the hidden layers. The difference between CBOW and Skip-gram is the training step aim: CBOW aims at predicting a specific word given its surrounding words, whereas Skip-gram does the opposite, \ie, predict the word in the middle based on the surrounding words \citep{mikolov2013efficient}. The papers that utilized Word2Vec used it for text representation only. 
{\citet{p36} leveraged Word2Vec for reduction of data sparsity. The authors developed their method for short-text classification, and one of the general approaches for this problem was to enrich the data. The authors evaluated both Word2Vec and BERT for the short-text representation, which was later applied to a CNN to extract higher-level feature information. }
Although Word2Vec is a neural architecture, it has incremental versions by using gensim\footnote{https://radimrehurek.com/gensim/} \citep{rehurek2011gensim} or other methods in the literature \citep{kaji2017incremental,may2017streaming,iturra2023rivertext}.

\subsubsection{Doc2Vec}
\citet{le2014distributed} proposed Doc2Vec to
obtain documents as distributional vectors. Doc2Vec is a generalization of Word2Vec. Similarly to Word2Vec, Doc2Vec is constituted by two architectures: Paragraph Vector - Distributed Memory (PV-DM) and Distributed bag-of-words version of Paragraph Vector (PV-DBOW). In PV-DM, the document vectors are trained with the word vectors in the architectures, while in PV-DBOW, the aim is to predict the words of a document from a document ID. \citet{p35} used a Doc2Vec model trained with the CCOHA and LatinISE datasets. The model was not updated during the process and may become obsolete as time passes. It was unclear whether \citet{p30} utilized a pre-trained model, trained a model from scratch, or if the model was updated over time. Since Doc2Vec is a neural architecture, training and updating it can be computationally costly.

\subsubsection{GloVe} Global Vectors (GloVe) is a method for generating co-occurrence-based word vector representations \citep{pennington2014glove}. According to the authors, GloVe utilizes global matrix factorization and local context window methods. The method is trained in a batch manner. \citet{p16}, \citet{p23}, and \citet{p27} used GloVe for semantic representation by using pre-trained models. \citet{p34} claimed that GloVe is used for feature extraction. \citet{p11} mentioned that the proposed system, \ie, Adaptive Social Sensor Event Detection (ASSED), supports GloVe. The authors in the original paper \citep{pennington2014glove} did not describe any incremental or adaptive training. Therefore, the vector representations can become outdated over time, constituting a potential disadvantage in streaming scenarios.

\subsubsection{FastText}
FastText \cite{bojanowski2016enriching}, an extension of the Skip-gram method, is one of the Word2Vec architectures. Instead of accounting for the entire words, FastText considers subword partitions using n-gram vectors. Using an example from the original paper, encoding the word $where$ in a 3-gram fashion results in a 5-sized vector containing (wh, whe, her, ere, re). In addition, the approach incorporates the word $where$ integrally. This method of splitting words in n-grams helps the model handle words unseen in the training step, also named out-of-vocabulary (OOV) words. An incremental update method is not mentioned in the paper. \citet{p16} utilized a pre-trained FastText model \citep{bojanowski2016enriching} as a text encoding method. In addition, FastText is used statically, implying that no method is presented in \citet{p16} for the incremental update of the text representations. However, \citet{p16} concatenated FastText representations to \textit{bag-of-words} representations generated in each step of an incremental procedure of accumulating data from past events.

\subsubsection{BERT}
Bidirectional Encoder Representation from Transformers (BERT) is a multi-purpose language model that enables several NLP tasks \citep{devlin2018bert}, such as sentiment analysis, sequence-to-sequence, paraphrasing, and question answering. In addition, BERT can provide vector representations of text to be used in a particular downstream task. \citet{p20} used an Italian version of the pre-trained BERT model, \ie, AlBERTo \citep{polignano2019alberto}, for measuring semantic similarity between tweets. In \cite{p22}, BERT was the primary model. The authors tested different sampling methods for fine-tuning to pursue an incremental update of the model. BERT was also used as a text encoding method in \cite{p30}, where the authors enhanced short-text clustering by combining pre-trained BERT's representations with a BiLSTM and a graph-of-words representation. \citet{p35} used BERT for word representation generation in both English and Latin by using pre-trained models. Considering the aforementioned papers, only \citet{p22} had an updating scheme for the representations. It was achieved by using fine-tuning strategies, which could enable the use of BERT in streaming scenarios, but it may also become a bottleneck in the process. {\citet{p38} leveraged BERT and variations in two moments. First, the authors used a pre-trained RoBERTa model \citep{liu2019roberta} specifically suited for sentiment classification. The RoBERTa model enabled automated training data generation. However, another BERT model was fine-tuned in the system whenever a sentiment drift happened. \citet{p36} used BERT to enrich short texts. Short texts are very sparse, and, according to the authors, using embeddings may improve the representation quality.}

\subsubsection{Sent2Vec} 
\citet{moghadasi2020sent2vec} proposed a sentence embedding method that considers the sentiment score behind the sentence. \citet{p34} used the Sent2Vec embeddings to compute the semantic representation of the input texts and then cluster these texts. If a new tweet was different from the histograms of the clusters, a concept drift was deemed to have occurred, and a new cluster was created for it. \citet{p34} did not describe an updating scheme. Thus, the Sent2Vec model can become obsolete over time, necessitating retraining.

\subsubsection{Incremental Word Context}
\citet{p32} proposed a vector representation method for texts %\akst{. This} \ak{that}
that can be considered a table-like representation, with the columns corresponding to words and rows similarly corresponding to words. However, the column (in the original paper, called context) and the words (called vocabulary) can have different sizes. The number of contexts defines the dimension size of the vector representation. The authors calculated the positive pointwise mutual information (PPMI) in each cell, considering the words and their co-occurrences. Although the vocabulary (rows) can be updated, similarly to bag-of-words, if the contexts are fixed, the system may incur obsolescence after the context words decrease or stop appearing. Furthermore, if certain context words are exchanged with other words, the changed dimensions will not represent the same contexts, and this will be reflected in an ML model dependent on vector inputs.

{\subsubsection{PSDVec} \citet{li2017psdvec} proposed the Positive-Semidefinite Vectors (PSDVec) as a toolbox for incremental word embedding. PSDVec is an eigendecomposition-based method. Similarly to Incremental Word Context, PSDVec uses a pointwise mutual information matrix. According to the authors, PSDVec has several advantages, including the ability to learn new words incrementally based on an original vocabulary. In their experiments, \citet{li2017psdvec} reached good results in the word similarity and analogy tasks.}

In this subsection, we analyzed the text representation methods used in the selected papers. However, we did not extrapolate the same analyses to incremental versions. Thus, when we discussed that a particular method only worked at least in batches, we did not extend the same conclusions to other versions, including incremental/adaptive versions when available. {Although not listed among the text representation methods found across the selected works, recently studied alternatives that could enable concept drift detection can be encountered in the literature, such as lexical replacements \cite{periti2024analyzing}, word senses representations \cite{giulianelli2023interpretable}, and the use of large language models (LLMs) for topic modeling \cite{mu2024large}.}

\subsection{Updating Mechanism of Text Representation Methods}
\label{subsec:text-representation-upd-scheme}

\begin{figure}[!htp]
\includegraphics[width=0.8\textwidth]{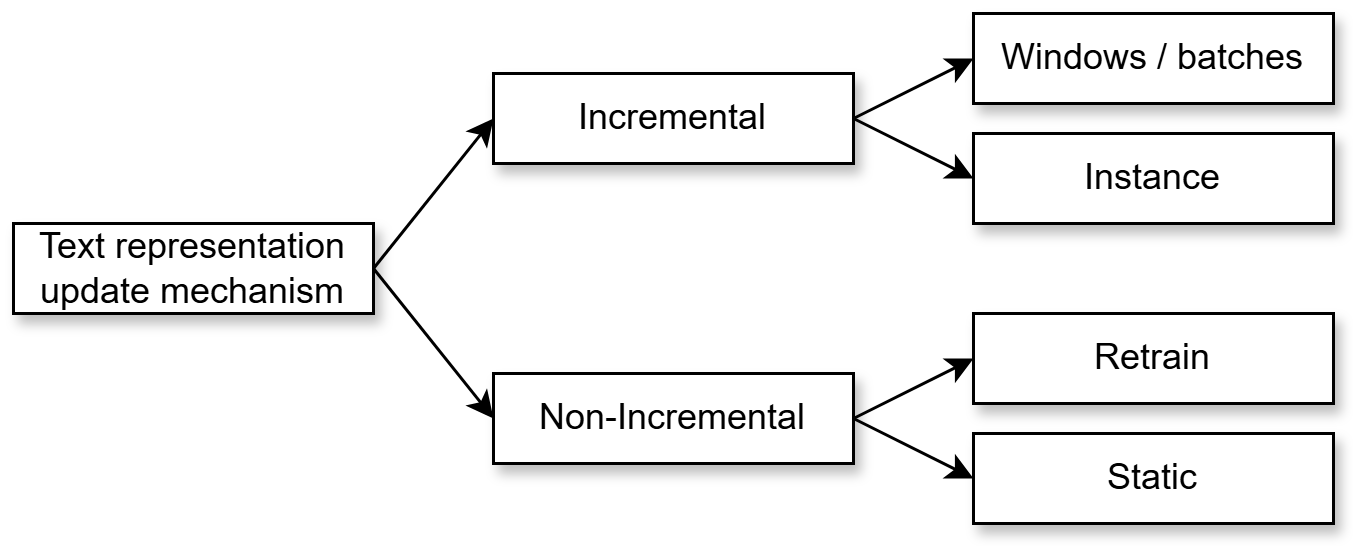}
\centering
\caption{Categories of mechanisms for text representation updating found in the selected papers.}
\label{fig:text-drift-text-representation-update}
\end{figure}

We also considered the updating mechanism of the text representation methods. Observing how the text representation behaves over time in text stream scenarios is critical. Because of stream characteristics, \ie, fast and potentially infinite, a static text model is a problem. It is even severe in text stream scenarios under concept drift because a representation vector may become obsolete, losing quality and, thus, negatively impacting the stream mining task. Therefore, we also obtained information on the text representation updating method. Fig.~\ref{fig:text-drift-text-representation-update} depicts the organization regarding the updating scheme of text representation methods. We organized in two dimensions: \textit{incremental} and \textit{non-incremental}. In \textit{Incremental}, we considered that the representation method can be updated over time, whether in batches or instances. In \textit{Non-Incremental}, we assumed that the text representation method was either static during the entire process or required complete retraining to be updated.

\subsubsection{Incremental}
We list text representation methods with incremental update capabilities organized in \textit{windows/batches} or \textit{instance}. Considering the update in windows/batches, this indicates that the text representation method requires a new amount of data to either be worth updating or satisfy a specific constraint of the text representation method. Using BERT as in \cite{p22} {and \cite{p38} are examples of this category. \citet{p22}, t}he BERT model is fine-tuned using texts selected by the sample methods proposed by the authors. {\citet{p38} perform the fine-tuning through an updated training set. The training set is updated whenever a sentiment drift is deemed to have occurred.}

Considering the incremental methods that can be updated in instances, it implies that it is unnecessary to accumulate data to update the text representation method: a single piece of information can be used for that. For example, we mention Incremental Word Context \citep{p32}. Furthermore, given a single new input, the Graph-of-Words \citep{p24} can be updated in real time.

\subsubsection{Non-Incremental}
Considering the text representation methods that do not allow any update but are retrained from scratch, we list bag-of-words, bigrams, biterm, and TF-IDF. However, while in use, a few text representation methods were kept static in the text streams: FastText, % BiLSTM, GRU, 
Doc2Vec, and GloVe. Most were used as pre-trained models, and they can become obsolete after some time, demanding complete retraining to maintain the performance of the dependent ML model. {\citet{p28}, \citet{p30} and \citet{p36} also leveraged static BERT and Word2Vec models.}

\section{Datasets}
\label{sec:datasets}
{Recalling the Research Question 4 (RQ4), \ie, \textit{``Which datasets were used to evaluate the proposed approach(es)?''}, w}e also included a list of real-world datasets to which the methods for stream mining tasks from the selected papers were applied. The synthetic datasets were excluded since they are generally numeric or contain a sequence of unrecognizable topics. 
%Even in cases where the synthetic datasets are not numeric, they may not represent reality, containing unrealistic texts or improbable topic transitions. 
Considering Table \ref{tab:datasets}, several datasets were used; however, most appeared in only one paper. {In addition, some papers that shared datasets in common frequently shared authors (or co-authors) or the task, \eg, short-text classification and topic modeling.} All the links in the column \textit{Information / Access} were verified on {19th September 2024}. In addition, some datasets were flagged as \textit{obtained by the authors}. It means that the authors collected the datasets, either manually or through APIs, but the datasets are not publicly available for download. 

Regarding the datasets as depicted in Table \ref{tab:datasets}, some may share the same name, such as Twitter, and New York Times. However, it was impossible to assert that they are the same dataset. Thus, we added a new line in the table instead of aggregating data regarding a particular dataset. In addition, at least three mechanisms were referred to as API providers for data collection: Twitter\footnote{https://developer.twitter.com/en/docs/twitter-api}, The Guardian\footnote{https://open-platform.theguardian.com/} and The New York Times\footnote{https://developer.nytimes.com/apis}. Thus, since the queries can be performed ranging from different dates and keywords, the datasets of the same name may correspond to different datasets. 

\subsection{Datasets description}

Below we provide short descriptions of each dataset listed in Table \ref{tab:datasets}. We highlight that some datasets included raw texts, while a few contain the bag-of-words representation of texts, \ie, preprocessed texts.

\subsubsection{20NewsGroup} This dataset contains approximately 20,000 news across 20 groups. In the link provided in this paper, there are three versions of this dataset, with slight variations. 

\subsubsection{Arxiv} According to \cite{p31}, this dataset contains approximately 2 million abstracts of papers published comprising the years between 2007 and 2021.

\subsubsection{{CLINC150 and CLINC150-SUR}}
{In the context of task-oriented dialog systems, CLINC150 \cite{larson2019evaluation} is a crowdsourced dataset containing 22,500 in-scope queries regarding 150 intents from 10 general domains and 1,200 out-of-scope queries. CLINC150-SUR \cite{p47} is an extension of the CLINC150 dataset, in which \citet{p47} generated more instances, added rephrased instances (generated with Parrot \cite{prithivida2021parrot}, and upsampled with LAMBADA \cite{anaby2020not}, reaching 600,000 instances.}

\subsubsection{CrisisLexT26}
\citet{p2} cited that CrisisLexT26 \citep{olteanu2015expect} is a collection of datasets related to several crises worldwide. However, \citet{p2} used only the datasets related to the Colorado Floods, containing 751 relevant and 224 irrelevant tweets, and Australian Bushfires, containing 645 relevant and 408 irrelevant tweets.

%\subsubsection{EmailData}
\subsubsection{EmailingList}
This dataset contains 1500 samples with 913 dimensions, \ie, boolean bag-of-words, corresponding to email messages, to be classified as junk or interesting. According to \citet{katakis2010tracking}, these samples were collected from Usenet posts existing inside the 20Newsgroup dataset.

\subsubsection{EveTAR}
EveTAR is an Arabic dataset that contains 1392 tweets on three terrorist events: (i) a suicide bombing in Ab, Yemen; (ii) Air strikes in Pakistan; and (iii) the Charlie Hebdo attack. \citet{p10} used this dataset to evaluate the ability of AIS-Clus to receive texts and detect events in languages other than English.

%%%%%%%%%%%%% TABELA

\begin{ThreePartTable}
\
\begin{landscape}
% \resizebox{.86\linewidth}{!}{
\begin{footnotesize}
\begin{longtable}{p{2.2cm}p{1cm}p{8.3cm}p{4cm}}
%\rowcolors{2}{gray!25}{white}
%\caption{List of datasets used in the papers and their respective resources, when available.}
\multicolumn{4}{c}{\textbf{Table \thetable} List of datasets used in the papers and their respective resources, when available. }
% \resizebox{.86\linewidth}{!}{
% \begin{tabular}{llll}
%\hline
\\
\toprule
\textbf{Dataset}  & \textbf{Papers}  & \textbf{Information / Access}  & \textbf{Stream Mining Tasks}  \\ 
\toprule
\endfirsthead

\multicolumn{4}{c}{\textbf{Table \thetable} List of datasets used in the papers and their respective resources, when available. 
\textit{(continued)}} \\ 
\toprule
\textbf{Dataset}  & \textbf{Papers}  & \textbf{Information / Access}  & \textbf{Stream Mining Tasks} *  \\\hline % \midrule
\endhead

%\hline
\rowcolor{gray!25}20NewsGroup                             & \citep{p29}  & \url{http://qwone.com/$\sim$jason/20Newsgroups/}  & Short-text clustering, Classification  \\
\rowcolor{gray!25}                                        & \citep{p30}  & &  \\
\rowcolor{gray!25}                                        & {\citep{p39}}  & &  \\
\rowcolor{gray!25}                                        & {\citep{p45}}  & &  {Multilabel classification}\\
\rowcolor{gray!25}                                        & {\citep{p48}}  & {Versions with gradual and abrupt drifts generated by the authors} & {Classification} \\
Arxiv                                   & \citep{p31}  & \url{https://www.kaggle.com/datasets/Cornell-University/arxiv}  & Classification  \\
\rowcolor{gray!25}{CLINC150}                            & {\citep{p47}}  & {\url{https://github.com/clinc/oos-eval} \cite{larson2019evaluation}}  & {Short-text classification}  \\
{CLINC150-SUR}                            & {\citep{p47}}  & {Based on the original CLINC150 dataset, with simulated user requests}  & {Short-text classification}  \\
& & {\url{https://huggingface.co/datasets/ibm/clinic150-sur}}  &  \\
\rowcolor{gray!25}CrisisLexT26                            & \citep{p2}  & obtained from \url{https://archive.org/details/twitterstream}  & Crisis management  \\
\rowcolor{gray!25}                                        &   & \citep{olteanu2015expect} &   \\
EmailingList                            & \citep{p5}  & \url{http://mlkd.csd.auth.gr/datasets.html}  & Classification  \\
                                        & \citep{p19}  &  &   \\
\rowcolor{gray!25}EveTAR                                  & \citep{p10}  & \url{http://qufaculty.qu.edu.qa/telsayed/evetar}  & Event detection  \\
Guardian, The                           & \citep{p15}  & obtained by the authors  & Classification  \\
%HCR                                     & \citep{p10}  & https://bitbucket.org/speriosu/updown/downloads  & Novelty detection  \\
\rowcolor{gray!25}Irish Times, The                        & \citep{p26}  & \url{https://www.kaggle.com/datasets/therohk/ireland-historical-news}  & Topic modeling  \\
\rowcolor{gray!25}                                        & \citep{p27}  &   & \\

{NELA-GT-2018}                        & \citep{p37}  & \url{https://doi.org/10.7910/DVN/ULHLCB} \citep{norregaard2019nela} & Classification  \\
\rowcolor{gray!25}{NELA-GT-2019}                        & \citep{p37}  & \url{https://doi.org/10.7910/DVN/O7FWPO}  \citep{gruppi2020nela} & Classification  \\
{NELA-GT-2020}                        & \citep{p37}  & \url{https://doi.org/10.7910/DVN/CHMUYZ} \citep{gruppi2021nela}   & Classification  \\
\rowcolor{gray!25}New York Times, The                     & \citep{p15}  & obtained by the authors  & Classification  \\  
\rowcolor{gray!25}                                        & \citep{p9}  & \url{https://ir-datasets.com/nyt.html}  & Classification  \\  
\rowcolor{gray!25}                                        & \citep{p26}  & \url{http://archive.ics.uci.edu/ml/datasets/Bag+of+Words}  & Topic modeling  \\  
\rowcolor{gray!25}                                        & \citep{p31}  & \url{https://www.dropbox.com/s/nifi5nj1oj0fu2i/data.zip?dl=0}  & Classification  \\
NOAA                                    & \citep{p11}  & not provided but probably from \url{https://data.noaa.gov/dataset/}  & Event detection  \\
                                        & \citep{p12}  &   &   \\
                                        & \citep{p14}  &   &   \\
\rowcolor{gray!25}NSDQ                                    & \citep{p18}  & \url{https://github.com/ChristophRaab/NASDAQ-Dataset}  & Classification  \\
\rowcolor{gray!25}                                        & \citep{p28}  &   &   \\
OffensEval                              & \citep{p22}  & \url{https://competitions.codalab.org/competitions/20011}  & Classification  \\
\rowcolor{gray!25}RCV1                                    & \citep{p9}  & Available via Scikit-learn library\footnote{\url{https://scikit-learn.org/stable/modules/generated/sklearn.datasets.fetch\_rcv1.html}}.  & Classification  \\
Reuters-21578                           & \citep{p1}  & \url{https://archive.ics.uci.edu/ml/machine-learning-databases/reuters21578-mld/}  & Topic modeling  \\
                          & {\citep{p45}}  &   & {Clustering, classification}  \\
\rowcolor{gray!25}SemEval2020 - Subtask 2 (CCOHA)         & \citep{p35}  & \url{https://www.english-corpora.org/coha/} & Semantic Shift Detection  \\
\rowcolor{gray!25}                                        &  & \citep{alatrash2020ccoha,schlechtweg2020semeval}  &   \\
SemEval2020 - Subtask 2 (LatinISE)      & \citep{p35}   & \url{https://lindat.mff.cuni.cz/repository/xmlui/handle/11234/1-2506} & Semantic Shift Detection  \\
                                        &               & \citep{mcgillivray2013tools,schlechtweg2020semeval}  &   \\
\rowcolor{gray!25}SO-T  & \citep{p23}  & obtained by the authors  & Short-text clustering  \\
SpamAssassin  & \citep{p19}  & \url{http://mlkd.csd.auth.gr/datasets.html}  & Classification  \\
\rowcolor{gray!25}SpamData                                & \citep{p5}  & \url{http://mlkd.csd.auth.gr/datasets.html}  & Classification  \\
\rowcolor{gray!25}                                        & \citep{p13}  &   &   \\
\rowcolor{gray!25}                                        & \citep{p19}  &   &   \\
Ts-T, Tw, Tw-T, Tweets, Tweets-T        & \citep{p6}  & \url{https://trec.nist.gov/data/microblog.html}  & Short-text clustering  \\
                                        & \citep{p23}  &   &   \\
                                        & \citep{p30}  &   &   \\
                                        & {\citep{p40}}  &   &   \\
\rowcolor{gray!25}Tweets, TweetSet, Twitter               & \citep{p4}            & obtained by the authors  & Short-text classification, Event detection \\
\rowcolor{gray!25}                                        & \citep{p21}           & obtained by the authors  & Short-text classification, Event detection \\
\rowcolor{gray!25}                                        & \citep{p34}           & obtained by the authors  & Short-text classification, Event detection \\
\rowcolor{gray!25}                                        & \citep{p16}           & obtained by the authors  & Stance detection  \\
\rowcolor{gray!25}                                        & \citep{p20}           & obtained by the authors  & Stance detection  \\
\rowcolor{gray!25}                                        & \citep{p33}           & obtained by the authors  & Stance detection  \\
\rowcolor{gray!25}                                        & \citep{p22}           & \url{https://archive.org/details/twitterstream}  & Classification  \\  
\rowcolor{gray!25}                                        & \citep{p24}           & obtained by the authors  & Short-text clustering  \\
\rowcolor{gray!25}                                        & \citep{p-25}          & obtained by the authors  & Concept drift detection \\
\rowcolor{gray!25}                                        & {\citep{p36}}          & \citep{wang2014summarization}  & Short-text classification \\
\rowcolor{gray!25}                                        & {\citep{p40}}          & obtained by the authors  & Short-text clustering \\
\rowcolor{gray!25}                                        & {\citep{p38}}          & obtained by the authors  & Sentiment drift detection \\
\rowcolor{gray!25}                                        & {\citep{p45}}          & \url{https://github.com/jackyin12/GSDMM/}  & {Clustering, classification} \\
\rowcolor{gray!25}                                        & {\citep{p41}}          & \url{https://github.com/cristianomg10/sentiment-drift-analysis-text-stream-football}  & {Short-text classification, Sentiment drift detection} \\
\rowcolor{gray!25}                                        & {\citep{p42}}          & \url{https://github.com/cristianomg10/temporal-analysis-of-drifting-hashtags-in-textual-data-streams-a-graph-based-application}  & {Topic modeling, Clustering} \\
\rowcolor{gray!25}                                        & {\citep{p46}}          & obtained by the authors & {Classification} \\
TwitterSentiment                        & \citep{p5}  & \url{https://bit.ly/twitter-sentiment-link}  & Classification  \\
\rowcolor{gray!25}UCI News                                & \citep{p27}  & \url{https://www.kaggle.com/datasets/uciml/news-aggregator-dataset}  & Topic modeling  \\
Usenet1                                 & \citep{p19}  & \url{http://mlkd.csd.auth.gr/datasets.html}  & Classification  \\
\rowcolor{gray!25}Usenet2                                 & \citep{p19}  & \url{http://mlkd.csd.auth.gr/datasets.html}  & Classification  \\
%Usenet3  & \citep{p19}  & http://mlkd.csd.auth.gr/datasets.html  & Classification  \\
USGS                                    & \citep{p11}  & not provided but probably from \url{https://www.usgs.gov/products/data}  & Event detection  \\
                                        & \citep{p12}  &   &   \\
                                        & \citep{p14}  &   &   \\
\rowcolor{gray!25}vg.no                                   & \citep{p8}  & obtained by the authors  & Classification  \\
Yelp datasets                           & \citep{p17}  & \url{https://www.yelp.com/dataset}  & Fake reviews detection \\
\rowcolor{gray!25}{Y-Art, Y-bus, Y-com, Y-Edu, Y-Ent, Y-Soc}  & {\citep{p43}}  & {\url{https://www.uco.es/kdis/mllresources/}} {\citep{read2012scalable,osojnik2017multi,nanculef2014efficient,nguyen2019multi,liu2018online}}  & Multilabel classification 
\\\hline
% \end{tabular}
% }
\label{tab:datasets}
\end{longtable}
\end{footnotesize}
% }
\end{landscape}
\end{ThreePartTable}

%%%%%%%%%%%%%%%%%%%%%%%%%%

\subsubsection{Guardian, The}
\citet{p15} collected a news stream from The Guardian using the API. The dataset contains 10 categories and 40,000 samples, represented using Word2Vec with 300 dimensions.

% \subsubsection{HCR}
% The Health Care Reform (HCR) dataset contains approximately 2,500 tweets distributed across eight topics regarding politics.

\subsubsection{Irish Times, The}
The Irish Times dataset corresponds to a set of 1.6 million news headlines published by the Irish Times, distributed in six classes. It comprises 25 years of publications.

{\subsubsection{NELA-GT}
NELA-GT \citep{norregaard2019nela,gruppi2020nela,gruppi2021nela} corresponds to a series of datasets regarding news and media outlets. In addition, conspiracy sources are included in this dataset. The authors incorporated ground-truth ratings of aspects such as reliability, transparency, and bias. NELA-GT-2018 \citep{norregaard2019nela} contains 713 thousand items from 194 media outlets and conspiracy sites; NELA-GT-2019 \citep{gruppi2020nela} contains 1.12 million media articles from 260 mainstream and alternative sources collected in 2019; NELA-GT-2020 \citep{gruppi2021nela} contains almost 1.8 million news stories from 519 sources. \citet{p37} used these datasets in the fake news detection, using the instances labeled as reliable and unreliable. The datasets were merged, but the temporal order was respected.}

\subsubsection{New York Times, The}
\citet{p15} used The New York Times' public API to collect news articles between January 2006 and January 2018. These news articles were encoded using Word2Vec, with 300 dimensions.
\citet{p9} used a dataset collected from The New York Times, containing news articles from 1987 and 2007, distributed in 26 categories \citep{sandhaus2008new}. 
\citet{p26} used only the title of news articles from the New York Times. The authors mentioned that the dataset contained 1,764,127 titles, with an average of five words per title.
\citet{p31} utilized a dataset collected from the News York Times containing 99,872 articles dating from 1990 to 2016.

\subsubsection{NOAA}
The National Oceanic and Atmospheric Administration (NOAA) is an agency in the United States government. It does not correspond directly to a dataset; however, \citet{p11}\citep{p14} and \citet{p12} used NOAA reports as ground truth for the automatic classification of tweets. No details were offered about the reports' processing or collection.

\subsubsection{NSDQ}
The NSDQ dataset (named after NASDAQ) corresponds to tweets regarding 15 companies listed in NASDAQ. NSDQ was compiled by the authors in \citet{p18} and \citet{p28} and comprised the months of February to December 2019. This dataset contains 30,278 tweets.

\subsubsection{OffensEval}
\citet{p22} used the OffensEval 2019 dataset \citep{zampieri2019semeval}. The dataset contains 14,000 tweets posted in 2019, categorized into offensive and inoffensive.  

\subsubsection{RCV1}
RCV1 \citep{lewis2004rcv1} is a dataset that contains 403,143 news from Reuters News between 1996 and 1997. The news articles are divided into three classes: industries, topics, and regions. This dataset is organized hierarchically. From this dataset, \citet{p9} obtained a corpus with 12 subtrees (labels).

%\subsubsection{Reuters}

\subsubsection{Reuters-21578}
\citet{p1} used this dataset, which contains articles with their respective categories temporally ordered. According to \citet{p1}, it contains 12,902 news, each classified into several categories, totaling 90 categories.

\subsubsection{SemEval2020 - Subtask 2}
\citet{p35} used the datasets corresponding to Task 2 of SemEval2020, regarding the semantic shift detection task. The datasets used were CCOHA \citep{alatrash2020ccoha} and LatinISE \citep{mcgillivray2013tools}. CCOHA contains texts in English that range from approximately 1810 to 2000, while LatinISE has Latin texts that range from the 2nd century BC to the 21st century AD. Both have target words, which are words that can be monitored to detect the semantic shift. These datasets were discovered in the selected papers that span the longest.

\subsubsection{SO-T}
\citet{p23} collected duplicated question titles regarding Python, Java, jQuery, R, and other programming languages/tools. In the paper, the authors carefully described the process of obtaining this dataset. In the end, this dataset contained 400,000 randomly selected pairs of question titles.

\subsubsection{SpamAssassin and SpamData}
These datasets correspond to emails collected from the \textit{Spam Assassin} collection. They are represented as bag-of-words, distributed across two classes, ham and spam, in imbalanced proportions (80\% and 20\%, respectively). Both contain 9,324 instances; however, SpamAssassin \cite{katakis2008ensemble} has 40,000 features, while SpamData has 499 \citep{katakis2009adaptive}. It is noted that these datasets contain gradual drifts \citep{p19}.

\subsubsection{Ts-T, Tw, Tw-T, Tweets, Tweets-T}
\citet{p6} used this dataset named \textit{Tweets}, containing 30,332 tweets distributed into 269 groups, with 7.97 words per tweet on average. The authors also generated a variant dataset from \textit{Tweets}, called \textit{Tweets-T}, where the dataset is sorted by topic. \citet{p23} used the same dataset \textit{Tweets}, called \textit{Ts-T}. \citet{p30} named the same datasets presented in \cite{p6} as \textit{Tw} and \textit{Tw-T}, respectively.

\subsubsection{Tweets, TweetSet, Twitter}
\citet{p4} used a Tweets dataset containing approximately 400,000 tweets. They stated that the data acquisition comprises November and December 2012, using the Twitter API. \citet{p21} also obtained a dataset through Twitter API and consists of 803,613 short texts distributed in four categories. 
\citet{p34} described the dataset used in their work as ``Twitter sentiment analysis training corpus'', from which they filtered 10\% of the data, totaling 104,857 tweets. \citet{p16} collected tweets by using a Java library named GetOldTweets\footnote{https://github.com/Jefferson-Henrique/GetOldTweets-java/}. They collected 112,397 tweets posted between September 2016 and January 2017, using vaccine-related keywords. \citet{p20} extended the dataset obtained in \cite{p16} until September 2019, corresponding to 806,672 tweets. \citet{p33} collected 486,688 tweets from July 2021 to December 2021 regarding the Green Pass, as the European Union COVID-19 Digital Certificate is known in Italy. \citet{p22} used tweets to perform country hashtag prediction in two different years: 2014 and 2017, consisting of 472,000 and 407,000 tweets, respectively. The tweets were obtained from the Internet Archive\footnote{https://archive.org/details/twitterstream}.
\citet{p24} experimented with their approach using a Twitter dataset, namely \textit{TweetSet}, containing about 144,000 tweets posted in June 2019, distributed into 16 categories. 
\citet{p-25} collected tweets by monitoring a set of users and hashtags, \ie, words with a \# at the beginning that simulate a tag for the tweet. The authors monitored, for instance, @dilmabr (former Brazilian president) and \#dolar (Portuguese for dollar). The dataset size was not mentioned. {\citet{p41} collected tweets regarding a specific soccer match between two South American clubs in an international cup. This dataset contains 37,126 tweets, and this is one of the very rare datasets with labeled drifts. \citet{p42} collected tweets comprising 2018 to 2022, totaling 255,131 tweets. These tweets were related to the hashtag \#mybodymychoice, and, in this specific study, the authors performed a community detection algorithm over an incremental graph method to detect hashtag drifts.}

{Although Twitter-based datasets were very frequent in the studied papers, as of February 2023, Twitter's API policies have changed\footnote{Available at: https://www.forbes.com/sites/jenaebarnes/2023/02/03/twitter-ends-its-free-api-heres-who-will-be-affected/?sh=36ad308a6266. Accessed on September 17th, 2023.}, and it became a paid service.}

\subsubsection{TwitterSentiment}
TwitterSentiment (or TSentiment, as in \cite{p5}) is a balanced dataset that contains 1.6 million tweets collected between April and June 2009. These tweets are labeled as positive or negative using distant supervision. In this case, emoticons were used for labeling.

\subsubsection{UCINews}
The UCINews dataset contains 422,937 news collected between March and August 2014. Each news item can be categorized as business, science and technology, entertainment, or health. This data collection also includes each news id, title, URL, publisher, story id, hostname, and timestamp information.

\subsubsection{Usenet1 and Usenet2}
Similarly to EmailingList, both Usenet1, and Usenet2 simulate a sequence of 1500 emails from the 20NewsGroup dataset to a particular user to be classified as junk or interesting \citep{katakis2008ensemble}. According to \citet{p19}, both datasets have 100 features corresponding to words.

%\subsubsection{Usenet3}
%Usenet3 is similar to Usenet1 and Usenet2. However, it contains many more features (\ie 27893) and 5997 instances. It is mentioned this dataset contains sudden/abrupt drifts \citep{p19}.

\subsubsection{USGS}
United States Geological Survey (USGS) is a scientific agency from the United States. Similarly to NOAA, USGS reports do not correspond to datasets and are also used as ground truth to classify tweets automatically by \citet{p11}\citep{p14} and \citet{p12}. 

\subsubsection{vg.no}
Vg.no is a Norwegian news website. 
\citet{p8} obtained news from four topics: European Union, economy, sports, and entertainment. However, the authors did not mention the size of the collected dataset.

\subsubsection{Yelp datasets} 
\citet{p17} used four real-world datasets based on the datasets provided by Yelp, namely Yelp CHI, Yelp NYC, Yelp ZIP, and Yelp Consumer Electronics. The authors used Yelp CHI (Chicago) \citep{mukherjee2013fake}, containing more than 67,000 reviews of restaurants and hotels distributed between 2004 and 2012. Yelp NYC \citep{rayana2015collective} contains approximately 322,000 reviews of restaurants in New York City. It comprises the years between 2004 and 2015. Yelp ZIP \citep{rayana2015collective} contains 608,598 reviews from New Jersey, Vermont, Connecticut, and Pennsylvania. Yelp Consumer Electronics \citep{barbado2019framework} contains almost 19,000 records evenly distributed between genuine and fake. These datasets include other data, such as user information, product information, rating, timestamp, and review, and were scraped/downloaded from Yelp.com. 

\subsubsection{{Y-Art, Y-bus, Y-com, Y-Edu, Y-Ent, and Y-Soc}} {\citet{p43} leveraged the datasets Y-Art, Y-bus, Y-com, Y-Edu, Y-Ent, and Y-Soc. These datasets are based on Yahoo, and each class has second-level categories. All datasets can be found in the link provided in Table \ref{tab:datasets}, together with several multilabel datasets.} 

Although SpamAssassin and EmailingList have known concept drifts (gradual and abrupt)\footnote{According to http://mlkd.csd.auth.gr/concept\_drift.html}, an interesting aspect is that {only two} datasets across the papers analyzed, {\ie, \citet{p41} and \citet{p48}}, have labeled concept drifts, due to the difficulty of defining the specific points of drift, which requires a deep study on a particular dataset.
%\JPBCOMMENT{frase estranha: }\CGCOMMENT{Ajustado.}
Thus, some works attempted to force concept drifts by: (i) placing data partitions temporally disordered in a stream, \ie, data from 2011 and 2015 before 2012 \citep{p17}; or (ii) rearranging the data, sorting by classes or topics \citep{p4}. This aspect is extended in Section \ref{sec:on-drift}.

Therefore, since we could not locate repeating datasets in more than three papers, we can conclude that the research area of concept drift detection in textual streams lacks benchmark datasets. Furthermore, all the datasets used for classification are instance-level labeled, \ie, sentences/tweets labeled. In addition, the resource of one of the most recurrent datasets, \ie, \textit{TagMyNews} {and \textit{Snippets} \citep{phan2010hidden}}, could not be encountered across the papers. Also, it is closely related to short-text applications, which constitutes an entirely new research area.

%\subsection{Concept Drift Visualization and Simulation}
%\label{subsec:on-drift-visualization-and-protocolos}

\section{Concept Drift Visualization and Simulation}
\label{sec:on-drift}
%\akst{Regarding texts,} 
It is challenging to clearly express or prove the existence of concept drifts in a particular textual dataset. However, a few works attempt to justify the existence of drifts by resorting to plots. {In this section, we only provide references for the figures due to copyright restrictions.} For example, \citet{p4} used normalized stacked bar plots to demonstrate the topic distribution over several batches (Figure 4 in \citep{p4}). %Fig. \ref{fig:p4-topic-distribution} depicts the plot used in \cite{p4}, with the y-axis representing the topic distribution and the x-axis, the index of the data chunk.
%Fig. \ref{fig:p4-topic-distribution} depicts the plot used in \cite{p4}, with the y-axis representing the topic distribution and the x-axis, the index of the data chunk.

% \begin{figure}[!htp]
% \includegraphics[width=0.55\textwidth]{fig11-p4-topic-distribution.png}
% \centering

% \caption{Normalized stacked bar plot depicting the topic distribution across batches \citep{p4}.}
% \label{fig:p4-topic-distribution}
% \end{figure}

%\akst{The authors in} 
\citet{p33} plotted the distribution of the stances across the analyzed period using a normalized stacked area plot, similar to the stacked bar plot, to show the topic distribution over time. %, as depicted in Fig. \ref{fig:p33-drift}. 
The background color regards the stance of tweets about the Green Pass, distributed in positive (in blue), neutral (in white), and negative (in red). Considering the color code aforementioned, the thicker line corresponds to the average stance at each moment in the timeline. This description relates to Figure 3 in \cite{p33}. {Similarly, \citet{p41} depicted the sentiment distribution regarding a soccer match over time (Figure 4 in their paper) by using a stacked area plot. In addition, the authors visualized the sentiment drift splitting the match into quarters, \ie, Figure 3 in \cite{p41}.}

% \begin{figure}[!htp]
% \includegraphics[width=1.0\textwidth]{fig12-p33-drift.png}
% \centering

% \caption{Normalized stacked area plot illustrating the sentiment distribution across the analyzed period \citep{p33}.}
% \label{fig:p33-drift}
% \end{figure}

However, \citet{p11} and \citet{p18}\citep{p28} used dimensionality reduction methods, \ie, either t-SNE or PCA, to reduce high-dimensional representations to two dimensions, which can easily be plotted. Thus, \citet{p14} and \citet{p18}\citep{p28} used t-SNE to confirm that there are drifts between texts of specific hashtags. Figure 4 in %\ref{fig:p18-drift} 
\cite{p18} depicts the visual representation of concept drift. The data points of different colors in different positions indicate that texts regarding particular stock tickers have different patterns. However, it does not highlight temporal changes. 

% \begin{figure}[!htp]
% \includegraphics[width=0.7\textwidth]{fig13-p18-drift.png}
% \centering

% \caption{Representation of concept drift in Twitter posts, considering 15 stock tickers \citep{p18}. Dimensionality reduction was performed using t-SNE.}
% \label{fig:p18-drift}
% \end{figure}

%\akst{The authors in} 
%The authors in 
\citet{p11} used PCA for dimensionality reduction for plotting and suggesting a direction of drift based on data from 2014 and from four months in 2018. It is not possible to categorize the drifts shown by the images considering the literature presented in Section \ref{sec:dsm}.
%and \ref{sec:semantic-shift}. 
Figure 10 in \cite{p11} %\ref{fig:p11-drift} 
is a plot of text representations reduced to bi-dimensional vectors using PCA. The authors colored the data points according to the month or year of the posts' timestamps. Posts from 2014 occupy the center left of the image, while the representations of the other posts published in 2018, identified as July, August, September, and October, occupy the center and bottom of the image. In addition, the authors drew an arrow to show the direction of the concept drift.

% \begin{figure}[!htp]
% \includegraphics[width=0.65\textwidth]{fig14-p11-drift.png}
% \centering

% \caption{Representation of concept drift in social posts, considering data from 2014 and four months in 2018, with dimensionality reduction performed using PCA \citep{p11}.}
% \label{fig:p11-drift}
% \end{figure}

{In an ad-hoc manner, we mention some interesting works that approach concept drift / semantic shift visualization. \citet{kazi2022visualization} proposed three visualization methods that emphasize the changes over time, starting with a reference word. The first proposed method is the radial bar chart, which can show top similar words, word re-occurrence, and degree of similarity. For example, considering Figure 1 in their work, the word \textit{cigarette}, in the 1980s, was related to \textit{tobacco}, while in the 2020s, it was related to \textit{vape} and \textit{ecigarettes}. Figure 2 in their work corresponds to a second proposal regarding the spiral line chart. This chart enables visualization of similar words, word re-occurrence, and continuity. Therefore, it eases understanding the appearance and fade of words related to a reference word. More specifically, Figure 2 in their paper considers the word \textit{anxiety}\footnote{Available at: \url{https://public.tableau.com/app/profile/raef6267/viz/SpiralLineChartConceptDrift/SpiralLineChart}. Accessed on September 23rd, 2024.}. To enhance the visualization of geographical information, the authors proposed a word cloud using maps of countries as silhouettes. Figure 3 in their paper shows this method. The authors analyzed the word \textit{divorce}, hypothesizing that the use of this word in the 1970s/1980s regarded the divorce itself, while in recent years, \ie, 2010s/2020s, it regarded the consequences of divorce, such as violence and self-harm. \citet{periti2023studying} also provided interesting highlights by using visualization. Although this work did not appear among the selected works, it uses WIDID \cite{p35}, which was among the selected papers. \citet{periti2023studying} studied the semantic shifts of the Italian parliamentary speeches over time. The authors exemplified using the word \textit{clean}. One visualization represents polysemy and the semantic shift of a word itself over time, \eg, Figure 4a in their paper. On the other hand, Figure 4b in their paper emphasizes the prominence and sense shift of the sense nodules of a given word over time. Although unrelated to streams and drifts, \citet{huang2023va}\footnote{Available at: \url{https://va-embeddings-browser.ivis.itn.liu.se/}. Accessed on September 23rd, 2024.} provided an interesting visual survey for embedding visualization. We included this work in this discussion since a considerable number of selected papers leveraged embeddings as text representation methods and, therefore, \citet{huang2023va} may inspire the development of new visualization methods towards drift visualization.}

As aforementioned, concept drift in texts is common and can occur over time. However, depending on the characteristics of the approach and datasets, it may be challenging to execute the experiments due to the lack of certainty of the existence of drift, their potential positions, and their behavior over time. Therefore, some papers simulate drifts. 
For example, \citet{p1,p26,p36} and \citet{p39} rearranged the topics sequentially in the stream. Thus, when a new topic emerges from the stream, it is considered a drift. \citet{p17} simulated drift by dividing the datasets into partitions and rearranging them in different orders. For example, one of the datasets is initially ordered temporally and divided into five partitions, \ie, $D1$, $D2$, $...$, $D5$. Thus, in a specific scenario, the authors merged $D1-D3$ for training and used the other partitions, \ie, $D2$, $D4$, and $D5$, for testing sequentially. Although it created a scenario of concept drift and worked for the experiment in the aforementioned papers, both scenarios are unrealistic, especially considering the temporal aspects of the partitions in the latter example. 

{Across the analyzed papers, a number of authors mentioned the difficulty of finding datasets with labeled drifts (\cite{p47} and \cite{p48}, to mention a few). To prove this aspect, considering the 48 papers analyzed, only two papers mentioned the existence of labeled drifts in their datasets: \cite{p41} (tweets regarding a soccer match) and \cite{p48} (for AGNews and 20NewsGroup), although those presented in \cite{p48} had their drifts (gradual and abrupt) artificially generated. This leads to the development of text drift generation methods to allow testing text classification methods and text drift detectors. \citet{p48} introduced drifts based on a procedure initially developed by \citet{katakis2008ensemble}. \citet{garcia2024methods} presented four text drift generation methods, \ie, class swap, class shift, time slice removal, and adjective swap, based on \citet{p32}, in which the former three involve manipulating classes, while the latter manipulates the sentence meaning by swapping adjectives with their antonyms.}

Ultimately, depending on the sort of text drift, it can be challenging to visualize due to several factors, such as the inherent high dimensionality of the most frequent text representations. In addition, visually representing changes in text behavior over time can be challenging. Furthermore, developing scenarios to force concept drift in text streams can be complex, depending on the type of text drift. Generally, the datasets are described in the papers; however, sometimes, they lack evidence for the existence of text drift. Thus, it is necessary to resort to data rearrangement to simulate drifts and data visualization to search for changes in temporal patterns. However, to maintain consistency, it may be essential to consider the temporal order, especially concerning streaming scenarios. 

\section{Conclusion\protect{, Open Challenges,} and Future Directions}
\label{sec:conclusion}

In this study, we performed a systematic literature review on concept drift adaptation, specifically in text streams {scenarios}. A text stream is a specialization of data streams in which several texts arrive sequentially at high speeds. Sequentially handling texts is challenging due to the constraints of data stream settings, \ie, processing time and memory consumption. In addition, we can mention characteristics of text-related settings, such as vocabulary maintenance, NLP, and text representation maintenance{; ideally, these tasks should be performed on the fly}.

We selected {48} papers and extracted information %\akst{from them} 
according to the defined criteria. We evaluated {and categorized} the papers regarding categories of drift, types of drift detection, the ML model update scheme, the stream mining tasks applied, the text representation method utilized, and the update scheme of the text representation methods. {In this study, }we also {provided} the metrics used in each stream mining task.

{Text drift may happen due to several reasons. The natural evolution of writing can lead to drift, such as the emergence or disappearance of new words. In addition, texts generally reflect changes in the real world. \citet{p41} mentioned that the drifts were generated by the goals scored by a team, leading to a positive sentiment. In \citet{p11,p12,p14}, the change in the volume of tweets regarding landslides could indicate the occurrence of the actual event. \citet{p4} used cosine distance between clusters generated from different chunks to indicate the existence of topic drifts. A topic drift may happen due to changes in user interests over time. \citet{p18} mentioned the existence of drift in the dataset comprising tweets regarding different stocks from NASDAQ. These changes may happen due to the increase of posts because of actual news posts, any positive event such as an increase in the profit, announcement of dividends, or even negative events, such as scandals and corruption. \citet{p17} worked on an adversarial problem, \ie, fake reviews detection, in which a classifier model needs to be updated frequently to overcome new writing patterns from unlawful reviewers. Therefore, drifts, in this case, corresponded to those changes in writing patterns to bypass the classifier. In \citet{p16,p20}, the drifts were the changes in stance distribution regarding specific topics, such as vaccination. \citet{p-25} considered drift the changes in the volume of tweets regarding news on Brazilian politics. In \citet{p2}, the drifts corresponded to changes in writing patterns to define whether the post was relevant to crisis management. In \citet{p42}, the drifts regarded the hashtag \textit{\#mybodymychoice} in different uses other than its original context. \citet{p47} mentioned that drifts in their scenario regarded the failure of a newly deployed feature in systems. To summarize, many different reasons can cause text drifts, generally reflected by actual changes in the real world. Although it is a frequent phenomenon, text drifts are rarely labeled. It is a clear outcome of the difficulty of finding the exact point of many of those scenarios mentioned above. To confirm this statement, only two papers provided datasets with labeled drifts \cite{p41,p47}. However, in \citet{p47}, the drifts were introduced in the datasets to evaluate their method.}

{Regarding categories of drift, we differentiated the types into \textit{real}, \textit{virtual}, \textit{feature drift}, and \textit{semantic shift}. Most works ({44}) approached the real drift problem, corresponding to the mapping changes between $X$ and $y$ over time. Only {four} works considered the virtual drift, and another three tackled the semantic shift problem. Please note that a work can approach more than one drift category simultaneously. Considering the drift detection method, we investigated the papers and observed that it is possible to categorize them into \textit{adaptive}, where the method adapts to the concept drift without detecting it, and \textit{explicit}, where there is an explicit concept drift detection that can trigger the ML model update.

Furthermore, we investigated the strategies employed by the methods and systems to update ML models as needed. We categorized the studied papers considering the ML update scheme into four groups: (i) ensemble update, (ii) incremental, (iii) keep-compare-evolve, and (iv) retraining. {In addition}, we analyzed the applications approached in the papers according to a stream mining task categorization. The stream mining tasks found in the studies were categorized into classification, clustering, general detection, and topic modeling. Several applications were found, such as fake review detection, sentiment analysis, and novelty detection.

In addition, we organized and presented the text representation methods since they are crucial for text streams subject to concept drift. {Sixteen} text representation methods were identified, where Bag-of-words and Word2Vec were the most frequent methods (each appeared in 11 studies). Moreover, when available, the update mechanisms of the text representations were also listed. Only two methods are fully incremental, while most studies used static text representation methods/language models. Therefore, it constitutes an open challenge.
}

Additionally, we listed the real-world datasets with their links when available and discussed concept drifts visualization and {drifts simulation}. Some papers argued that the datasets in use have drift, although %\akst{they} \ak{such drifts}
such drifts are unlabeled or uncategorized. A few papers resorted to visualization techniques or data rearrangement to simulate drift to justify the existence of drifts. Concept drifts in text streams can manifest in various ways, including feature drift, semantic shift, real and virtual drifts, and topic drift. Thus, different approaches are required to manage these types of drifts. 

{It is worth mentioning the extraordinary advances that have been made regarding LLMs recently. There is some discussion about the requirements for a language model to be considered large. For example, BERT is considered an LLM \cite{kurtic2022optimal}, although a pre-trained BERT large uncased has 340 million parameters.
Considering BERT-like families, some papers addressed the temporal adaptation in these language models. For example, \citet{hu2023learn} developed a framework to address temporal shifts in news posts. \citet{su2022improving} also directed their efforts to address semantic changes using language models. \citet{agarwal2022temporal}, on the other hand, evaluated the temporal effects on pre-trained language models. \citet{p22} also addressed concept drift but with a focus on text stream scenarios.}

{More recently, other works mentioned that LLMs are generally constituted of billions of parameters capable of performing tasks based on prompts, sometimes in a zero-shot fashion \cite{raina2024llm,kojima2022large}. However, combining LLMs such as Llama and GPT-3 (or more recent versions) in text stream scenarios subject to concept drift is still open.}

\subsection{{Open Challenges and Future Directions}}

During this study, we discovered aspects that can be addressed in future research and remain as open challenges. 

\subsubsection{{\textbf{Text drift visualization}}}
The research area still requires visualization methods that highlight the existence of text drift. There is no standard for generating those visualizations, especially regarding changes over time. {Due to the variety of tasks and applications, the existence of different visualizations with no standard is understandable. However, developing visualization methods that are easy to interpret and generate may help justify the presence of drifts.}

\subsubsection{{\textbf{Benchmark for text streams datasets subject to concept drift}}} 
{As verified in this paper, there is no benchmark to compare the ability of learning methods in text stream scenarios subject to concept drift. In particular, only two papers provided datasets with labeled drifts.} Therefore, different approaches for text drift simulation have been used in the literature. Standardization in these processes may be an advantage, enabling faster development of the research area.
{In addition, as it could be seen early in this section, the source of text drifts can be domain-specific, demanding further analysis for a deep understanding of the phenomenon.}

The authors in the selected papers collected many datasets; however, the
most frequent datasets across the papers were related to short-text scenarios or topics. Thus, it is crucial to develop benchmark datasets for text drift detection focused on text stream scenarios in the future.

\subsubsection{{\textbf{Incremental methods for semantic shift detection}}}
%enabling the next researches ``to play in the same field''. 
Considering the semantic shift, it can be advanced in linguistics and be studied in depth. According to the information obtained from the papers that approach semantic shift detection studied in this work, 
\review{a challenging aspect is the need to monitor all the words in the vocabulary.}
%a challenging aspect is that the target words are known \textit{a priori}, {although some methods can track changes in an entire vocabulary}. 
Thus, it appears that methods that can indicate words that suffer semantic shift in text streams are desired {to reduce computational load}. {Besides, a reduced number of papers approached semantic shift \review{in text stream scenarios, \eg, \citep{p35}}. Given that text streams have their constraints and incremental approaches are more suitable to these scenarios, producing incremental methods for semantic shift detection may help develop the area.}
%Additionally, we realized that most text representation methods are not updated during the process, and, as aforementioned, it constitutes a problem over time. Therefore, developing new, easy-to-use representation methods that can be updated over time may benefit the research area.

%Finally, as discussed in Section \ref{sec:datasets}, there are no benchmark datasets for text drift detection in text stream scenarios. 

\subsubsection{{\textbf{Incremental text representation methods}}} {As verified in this study, a few text representation methods were able to embed updates over time. For example, frequency-based approaches, such as Bag-of-Words and TF-IDF, may suffer from the appearance and disappearance of words over time, considering the case of defining the reference tokens at the beginning of the text stream processing. }
{Being able to incorporate updates over time to representations provided by pre-trained language models or effectively modeling dense representations over time without creating a bottleneck in the process may be a future direction regarding this regard.}

\subsubsection{{\textbf{Text drift detection in LLM environments}}} {Although studying LLMs such as Llama and GPT family are outside the scope of this paper, text drift detection may be important in scenarios that leverage LLMs. For example, the tasks of preventing jailbreaks and prompt injection can be modeled as text stream scenarios, in which the input is the user interactions with the LLM, and the jailbreaks and prompt injection could be analyzed as drifts in the user input stream. Prompt injection is a type of attack that introduces instructions to manipulate the LLM to perform the attacker's intention \cite{liu2023prompt}. At the same time, jailbreak, in this sense, means the input of malicious instructions to provoke undesired LLM's behavior \cite{deng2023multilingual}. Although there are pre-trained models for this task, such as the Prompt-Guard-86M\footnote{\url{https://huggingface.co/meta-llama/Prompt-Guard-86M}}, a problem in this regard is the nature of the scenario, which is clearly adversarial, meaning that incremental learning/adaptation is frequently desired. }

In summary, this {systematic} review provides a detailed analysis and evaluation of concept drift adaptation methods in text stream scenarios, offering valuable insights that may help readers understand the strengths and weaknesses of the current techniques and open issues that need to be addressed.

\section*{\protect{Acknowledgements}}
{We are grateful to the anonymous reviewers for their significant comments and suggestions to improve the manuscript.}
% \begin{acks}
% \end{acks}

\bibliographystyle{ACM-Reference-Format}
\bibliography{sample-base}

\appendix
\section{List of Acronyms}
\label{appendix:acronyms}

\review{In order to ease the reader to locate acronyms' meanings, we developed the Table \ref{table:acronyms}. This list provides the acronyms alphabetically ordered. Please, note that we did not include acronyms without a clear meaning provided by the acronym's author(s).}

\rowcolors{1}{white}{gray!25} % Alterna entre cinza claro e branco
\begin{longtable}{ l l }
\caption{List of acronyms, alphabetically ordered.}\\
\label{table:acronyms}

\textbf{Acronym} & \textbf{Meaning} \\
\hline
\endfirsthead

\rowcolors{1}{white}{gray!25} % Reinicia o padrão de cor nas páginas subsequent
\textbf{Acronym} & \textbf{Meaning} \\
\hline
\endhead
AdaNEN & Adaptive Neural Ensemble Method \citep{p48} \\
ADWIN & Adaptive Windowing \citep{bifet2007learning}\\
AE & Autoencoder \citep{p47} \\
AEE & Additive Expert Ensemble \citep{kolter2005using} \\
AIS & Artificial Immune System \\
AIS-Clus & Artificial Immune System - Clustering \citep{p3,p10} \\
API & Application Programming Interface \\
ARIMA & Auto-regressive Integrated Moving Average \\
ARL & Average run length \\
ASSED & Adaptive Social Sensor Event Detection \citep{p11} \\
AUC & Area Under the Curve \\
AWILDA & Adaptive Window based Incremental LDA \citep{p1} \\
BERT & Bidirectional Encoder Representation from Transformers \citep{devlin2018bert} \\
BOW & Bag-of-words \\
BSP & Balancing Stability and Plasticity \citep{p27} \\
CBOW & Continuous bag-of-words \citep{mikolov2013efficient} \\
CCOHA & Clean Corpus of Historical American English \\
CFS & Correlation-based feature selection \\
CNB & Complement Naive Bayes \\
CNN & Convolutional Neural Network \\
CRQA & Cross Reference Quantification Analysis \citep{p-25} \\
CUSUM & Cumulative sum \\
DBSCAN & Density-Based Spatial Clustering of Applications with Noise \\
DC & Drift categories \\
DCFS & Dynamic Correlation-based Feature Selection \\
DD & Drift detection types \\
DDAW & Drift Detection-based Adaptive Window \citep{p29,p40} \\
DDM & Drift detection method \\
DPMM & Dirichlet Process Multinomial Mixture \\
DSM & Data stream mining \\
EC & Exclusion criteria \\
EDDM & Early Drift Detection method \citep{baena2006early} \\
EWMA & Exponentially Weighted Moving Average \\
EWNStream+ & Evolutionary Word relation Network for short text Streams clustering \citep{p24} \\
FAR & False alarms rate \\
FFCA & Fuzzy Formal Concept Analysis \citep{p37} \\
FNN & Feedforward Neural Network \\
GCTM & Graph Convolutional Topic Model \citep{p26} \\
GDWE & Graph-based Dynamic Word Embeddings \citep{p31} \\
GloVe & Global Vectors \citep{pennington2014glove} \\
GPU & Graphic Processing Unit \\
HDDM & Hoeffding-inequality-based Drift Detection Method \citep{frias2014online} \\
IC & Inclusion criteria \\
IS & Intelligent Systems \\
IWC & Incremental Word-Context \citep{p32} \\
kNN & k-Nearest Neighbors \\
KSWIN & Kolmogorov-Smirnov Windowing \citep{raab2020reactive} \\
LAMBADA & Language-model-based data augmentation \citep{anaby2020not} \\
LDA & Latent Dirichlet Allocation \citep{blei2003latent} \\
LLM & Large language model \\
LPP & Log Predictive Probability \\
LSTM & Long short-term memory \\
MDR & Missing detection rate \\
ML & Machine Learning \\
MLM & Masked language modeling \\
MOA & Massive Online Analysis \citep{bifet2010moa,bifet2018machine} \\
MSE & Mean squared error \\
MTD & Mean time to detection \\
MTFA & Mean time between false alarms \\
MU & Model update \\
NCD & Normalized Compression Distance \\
NLP & Natural Language Processing \\
NMI & Normalized Mutual Information \\
NOAA & National Oceanic and Atmospheric Administration \\
NPMI & Normalized Pointwise Mutual Information \\
OBAL & Online Batch-based Active Learning \citep{p2} \\
OFSER & Online Feature Selection with Evolving Regularization \citep{p19} \\
OM & Opinion mining \\
OOV & Out-of-vocabulary \\
OSMTS & Online Semi-Supervised Classification on Multilabel Text Streams \citep{p43} \\
PCA & Principal component analysis \\
PH & Page-Hinkley \\
PPMI & Positive Pointwise Mutual Information \citep{p32} \\
PSDVec & Positive-Semidefinite Vectors \citep{li2017psdvec} \\
PV-DBOW & Paragraph vector - Distributed bag-of-words \citep{le2014distributed} \\
PV-DM & Paragraph vector - Distributed memory \citep{le2014distributed} \\
RDDM & Reactive drift detection method \citep{barros2017rddm} \\
RoBERTa & Robustly Optimized BERT Pre-training Approach \citep{liu2019roberta} \\
ROC & Receiver Operating Characteristic \\
RQ & Research Question \\
SA & Sentiment analysis \\
SMAFED & Social Media Analysis Framework for Event Detection \citep{p34} \\
SPC & Statistical Process Control \\
SVM & Support Vector Machine \\
t-SNE & t-Distributed Stochastic Neighbor Embedding \\
TCR-M & Topic Change Recognition-based Method \citep{p44} \\
TF-IDF & Term frequency-Inverse Document Frequency \\
TR & Text representation \\
TRUS & Text representation update scheme \\
TSDA-BERT & Twitter Sentiment Drift Analysis - BERT \citep{p38} \\
VFDT & Very Fast Decision Tree \citep{gama2006decision} \\
WIDID & What is Done is Done \citep{p35} \\\hline
% Continue inserindo suas siglas aqui
\end{longtable}

\end{document}